\pdfoutput=1

\documentclass[11pt]{article}

\usepackage[final]{acl}

\usepackage{times}
\usepackage{latexsym}

\usepackage[T1]{fontenc}

\usepackage[utf8]{inputenc}

\usepackage{microtype}

\usepackage{inconsolata}

\usepackage{graphicx}

\usepackage{multirow}
\usepackage[normalem]{ulem}
\useunder{\uline}{\ul}{}

\usepackage{subcaption}
\usepackage{xcolor}
\usepackage{amsmath,amssymb}
\usepackage{pifont}
\usepackage{tikz}
\usetikzlibrary{shapes.geometric, shapes.misc, arrows.meta, positioning}
\usepackage{comment}
\usepackage{caption}
\usepackage{placeins}

\title{MATCHED: Multimodal Authorship-Attribution To Combat Human Trafficking in Escort-Advertisement Data}

\author{Vageesh Saxena \\
  Law \& Tech Lab \\
  Maastricht University \\
  v.saxena@maastrichtuniversity.nl \\\And
  Benjamin Bashpole \\
  Bashpole Software, Inc. \\    
  bashpole@idtraffickers.com \\ \AND
  Gijs Van Dijck \\
  Law \& Tech Lab \\
  Maastricht University \\
  gijs.vandijck@maastrichtuniversity.nl \\
  \And
 Gerasimos Spanakis \\
  Law \& Tech Lab \\
  Maastricht University \\
  jerry.spanakis@maastrichtuniversity.nl \\
  }

\definecolor{pretrained}{HTML}{FF0000}
\definecolor{ce}{HTML}{0000FF}
\definecolor{supcon}{HTML}{00FF00}
\definecolor{cesupcon}{HTML}{FFA500}
\definecolor{blipdeclutrvit}{HTML}{ABAD48}
\definecolor{declutrvit}{HTML}{3AA8C1}
\definecolor{triplet}{HTML}{FFFF00}
\definecolor{cetriplet}{HTML}{FF00FF}

\begin{document}
\maketitle
\begin{abstract}
Human trafficking (HT) remains a critical issue, with traffickers increasingly leveraging online escort advertisements (ads) to advertise victims anonymously. Existing detection methods, including Authorship Attribution (AA), often center on text-based analyses and neglect the multimodal nature of online escort ads, which typically pair text with images. To address this gap, we introduce MATCHED, a multimodal dataset of 27,619 unique text descriptions and 55,115 unique images collected from the Backpage escort platform across seven U.S. cities in four geographical regions. Our study extensively benchmarks text-only, vision-only, and multimodal baselines for vendor identification and verification tasks, employing multitask (joint) training objectives that achieve superior classification and retrieval performance on in-distribution and out-of-distribution (OOD) datasets. Integrating multimodal features further enhances this performance, capturing complementary patterns across text and images. While text remains the dominant modality, visual data adds stylistic cues that enrich model performance. Moreover, text-image alignment strategies like CLIP and BLIP2 struggle due to low semantic overlap and vague connections between the modalities of escort ads, with end-to-end multimodal training proving more robust. Our findings emphasize the potential of multimodal AA (MAA) to combat HT, providing LEAs with robust tools to link ads and disrupt trafficking networks. 

\end{abstract}

\section{Introduction}

Human trafficking (HT) is a widespread crime that exploits vulnerable individuals of all ages and genders \citep{Europol, Undoc}. Among its various forms, sex trafficking is notably prevalent, with traffickers coercing victims into commercial sex through violence, threats, deception, and debt bondage \citep{ILO}. Women and girls are particularly affected, especially within the commercial sex industry \citep{ILO}. With the rise of digital platforms, traffickers increasingly exploit online advertisements (ads) to take advantage of anonymity, with approximately 65\% of HT victims in the United States being advertised through online escort services \citep{Polaris_stats2020}. The sheer volume of online ads makes manual tracking a daunting task, causing many trafficking cases to go undetected \citep{Polaris_stats2018}. 

\hspace{\parindent} Direct detection of HT instances through end-to-end classification techniques demonstrates assuring performances \citep{7745456, tong-etal-2017-combating, alvari2017semisupervisedlearningdetectinghuman}, but reliance on expert-generated labels risks overfitting, failing to generalize beyond training data. To address this challenge, LEAs and researchers have developed a list of HT indicators for identifying suspicious ads \citep{6758797, 7752332, jrsa}. However, these indicators can only be studied in a group of ads linked to individuals or trafficking networks. While traditional methods suggest connecting these ads through phone numbers and email addresses \citep{chambers-etal-2019-character}, recent research demonstrates that only 37\% of ads contain such identifiers \citep{saxena-etal-2023-idtraffickers}. Other supervised \citep{Nagpal2015AnER, li-etal-2022-extracting, liu-etal-2023-sweet} and unsupervised techniques \citep{rabbany2018active, 10.1145/3487553.3524263, 9912645} often rely on explicit similarities, like name, identical phrases, or near duplicates, to connect these escort ads, limiting their effectiveness, particularly when vendors alter details to evade detection. Another research direction suggests using AA approaches, offering a holistic approach by identifying unique language patterns and stylistic features consistent across ads from the same vendor or vendor group. Existing AA approaches in NLP have demonstrated considerable success by examining these subtle written expressions, allowing for stronger connections between ads even when explicit markers differ \citep{ardakani2020identifying, saxena-etal-2023-idtraffickers}. 

\noindent However, a key limitation in AA research for linking escort ads is the underutilization of multimodal data—specifically, the integration of images with text—even though most escort ads include a title, ad description, and multiple images. Combining textual and visual data can provide a richer context for AA, as images often reveal stylistic consistencies, locations, or poses that uniquely characterize a vendor’s profile. Incorporating images alongside text thus strengthens the “anchors” needed for linking ads. In larger trafficking networks, vendors may reuse images with varying writing patterns in text ads, providing a visual link for AA models to leverage. Conversely, vendors might pair similar text descriptions with different escort images. Additionally, existing research expects a minimum of five ads per vendor for effective AA \citep{saxena-etal-2023-idtraffickers}. Since most ads already include multiple images, adopting Multimodal AA (MAA) can capitalize on these data sources, improving performance beyond single-modality approaches for vendors with lower ad frequencies. Through this work, we aim to support LEAs in building AA-driven knowledge graphs and enable targeted investigations across extensive collections of raw escort ads by making the following contributions:

\noindent \textbf{(i) MATCHED Dataset and Comprehensive Benchmarking:} We present MATCHED, a novel dataset for MAA, featuring 27,619 unique text descriptions and 55,115 images from Backpage escort ads collected across seven U.S. cities between December 2015 and April 2016. Extensive experiments establish benchmarks across text, vision, and multimodal domains, with evaluations conducted on in-distribution and Out-Of-Distribution (OOD) datasets. MATCHED provides a robust foundation for future research in MAA. Due to the dataset's sensitivity, the anonymized metadata is attached via \href{https://doi.org/10.34894/UR3RVE}{DataverseNL}, while the full dataset will remain restricted. Code for the experiments is available on \href{https://github.com/vageeshSaxena/MATCHED.git}{MATCHED}.

\noindent \textbf{(ii) Enhanced Performance through Multitask Training:} We propose a joint multitask framework that simultaneously optimizes vendor identification and verification, outperforming traditional single-task models by 1.61\% (text) and 1.52\% (vision) on macro-F1 score for classification and 1.68\% (text) and 6.75\% (vision) on R-Precision for retrieval task. Although these gains may seem subtle, this dual-focus approach empowers LEAs to identify known vendors and discover emerging ones in OOD ads, enhancing their investigative capabilities.

\noindent \textbf{(iii) Advancements in Model Performance through Multimodal Training:} Traditional AA methods rely heavily on textual data, often ignoring valuable stylistic cues from images and excluding vendors with fewer ads. Our multimodal approach integrates text and image data, improving performance even for vendors with limited postings. Pairing a single text description with multiple images (e.g., one text with five images produces five samples) expands the training set and enriches feature representation. While text remains the dominant modality, incorporating images with text enhances text-only results by 5.43\% on retrieval R-Precision, marginally improves vision-only results by 0.75\% on retrieval R-Precision, and increases classification macro-F1 by 32.62\%—ultimately providing a more comprehensive and robust AA framework.

\section{Related Research}
AA in NLP has evolved from basic stylometric analysis \citep{bhargava2013stylometric, ramnial2016authorship} to advanced models that detect distinct linguistic patterns for identifying authorship across text segments \citep{fabien-etal-2020-bertaa, ai-etal-2022-whodunit, wegmann-etal-2022-author}. AA’s applications include forensic linguistics, where it attributes authorship in forensic contexts \citep{IQBAL2008S42, Nirkhi2013, Fobbe2021TextLinguisticAI}, to cybersecurity and cybercrime, where AA tracks malicious actors and identifies criminal activity across platforms \citep{10.1145/3308558.3313537, saxena-etal-2023-vendorlink}. However, a unique challenge emerges in applying AA within online criminal markets: conventional models struggle to efficiently capture the specialized jargon, coded language, and noise of illicit environments like illegal marketplaces, necessitating models that recognize these nuanced linguistic shifts \citep{choshen-etal-2019-language, manolache2022veridarklargescalebenchmarkauthorship}. This limitation underscores the need for fine-tuned models that capture the nuanced linguistic shifts and stylistic patterns unique to these illicit activities. Addressing this gap, \citet{ardakani2020identifying} propose supervised neural networks for AA on Backpage escort ads, demonstrating the potential to uncover stylistic consistencies even when explicit identifiers are altered. Building on this, \citet{saxena-etal-2023-idtraffickers} leverage state-of-the-art transformer-based models for vendor identification and verification, showcasing their ability to link ads across escort markets spanning 41 cities effectively.

\hspace{\parindent} In addition to text, images in criminal markets may reveal recurring stylistic patterns that can help identify vendors' accounts \citep{10546301}. Vision-based AA approaches can leverage visual patterns—such as backgrounds, lighting, or object placement—that complement linguistic cues, especially when text data is sparse or inconsistent \citep{10.1145/3196494.3196529}. MAA approaches, on the other hand, combine text and images, enhancing accuracy by merging stylistic patterns across media. They can link and create features that account for linguistic and visual patterns, creating a more comprehensive vendor profile \citep{10.1145/3308558.3313537}. 

\hspace{\parindent} Building on these insights, we introduce a multimodal dataset, MATCHED, of escort ads collected from seven U.S. cities across four geographical regions. This dataset and our multitask training approach establish new benchmarks in text, vision, and multimodal domains for escort market ads, setting a strong foundation for future research in MAA. Our models optimize vendor identification (a classification task that assesses the likelihood of an ad belonging to a specific vendor within a candidate set) and vendor verification (a similarity task that assesses whether two ads came from the same vendor). By employing a multitask training objective, our approach allows LEAs to identify known vendors in a closed-set environment while connecting emerging vendors across out-of-distribution ads for open-set scenarios. Integrating multimodal data, especially for vendors with limited text ads, further enhances model performance by creating multiple samples per ad. This comprehensive approach leverages textual and visual cues, enabling LEAs to track HT networks more precisely across various online markets and platforms, laying the groundwork for advanced AA research.

\section{Dataset}
\label{sec:dataset}

\begin{table}[h]
\resizebox{\linewidth}{!}{%
\begin{tabular}{c|ccccc}
\textbf{Regions} & \textbf{Ads} & \textbf{Text} & \textbf{Images} & \textbf{\% Faces} & \textbf{Vendors} \\ \hline
South            & 14088        & 13661         & 27423           & 0.4928            & 1450             \\
Midwest          & 8564         & 8259          & 14883           & 0.5542            & 1008             \\
West             & 3262         & 3153          & 5049            & 0.6052            & 507              \\
Northeast        & 2599         & 2546          & 7760            & 0.6183            & 584              \\ \hline
All              & 28513        & 27619         & 55115           & 0.5676            & 3549            
\end{tabular}
}
\caption{Number of advertisements, unique text descriptions, images, \% of Faces in the image datasets, and vendors per region in the MATCHED dataset. }
\label{tab:data-stats}
\end{table}

\noindent \citet{jrsa} provides compelling evidence that Backpage escort advertisements are frequently associated with HT activities. Motivated by this finding, we focus our experiments on Backpage ads and curate a dataset of 28,513 advertisements, including 27,619 unique text descriptions and 55,115 unique images linked to 3,549 vendors. Approximately 56\% of these images feature an escort’s face, while the remainder display only parts of their body (without any faces). Following \citet{saxena-etal-2023-idtraffickers}, we utilize \citet{chambers-etal-2019-character} to extract phone numbers and apply NetworkX \citep{SciPyProceedings_11} to form vendor communities. Each community is assigned a unique vendor label, establishing the ground truth for AA tasks across the dataset. The ads are collected from seven major U.S. cities—Chicago, Houston, Detroit, Dallas, San Francisco, New York, and Atlanta—representing four geographic regions: South, Midwest, West, and Northeast. These zones group ads from relevant cities, with average text sequence lengths of 125, 118, 113, and 132 tokens, respectively. Table \ref{tab:data-stats} provides detailed insights for each regional dataset. This organization ensures geographic diversity and allows for a more nuanced AA analysis across different regions.

\hspace{\parindent} The south region dataset, which contains the most text and image ads, is selected as the primary dataset for model training and in-distribution evaluation. We use the remaining Midwest, West, and Northeast datasets as OOD datasets to assess our model's capability to adapt to new distributions. Notably, many vendors in our dataset appear across multiple geographic regions. As a result, the OOD datasets include ads from vendors present in the training dataset (South) and additional vendors unique to each region. Further dataset details can be found in the appendix \ref{app:dataset} and \ref{app:datasheet}.

\section{Experimental Setup}
In our approach to vendor identification and verification tasks, we conduct our experimental setup through a closed-set classification task using the south region dataset, where the model learns to identify vendors from a pre-defined candidate set and an open-set metric learning task, where we leverage the ad representations (embeddings) from the trained classifier to perform similarity searches that retrieve all relevant ads from a vendor's collection. Before establishing the multimodal benchmark, we establish individual baselines in the text-only and vision-only modalities for a fair comparison. With practical utility in mind, the baselines are determined based on the performance of the trained model on vendor identification and verification objectives, providing a point of reference for identifying vendors in LEA databases and connecting them to new, emerging vendors. 

\noindent \textbf{(i) Vendor Identification - A Closed-Set Classification Task:} For the vendor identification, we perform a multi-class classification using pre-trained backbones on the South region dataset using cross-entropy (CE) loss \citep{juola2005controlled}. Following \citet{ai-etal-2022-whodunit}, we also experiment with a multitask joint objective combining CE, Supervised Contrastive (SupCon) \citep{ye2023supervised, huertas2024understanding}, and Triplet losses \citep{hu2020deepstyle, tyo2021siamese} to enhance feature discrimination by aligning representations of the same vendor while separating those of different vendors. 

\noindent \textbf{(ii) Vendor Verification - An Open-Set Metric Learning Task:} Similar to \citet{wegmann-etal-2022-author}, we employ contrastive learning for a metric learning task \citep{kaya2019deep} using Triplet and SupCon losses on the South region dataset. The objective is to bring together ad representations from the same vendor while pushing apart representations from different vendors. 

\noindent \textbf{(iii)(A) Text-Only Baselines:} Following \citet{saxena-etal-2023-idtraffickers}, we establish text-only baselines using the Style-Embedding \citep{wegmann-etal-2022-author} and DeCLUTR-small \citep{giorgi-etal-2021-declutr} backbones. For vendor identification, we fine-tune these models with CE, CE+Triplet, and CE+SupCon objectives. For vendor verification, we apply Triplet and SupCon objectives.

\noindent \textbf{(iii)(B) Vision-Only Baselines:} In the absence of prior vision-only baselines on similar datasets, we fine-tune VGG-16 \citep{simonyan2015deepconvolutionalnetworkslargescale}, ResNet-50 \citep{he2015deepresiduallearningimage}, DenseNet-121 \citep{huang2018denselyconnectedconvolutionalnetworks}, InceptionNetV3 \citep{szegedy2015rethinkinginceptionarchitecturecomputer}, EfficientNetV2 \citep{tan2021efficientnetv2smallermodelsfaster}, ConvNext-small \citep{woo2023convnextv2codesigningscaling}, and ViT-base-patch16-244 \citep{dosovitskiy2021imageworth16x16words} 
backbones with CE, CE+Triplet, and CE+SupCon objectives for the vendor identification task on the South Dataset. Once again, we employ Triplet and SupCon objectives on these backbones for the vendor verification task.

\noindent \textbf{(iii)(C) Multimodal Baselines:} We evaluate two widely-used multimodal architectures, VisualBERT \citep{li2019visualbertsimpleperformantbaseline} and ViLT \citep{kim2021viltvisionandlanguagetransformerconvolution}, fine-tuned on the South dataset for vendor identification using CE loss. Building on insights from unimodal experiments, we also design a custom backbone, "DeCLUTR-ViT", which combines DeCLUTR for text and ViT for images, leveraging their strong performance in respective modalities. To integrate text and image representations, we explore four fusion strategies: concatenation \citep{8615789, li2024compact}, mean pooling \citep{10.1145/3543848}, self-attention \citep{kiela2020supervisedmultimodalbitransformersclassifying, GAN2024122731}, and adaptive auto fusion via a neural network \citep{sahu-vechtomova-2021-adaptive}, enabling nuanced cross-modal interactions by combining complementary signals. Additionally, we perform image-text alignment pre-training tasks using the DeCLUTR-ViT backbone on the combined dataset from all regions, applying three alignment strategies: Image-Text Contrastive (ITC, or CLIP) \citep{radford2021learningtransferablevisualmodels}, ITC+ITM (Image-Text Contrastive and Image-Text Matching) \citep{villegas2024improvingmultimodalclassificationsocial}, and BLIP2 \citep{li2023blip2bootstrappinglanguageimagepretraining}. These alignment techniques ensure that text and images from the same ad are represented closely in the latent space, particularly when a single text ad is associated with multiple images. Finally, we fine-tune the aligned models on the South dataset for vendor identification using CE and CE+SupCon objectives, effectively combining alignment and fusion strategies to enhance performance.

\noindent \textbf{(iv) Evaluation:} To address the class imbalance in our dataset (Appendix Figure \ref{fig:dataStats}(C)), we prioritize the Macro-F1 metric for the classification task. Additionally, we assess classification, metric-learning, and text-image alignment baselines through a retrieval task, linking query ads to relevant ads from the same vendor. The dataset is split into training ("documents") and test ("queries") sets, with embeddings generated by trained models used to compute cosine similarity via FAISS-based search \citep{johnson2019billion}. Text-only and vision-only baselines extract embeddings directly from their respective encoders, while multimodal baselines combine text and vision embeddings from the DeCLUTR-ViT backbone using mean pooling. For ITC+ITM and BLIP2-based baselines, we take these vision embeddings from the QFormer encoder. Retrieval tasks are categorized as text-to-text, image-to-image, or multimodal based on whether query and document embeddings are derived from text, vision, or pooled multimodal representations. This framework evaluates the models' ability to link ads to vendors, including unseen ones, offering robust insights into performance across tasks.

The retrieval task is evaluated using R-Precision@X, which measures precision when the number of retrieved items equals the number of relevant ads per vendor, with higher scores reflecting more accurate representations of vendor activity \citep{saxena-etal-2023-idtraffickers}. Additionally, Mean Reciprocal Rank (MRR@10) evaluates the average ranking position of the first ten correctly retrieved ads for each query, with scores closer to 1 indicating higher relevance ranking, thereby reducing manual search efforts for LEAs \citep{striebel-etal-2024-scaling}. Lastly, Macro-F1@X independently calculates F1 scores for each vendor class and averages them, ensuring equal weight for all vendors regardless of sample size. In Macro-F1@X and R-Precision@X, X represents the cutoff, defined as the number of relevant items per vendor. 

\section{Results}
\label{sec:results}
This section evaluates text-only, vision-only, and multimodal baselines for vendor identification and verification tasks. The classification task leverages CE, CE+Triplet, and CE+SupCon objectives to enhance feature discrimination among vendor classes. For metric learning, Triplet and SupCon contrastive losses are employed to cluster representations of the ads from the same vendor while distinguishing those of different vendors. Retrieval performance is assessed using similarity search and metrics such as MRR@10, R-Precision@X, and Macro-F1@X across the South, Midwest, West, and Northeast datasets. The Zero-Shot (ZS) average reflects model performance across all datasets without task-specific training. In contrast, the OOD average score evaluates how models trained on the South dataset generalize to unseen ads and vendors in the Midwest, West, and Northeast datasets. Since classification models cannot handle unseen vendors in OOD scenarios, OOD performance is reported solely through retrieval metrics. For detailed quantitative insights and performance comparisons, we direct our readers to Appendix Tables \ref{tab:all_vendor_identification_results}-\ref{tab:multimodal_retrieval}.

\begin{table}[]
\centering
\resizebox{\linewidth}{!}{%
\begin{tabular}{|ccc|}
\hline
\multicolumn{1}{|c|}{\textbf{Model}}                                                                                                   & \multicolumn{1}{c|}{\textbf{Loss}}                             & \textbf{Macro-F1}                      \\ \hline
\multicolumn{3}{|c|}{\textbf{Text-Baseline}}                                                                                                                                                                                                     \\ \hline
\multicolumn{1}{|c|}{{\color[HTML]{303498} }}                                                                                          & \multicolumn{1}{c|}{CE}                                        & 0.6379                                 \\ \cline{2-3} 
\multicolumn{1}{|c|}{{\color[HTML]{303498} }}                                                                                          & \multicolumn{1}{c|}{CE+Triplet}                                & 0.5503                                 \\ \cline{2-3} 
\multicolumn{1}{|c|}{\multirow{-3}{*}{{\color[HTML]{303498} \textbf{DeCLUTR-small}}}}                                                  & \multicolumn{1}{c|}{{\color[HTML]{303498} \textbf{CE+SupCon}}} & {\color[HTML]{303498} \textbf{0.6540}} \\ \hline
\multicolumn{3}{|c|}{\textbf{Vision-Baseline}}                                                                                                                                                                                                   \\ \hline
\multicolumn{1}{|c|}{{\color[HTML]{303498} }}                                                                                          & \multicolumn{1}{c|}{CE}                                        & 0.6142                                 \\ \cline{2-3} 
\multicolumn{1}{|c|}{{\color[HTML]{303498} }}                                                                                          & \multicolumn{1}{c|}{CE+Triplet}                                & 0.6378                                 \\ \cline{2-3} 
\multicolumn{1}{|c|}{\multirow{-3}{*}{{\color[HTML]{303498} \textbf{ViT-base-patch16}}}}                                               & \multicolumn{1}{c|}{{\color[HTML]{303498} \textbf{CE+SupCon}}} & {\color[HTML]{303498} \textbf{0.6294}} \\ \hline
\multicolumn{3}{|c|}{\textbf{Multimodal-Baselines}}                                                                                                                                                                                              \\ \hline
\multicolumn{1}{|c|}{{\color[HTML]{303498} }}                                                                                          & \multicolumn{1}{c|}{CE}                                        & 0.9670                                 \\ \cline{2-3} 
\multicolumn{1}{|c|}{\multirow{-2}{*}{{\color[HTML]{303498} \textbf{\begin{tabular}[c]{@{}c@{}}End2End \\ DeCLUTR-ViT\end{tabular}}}}} & \multicolumn{1}{c|}{{\color[HTML]{303498} \textbf{CE+SupCon}}} & {\color[HTML]{303498} \textbf{0.9802}} \\ \hline
\multicolumn{1}{|c|}{DeCLUTR-ViT}                                                                                                      & \multicolumn{1}{c|}{BLIP2+CE+SupCon}                           & 0.9420                                 \\ \hline
\end{tabular}
}
\caption{Macro-F1 performance of the text, vision, and multimodal classifiers on the south region dataset. The benchmarks are highlighted by \color[HTML]{303498} \textbf{color}.}
\label{tab:classifier_performance}
\end{table}

\noindent \textbf{(i) Classification task:} Consistent with the prior findings \citep{saxena-etal-2023-idtraffickers}, the DeCLUTR outperforms the Style-Embedding backbone with CE loss amongst the text-baselines by about 12\%. Furthermore, as shown in text-baseline results in Table \ref{tab:classifier_performance}, the DeCLUTR model performs the best when trained with the CE+SupCon joint objective.

\hspace{\parindent} For the vision baseline, ResNet-50 with CE loss achieves the highest macro-F1 score (0.6394) among classifiers, followed by EfficientNetV2 (0.6285), DenseNet-121 (0.6262), ConvNext-small (0.6215), and ViT-base-patch16 (0.6141). Despite its slight underperformance in classification tasks, our analysis in appendix table \ref{tab:image-baselines-retrieval} reveals that ViT-base-patch16 outperforms all other models in retrieval tasks for both in-distribution and OOD datasets. This finding aligns with prior research \citep{gkelios2021investigatingvisiontransformermodel, elnouby2021trainingvisiontransformersimage}, which highlights ViT's ability to produce rich, contextualized embeddings that capture global relationships and stylistic patterns, even across diverse visual data (e.g., images with or without faces), making ViT-base-patch16 the most suitable backbone for our task. Finally, as shown in Table \ref{tab:classifier_performance}, the ViT baseline trained with the CE+Triplet objective achieves the best macro-F1 score of 0.6378, with CE+SupCon closely following at 0.6294.

\hspace{\parindent} The multimodal DeCLUTR-ViT backbone, trained end-to-end with mean pooling as the fusion technique, achieves the highest macro-F1 score (0.9670) on the South region dataset, surpassing VisualBERT and ViLT. Notably, the DeCLUTR-ViT baselines pre-trained with text-image alignment strategies, such as CLIP, ITC+ITM, or BLIP2, underperform when fine-tuned for the authorship identification task, though the BLIP2-pretrained backbone comes closest to matching DeCLUTR-ViT’s performance (0.9420). When trained with the joint CE+SupCon objective, the DeCLUTR-ViT backbone demonstrates exceptional robustness in capturing multimodal relationships. This strong performance may also be attributed to the dataset’s structure, where each text ad is paired with multiple images and vice versa, ensuring the model encounters diverse combinations during training.

\newsavebox{\cetikzbox}
\sbox{\cetikzbox}{\tikz[baseline] {\draw[ce, line width=0.8mm] (-0.1,0) -- (0.1,0) (0,-0.12) -- (0,0.12);}}

\newsavebox{\declutrvittikzbox}
\sbox{\declutrvittikzbox}{\tikz[baseline] {\draw[declutrvit, line width=0.8mm] (-0.1,0) -- (0.1,0) (0,-0.12) -- (0,0.12);}}

\newsavebox{\blipdeclutrvittikzbox}
\sbox{\blipdeclutrvittikzbox}{\tikz[baseline] {\draw[blipdeclutrvit, line width=0.8mm] (-0.1,0) -- (0.1,0) (0,-0.12) -- (0,0.12);}}

\begin{figure*}[!htbp] 
    \centering
    \includegraphics[width=\textwidth,height=\textheight,keepaspectratio]{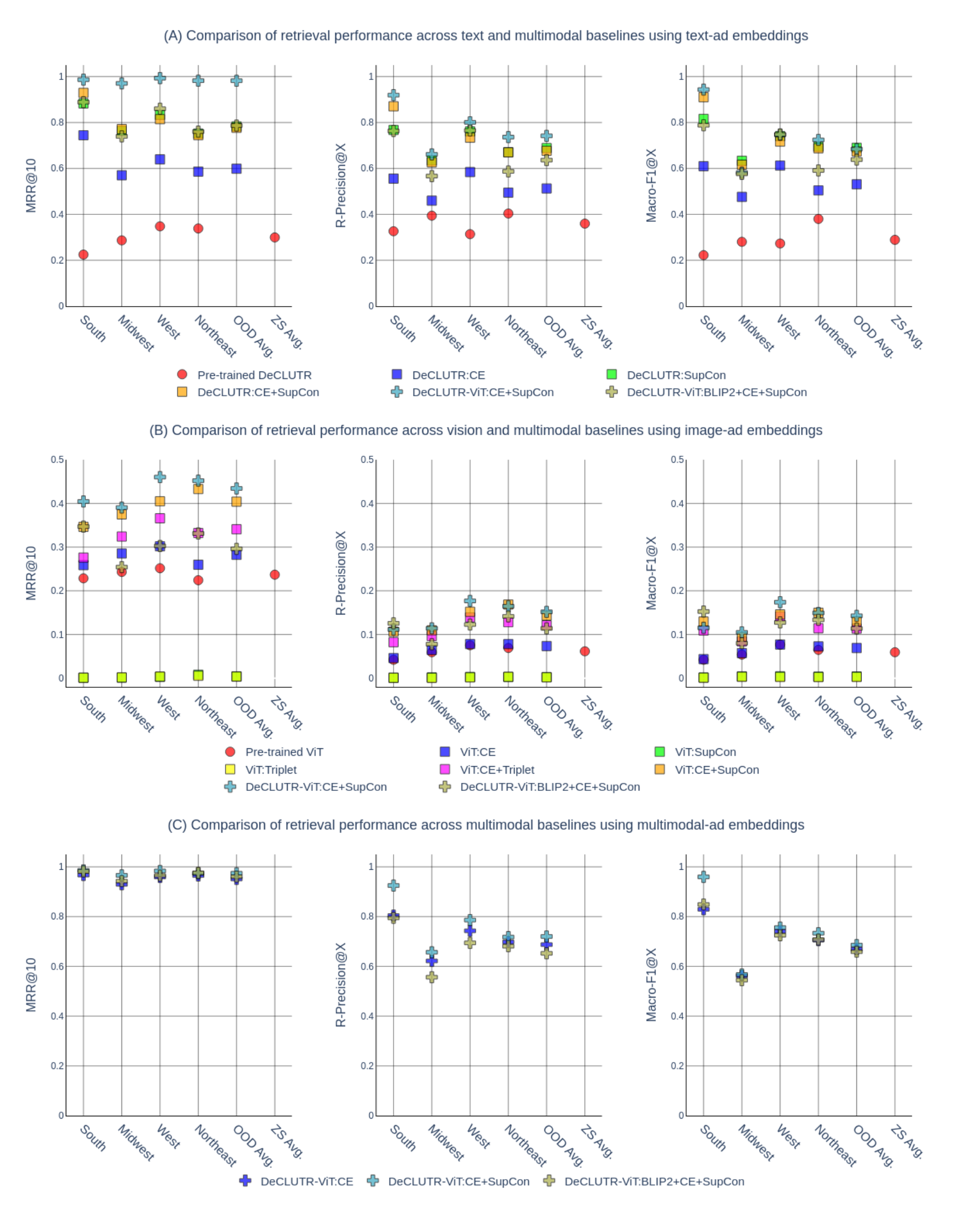}
    \captionsetup{width=\textwidth, justification=centering} 
    \caption{Comparison of retrieval performance across multiple baselines for text-to-text, image-to-image, and multimodal ads retrieval tasks on South, Midwest, West, and Northeast datasets. The text-to-text retrieval baselines include the pre-trained DeCLUTR checkpoint ({\color{pretrained}\ding{108}}), DeCLUTR classifiers trained on CE ({\color{ce}\ding{110}}) and CE+SupCon losses ({\color{cesupcon}\ding{110}}), and the DeCLUTR backbone trained with SupCon loss ({\color{supcon}\ding{110}}). Image-to-image retrieval baselines include the pre-trained ViT checkpoint ({\color{pretrained}\ding{108}}), ViT classifiers trained on CE ({\color{ce}\ding{110}}), CE+Triplet ({\color{cetriplet}\ding{110}}), and CE+SupCon losses ({\color{cesupcon}\ding{110}}), and ViT backbones trained with SupCon ({\color{supcon}\ding{110}}) and Triplet ({\color{triplet}\ding{110}}) losses. Multimodal baselines include End2End DeCLUTR-ViT classifiers trained with CE (\usebox{\cetikzbox}), CE+SupCon (\usebox{\declutrvittikzbox}), and BLIP2-aligned DeCLUTR-ViT classifiers trained with CE+SupCon (\usebox{\blipdeclutrvittikzbox}) objectives.}
    \label{fig:allRetrievalPlot} 
\end{figure*}

\noindent \textbf{(ii) Retrieval Task:} Since the metric-learning task employs triplet and SupCon losses, its effectiveness is assessed by the model's ability to cluster ad representations based on stylometric patterns from the same vendor. Similarly, the joint-objective classifiers also incorporate classification and metric-learning losses, enabling direct comparison between the baselines through a retrieval task. 

\hspace{\parindent} Figure \ref{fig:allRetrievalPlot}(A) compares the text-to-text retrieval performance of text-only pre-trained ({\color{pretrained}\ding{108}}), fine-tuned, and multimodal baselines. Fine-tuning on the South region dataset significantly improves performance across all metrics. Among text-only baselines, the DeCLUTR backbone trained with the joint CE+SupCon objective ({\color{cesupcon}\ding{110}}) outperforms the CE-only baseline ({\color{ce}\ding{110}}) and performs on-par with the SupCon-only baseline ({\color{supcon}\ding{110}}) on OOD avg score, while surpassing it on the training dataset. Given the consistent performance of the DeCLUTR backbone with CE+SupCon objective on the classification and retrieval task, we establish it as the benchmark for text-only modality. This benchmark is further compared against the text representations from the multimodal DeCLUTR-ViT backbone trained end-to-end with CE+SupCon (\usebox{\declutrvittikzbox}) and the fine-tuned DeCLUTR-ViT backbone, pre-trained for text-image alignment task using BLIP2 objective (\usebox{\blipdeclutrvittikzbox}). The multimodal backbone trained end-to-end with CE+SupCon consistently outperforms all baselines on training and OOD datasets.

\hspace{\parindent} Figure \ref{fig:allRetrievalPlot}(B) highlights image-to-image retrieval performance, comparing vision-only pre-trained ({\color{pretrained}\ding{108}}), fine-tuned, and multimodal baselines. Fine-tuning on image ads also improves retrieval performance. Amongst vision-only baselines, the ViT backbone trained with the CE+SupCon objective ({\color{cesupcon}\ding{110}}) achieves superior performance over other baselines on both training and OOD datasets, establishing itself as the benchmark for the vision-only modality. Despite the better performance of the ViT backbone with CE+Triplet objective ({\color{cetriplet}\ding{110}}) on classification, it underperforms on the retrieval task. We further compare this vision benchmark against the vision representations from the multimodal DeCLUTR-ViT backbone trained end-to-end with CE+SupCon (\usebox{\declutrvittikzbox}) and the fine-tuned DeCLUTR-ViT backbone, pre-trained for text-image alignment task using BLIP2 objective (\usebox{\blipdeclutrvittikzbox}). The end-to-end multimodal backbone with CE+SupCon objective consistently outperforms other baselines on OOD datasets. However, it underperforms the fine-tuned BLIP2 baseline on R-Precision and Macro-F1 metrics for the training dataset.

\hspace{\parindent} Figure \ref{fig:allRetrievalPlot}(C) compares retrieval performance among multimodal baselines, evaluating the multimodal representation from the end-to-end multimodal DeCLUTR-ViT backbone trained end-to-end with CE+SupCon (\usebox{\declutrvittikzbox}) and the fine-tuned DeCLUTR-ViT backbone, pre-trained for text-image alignment task using BLIP2 objective (\usebox{\blipdeclutrvittikzbox}). The end-to-end multimodal backbone with CE+SupCon objective consistently outperforms the other baseline across the training and OOD datasets. Our analysis (Appendix Table \ref{tab:alignment_retreival_results}) indicates that the low performance of the text-image alignment strategies can be attributed to the lack of semantic similarity between images and text, as images in escort ads often do not directly reflect the context of the accompanying text.\\

\noindent \textbf{Takeaways:} Key takeaways from our results are: \\

\noindent \textbf{(i) MATCHED Dataset:} Fine-tuning on our dataset significantly improves task performance, emphasizing its importance and revealing the limitations of pre-trained checkpoints in adapting to its unique linguistic and stylistic patterns.

\noindent \textbf{(ii) Benchmarks:} Given the dual objective of vendor identification and verification, we establish DeCLUTR-small and ViT-base-patch16-244 backbones with CE+SupCon joint objective as our text- and vision-only benchmarks. The multimodal benchmark integrates these backbones with mean pooling as the fusion technique.

\noindent \textbf{(iii) Benefits of Joint Objective:} The CE+SupCon joint objective consistently outperforms or matches other objectives on both in-distribution and OOD datasets, demonstrating its robustness and generalization capability. Furthermore, employing joint objectives allows AA models to address closed-set vendor identification and open-set vendor verification tasks effectively.

\noindent \textbf{(iv) Effects of Multimodal Integration:} Integrating multimodal features significantly enhances AA performance across all tasks, outperforming text-only and vision-only benchmarks. Multimodal setups effectively leverage complementary features from both modalities, capturing richer and more comprehensive authorship patterns.

\noindent \textbf{(v) Text-Image Alignment Challenges:} Text-image alignment strategies, inspired by CLIP and BLIP2 research, struggle to connect text to image pairs in our dataset due to a lack of semantic similarity, as images in escort ads often do not directly reflect the context of the accompanying text. 

\noindent \textbf{(vi) Modality-Specific Performance:} In the multimodal setup, combining text and vision features improves vision retrieval performance compared to the vision-only backbone, though it remains unreliable. However, incorporating vision features into the text modality significantly enhances text retrieval performance, with text consistently delivering the best results across text, vision, and multimodal representations. This highlights the superiority of the text representations from the DeCLUTR-ViT backbone, making it the most effective option for retrieval tasks on our dataset.
    
\section{Discussion \& Further Insights}
This research introduces a novel multimodal dataset and conducts extensive benchmarking to demonstrate that multitask joint objectives and multimodal data integration enhance the performance of AA tasks on both in-distribution and OOD datasets. By linking escort ads through these techniques, we aim to assist researchers, investigators, and LEAs study HT indicators. Given the space constraints of the main manuscript, we prioritized presenting the most critical claims and experimental results. However, additional insights and detailed analyses are provided in the Appendix.

\hspace{\parindent} Specifically, data analysis, preprocessing steps, and a datasheet following \citep{gebru2021datasheetsdatasets} are detailed in Appendix sections \ref{app:dataset} and \ref{app:datasheet}. Given the sensitive nature of our research, we decided not to release any models publicly and, therefore, haven't attached any model cards for our research. Appendix \ref{app:experimental_details} outlines the training setup and computational considerations, while Appendix \ref{app:model_performance} presents comprehensive performance metrics for all baselines. Key insights into model behavior and learning dynamics are discussed in Appendix \ref{app:insights}, and the practical application of AA tasks in building knowledge graphs for investigative purposes is explored in Appendix \ref{app:utility} \footnote{While we provide extensive insights into our experiments through Appendix, we would like to specifically encourage readers to find:\\
1. Appendix Figure \ref{fig:data_stats_main} for key details about our dataset\\
2. Appendix Table \ref{tab:alignment_retreival_results} for text-to-image retrieval results highlighting the challenges of pre-trained text-image alignment task in binding modalities, and \\
3. Appendix Table \ref{tab:sanity_check_unique_and_common_performance} for an analysis demonstrating that our benchmark performs equally well for both shared and unique vendors across the South and OOD datasets.}. 

By structuring our paper this way, we aim to balance clarity and depth. The main manuscript provides a concise overview, while the supplementary material offers thorough details and extended findings. This approach ensures transparency, rigor, and accessibility, allowing domain experts and practitioners to derive value from our work.

\section{Conclusion}
Through this research, we demonstrate the potential of MAA in addressing the complexities of vendor identification and verification within online escort markets. Using our novel MATCHED dataset, we extensively benchmark text-only, vision-only, and multimodal approaches, showcasing the advantages of CE+SupCon multitask training objectives. Our analysis reveals that this dual-objective consistently outperforms single-task approaches across in-distribution and OOD datasets, enabling LEAs to identify known vendors while linking emerging ones in new markets. Additionally, multimodal integration significantly enhances model performance by capturing complementary patterns across text and images. While text remains the dominant modality, integrating image data along text descriptions adds stylistic cues that enrich the model's capabilities. Among text, vision, and multimodal representations, text representations from the DeCLUTR-ViT backbone emerge as the most effective for retrieval tasks, achieving the best results across all modalities. While pre-trained text-image alignment strategies like CLIP and BLIP2 fail to establish meaningful cross-modal connections due to low semantic overlap and ineffective use of stylistic features, end-to-end multitask training is a more robust approach for leveraging multimodal data in AA tasks. Finally, the performance gap between pre-trained checkpoints and fine-tuned baselines highlights the importance of domain-specific adaptations and task-specific training, providing a strong foundation for future research. By addressing real-world challenges and emphasizing scalability, we aim to equip LEAs with actionable tools to uncover and disrupt trafficking networks effectively.

\section{Limitations}
\label{sec:limitations}
\paragraph{Assumption:} Similar to existing research, our research assumes that each class label corresponds to a distinct vendor during the classification task, enabling the model to leverage domain knowledge effectively. However, our qualitative analysis identifies cases where the trained classifier misclassifies ads, likely due to similarities in writing style and content, suggesting the possibility that multiple vendors might belong to the same entity. While we lack definitive ground truth to confirm this hypothesis, it represents a notable challenge in ensuring label accuracy. We recognize that improving the quality of vendor labels would likely lead to enhanced benchmark performance and more robust model evaluations.

\paragraph{Dataset Limitations and Generalization Challenges:} Our research utilizes escort ads collected from the Backpage platform between December 2015 and April 2016. These ads, sourced from seven U.S. cities and categorized into four geographical regions, exhibit significant vendor overlap across regions, as shown in appendix figure \ref{fig:dataSimilarity}. Manual inspection also reveals the presence of near-duplicate ads that are challenging to identify and remove due to extensive noise and variability within the dataset. While we aim to evaluate the OOD generalization capabilities of our trained models, a more realistic scenario would involve incorporating ads from multiple escort platforms and broader geographical areas to better simulate cross-platform generalization performance. Finally, our dataset covers only briefly from December 2015 to April 2016. Extending the collection period would allow us to understand how vendors change their writing styles to evade detection, providing deeper insights into temporal dynamics that our current work does not address.

\paragraph{Selective Feature Extraction and Fine-Tuning:}
In this work, we extract text and vision representations exclusively from the final layers of our models, which may not fully capture nuanced features learned at earlier layers. Representations extracted from intermediate layers could yield different or potentially better outcomes. Additionally, while fine-tuning pre-trained text-image alignment models, we fine-tune all layers uniformly, which may not be optimal. Techniques like Centered Kernel Alignment (CKA) \citep{kornblith2019similarityneuralnetworkrepresentations} can provide insights into which layers learn the most relevant features, enabling more informed decisions about representation extraction and selective layer freezing during fine-tuning. Addressing these concerns is currently beyond the scope of this research, but we plan to explore these aspects in future work.

\paragraph{Computational Constraints:} While our research employs relatively large model architectures and advanced training strategies, it is limited by the computing resources available to us. Larger model architectures could potentially enhance performance across classification and retrieval tasks. However, when applied to text-image alignment tasks, the computational demands of scaling these models exceeded our resource capacity. As a result, we opted for smaller, more efficient architectures that fit within our computational constraints, ensuring a fair and balanced comparison across baselines. Similarly, our research relies heavily on contrastive learning objectives, and prior studies \citep{gao-etal-2021-scaling, vaessen2024effectbatchsizecontrastive} highlight the benefits of larger batch sizes for such tasks. However, to maintain consistency and fairness among baselines, we limited our batch size to 32, as larger sizes led to memory errors, particularly with text-image alignment models. This computational limitation also influenced our decision to forego fine-tuning pre-trained CLIP and BLIP2 checkpoints, as the memory requirements for fine-tuning BLIP2 architecture caused GPU crashes. These decisions reflect deliberate trade-offs made to ensure the reproducibility and fairness of our experimental comparisons while working within resource limitations.

\paragraph{Explainability:} Although this research does not explicitly address the explainability or interpretability of our models, we recognize their critical role in fostering trust among researchers, investigators, and law enforcement agencies. Previous studies \citep{saxena-etal-2023-idtraffickers} have explored explainability in AA through local feature attribution techniques applied to text ads. However, while numerous frameworks exist for explainability in unimodal data \citep{ribeiro2016whyitrustyou, lundberg2017unifiedapproachinterpretingmodel, kokhlikyan2020captumunifiedgenericmodel}, these methods cannot be directly extended to the multimodal AA context. Additionally, research highlights the limitations of existing explainability techniques, including their susceptibility to adversarial attacks, network sparsity, and inconsistencies in results \citep{das2020opportunities, krishna2022disagreement, saxena-etal-2023-vendorlink}. We aim to address these challenges in future work by developing a robust explainability framework tailored specifically for multimodal AA scenarios. Such a framework will help uncover the contributions of textual and visual features in decision-making processes, ensuring transparency and reliability in the application of multimodal AA models.

\paragraph{Generative Models:} Vendors can potentially leverage advanced Large Language Models (LLMs), like ChatGPT, to generate text ads with varying linguistic styles, making them harder to trace back to individual authors. While many publicly available LLMs restrict content generation for illegal purposes, vendors could fine-tune open-source LLMs or develop models tailored to evade AA by mimicking diverse stylistic patterns. This poses a significant challenge to our AA systems, which rely on detecting unique stylometric signatures. Similarly, the advent of vision-based generative models enables vendors to create or manipulate images to obscure identifiable stylistic cues. These advancements in generative technologies could diminish the effectiveness of both text- and vision-based AA models. In the future, we plan to recollect and analyze updated datasets to ensure our AA systems can differentiate between content generated by automated systems and authentic user-generated data, adapting to these emerging threats. 

\section{Ethical Considerations}
\label{sec:broader_impact}

\subsection{Data Protocols} 
We collect our dataset from the Backpage Escort Markets spanning seven U.S. cities, posted between December 2015 and April 2016. Following ethical guidelines outlined by \citet{krotov2020tutorial}, which presents a framework of seven principles for responsible web scraping, we ensured our approach complied with these standards. The Backpage website's use policy does not explicitly prohibit data scraping.

\subsection{Privacy Considerations and Potential Risks} 
\label{sec:risks}
In undertaking this research, we recognize the significant privacy concerns associated with using data from escort advertisements, particularly given that individuals within these ads may be at risk. However, the prevalence of human trafficking, a grave societal issue that affects countless lives, drives our commitment to contribute positively to anti-trafficking efforts. We believe our intentions align with the broader ethical imperative to support the fight against exploitation and to aid LEA in identifying and disrupting trafficking networks.

\hspace{\parindent} To address privacy concerns, we have extensively tried to mask personal identifiers within the dataset. Following methods from \citet{saxena-etal-2023-idtraffickers}, we mask phone numbers, email addresses, post IDs, dates, and links in text data, transforming them into generalized formats such as "<EMAILID-23>" or "<LINK>," which minimizes the risk of reverse engineering and personal identification (please refer appendix section \ref{app:dataset} for more details). At the same time, we explored various entity recognition tools to mask names \citep{li-etal-2022-extracting, liu-etal-2023-sweet} and locations, the inherent noise in the data led to inaccuracies, with some false positives in entity predictions. Since research indicates that individuals in these ads often use pseudonyms \citep{CARTER2021100065, GraulichSexTrafficking} and Backpage ads are no longer publicly accessible after the 2016 seizure, we find it unlikely that masked text data could be misused for individual identification. 

\hspace{\parindent} Privacy risks are more challenging to mitigate for the image data, as AA relies on preserving stylistic cues. Although we initially considered blurring faces to protect identities, we ultimately decided against it to avoid introducing biases that could compromise the authenticity of stylometric patterns. This decision was made after careful consideration of the potential impact on the accuracy and integrity of the AA task. Many ads already feature images with blurred or cropped faces, which suggests an attempt by individuals to maintain anonymity. For similar reasons, we also opted not to use other image augmentations, such as flipping or rotating, as these transformations could alter stylistic features tied to individual vendors, thus potentially impacting the accuracy and integrity of the AA task.

\hspace{\parindent} Our efforts to balance privacy with societal benefit align with the principles outlined in Article 6 of the General Data Protection Regulation (GDPR), \href{https://gdpr-info.eu/art-6-gdpr/}{the lawfulness of processing}. By minimizing identifiable information and rigorously managing data access, we strive to uphold this balance.

\hspace{\parindent} To further safeguard against misuse, we have established strict access controls for the MATCHED dataset. Access will be limited to vetted researchers and organizations with legitimate research goals, particularly those focused on anti-trafficking and public welfare. Each access request will undergo a thorough review by an ethics review board, assessing the legitimacy of the research goals and the adequacy of the applicant's security measures. This process ensures that only those committed to ethical and secure usage standards gain access. Applicants must also sign non-disclosure and data protection agreements legally binding them to these standards. Any violation of these guidelines will result in legal consequences. Only metadata on the \href{https://doi.org/10.34894/UR3RVE}{DataverseNL} platform will offer a high-level overview without compromising sensitive information. 

\noindent \textbf{Note:} Our research has undergone ethical scrutiny within our institution, and we have received internal approval to proceed with the project. The ethical review details and additional documentation will be provided in the camera-ready version of this paper, demonstrating our commitment to transparency and responsibility in our efforts. We are guided by the principle that our work should ultimately serve to protect and support vulnerable individuals, advancing a cause deeply rooted in societal benefit.

\subsection{Legal Impact} 
We acknowledge that the specific impact of our research on law enforcement processes is difficult to predict. Our primary goal is to support LEAs in better understanding vendor connections within online escort markets, offering a tool to assist in their investigative efforts. We strongly recommend that LEAs and researchers treat our analysis as an investigative aid rather than direct evidence for criminal prosecution. Our findings should be supplementary tools to guide investigations, not standalone proof of criminal activity.

\subsection{Environmental Impact} 
Our experiments are conducted on a private infrastructure equipped with an NVIDIA H100 80GB GPU (TDP of 350W) and a carbon efficiency of 0.475 kgCO$_2$eq/kWh. Establishing all baselines required a cumulative training time of 45.79 hours. Using the \href{https://mlco2.github.io/impact#compute}{Machine Learning Impact calculator} from \citet{lacoste2019quantifying}, we estimate the total emissions for these experiments to be approximately 16.625 kgCO$_2$eq.

\bibliography{MATCHED}

\begin{thebibliography}{92}
\providecommand{\natexlab}[1]{#1}

\bibitem[{Ai et~al.(2022)Ai, Wang, Tan, and Tan}]{ai-etal-2022-whodunit}
Bo~Ai, Yuchen Wang, Yugin Tan, and Samson Tan. 2022.
\newblock \href {https://aclanthology.org/2022.aacl-main.84} {Whodunit? learning to contrast for authorship attribution}.
\newblock In \emph{Proceedings of the 2nd Conference of the Asia-Pacific Chapter of the Association for Computational Linguistics and the 12th International Joint Conference on Natural Language Processing (Volume 1: Long Papers)}, pages 1142--1157, Online only. Association for Computational Linguistics.

\bibitem[{Alansari et~al.(2023)Alansari, Hay, Javed, Shoufan, Zweiri, and Werghi}]{10098610}
Mohamad Alansari, Oussama~Abdul Hay, Sajid Javed, Abdulhadi Shoufan, Yahya Zweiri, and Naoufel Werghi. 2023.
\newblock \href {https://doi.org/10.1109/ACCESS.2023.3266068} {Ghostfacenets: Lightweight face recognition model from cheap operations}.
\newblock \emph{IEEE Access}, 11:35429--35446.

\bibitem[{Alvari et~al.(2017)Alvari, Shakarian, and Snyder}]{alvari2017semisupervisedlearningdetectinghuman}
Hamidreza Alvari, Paulo Shakarian, and J.~E.~Kelly Snyder. 2017.
\newblock \href {https://arxiv.org/abs/1705.10786} {Semi-supervised learning for detecting human trafficking}.
\newblock \emph{Preprint}, arXiv:1705.10786.

\bibitem[{Alvari et~al.(2016)Alvari, Shakarian, and Snyder}]{7745456}
Hamidreza Alvari, Paulo Shakarian, and J.E.~Kelly Snyder. 2016.
\newblock \href {https://doi.org/10.1109/ISI.2016.7745456} {A non-parametric learning approach to identify online human trafficking}.
\newblock In \emph{2016 IEEE Conference on Intelligence and Security Informatics (ISI)}, pages 133--138.

\bibitem[{Ardakani(2020)}]{ardakani2020identifying}
Hassan~Marzoughi Ardakani. 2020.
\newblock \emph{Identifying Human Trafficking Networks in Louisiana by Using Authorship Attribution and Network Modeling}.
\newblock Louisiana State University and Agricultural \& Mechanical College.

\bibitem[{Bhargava et~al.(2013)Bhargava, Mehndiratta, and Asawa}]{bhargava2013stylometric}
Mudit Bhargava, Pulkit Mehndiratta, and Krishna Asawa. 2013.
\newblock Stylometric analysis for authorship attribution on twitter.
\newblock In \emph{Big Data Analytics: Second International Conference, BDA 2013, Mysore, India, December 16-18, 2013, Proceedings 2}, pages 37--47. Springer.

\bibitem[{Cao et~al.(2018)Cao, Shen, Xie, Parkhi, and Zisserman}]{cao2018vggface2datasetrecognisingfaces}
Qiong Cao, Li~Shen, Weidi Xie, Omkar~M. Parkhi, and Andrew Zisserman. 2018.
\newblock \href {https://arxiv.org/abs/1710.08092} {Vggface2: A dataset for recognising faces across pose and age}.
\newblock \emph{Preprint}, arXiv:1710.08092.

\bibitem[{Carter et~al.(2021)Carter, Gee, McIlhone, Lally, and Lawson}]{CARTER2021100065}
Pelham Carter, Matt Gee, Hollie McIlhone, Harkeeret Lally, and Robert Lawson. 2021.
\newblock \href {https://doi.org/10.1016/j.metip.2021.100065} {Comparing manual and computational approaches to theme identification in online forums: A case study of a sex work special interest community}.
\newblock \emph{Methods in Psychology}, 5:100065.

\bibitem[{Chambers et~al.(2019)Chambers, Forman, Griswold, Lu, Khastgir, and Steckler}]{chambers-etal-2019-character}
Nathanael Chambers, Timothy Forman, Catherine Griswold, Kevin Lu, Yogaish Khastgir, and Stephen Steckler. 2019.
\newblock \href {https://doi.org/10.18653/v1/D19-5507} {Character-based models for adversarial phone extraction: Preventing human sex trafficking}.
\newblock In \emph{Proceedings of the 5th Workshop on Noisy User-generated Text (W-NUT 2019)}, pages 48--56, Hong Kong, China. Association for Computational Linguistics.

\bibitem[{Chen et~al.(2020)Chen, Kornblith, Norouzi, and Hinton}]{chen2020simpleframeworkcontrastivelearning}
Ting Chen, Simon Kornblith, Mohammad Norouzi, and Geoffrey Hinton. 2020.
\newblock \href {https://arxiv.org/abs/2002.05709} {A simple framework for contrastive learning of visual representations}.
\newblock \emph{Preprint}, arXiv:2002.05709.

\bibitem[{Choshen et~al.(2019)Choshen, Eldad, Hershcovich, Sulem, and Abend}]{choshen-etal-2019-language}
Leshem Choshen, Dan Eldad, Daniel Hershcovich, Elior Sulem, and Omri Abend. 2019.
\newblock \href {https://doi.org/10.18653/v1/P19-1419} {The language of legal and illegal activity on the {D}arknet}.
\newblock In \emph{Proceedings of the 57th Annual Meeting of the Association for Computational Linguistics}, pages 4271--4279, Florence, Italy. Association for Computational Linguistics.

\bibitem[{Cotogni et~al.(2024)Cotogni, Arazzi, and Cusano}]{10546301}
Marco Cotogni, Marco Arazzi, and Claudio Cusano. 2024.
\newblock \href {https://doi.org/10.1109/TMM.2024.3408683} {Photostyle60: A photographic style dataset for photo authorship attribution and photographic style transfer}.
\newblock \emph{IEEE Transactions on Multimedia}, pages 1--12.

\bibitem[{Das and Rad(2020)}]{das2020opportunities}
Arun Das and Paul Rad. 2020.
\newblock \href {https://arxiv.org/abs/2006.11371} {Opportunities and challenges in explainable artificial intelligence (xai): A survey}.
\newblock \emph{Preprint}, arXiv:2006.11371.

\bibitem[{Dosovitskiy et~al.(2021)Dosovitskiy, Beyer, Kolesnikov, Weissenborn, Zhai, Unterthiner, Dehghani, Minderer, Heigold, Gelly, Uszkoreit, and Houlsby}]{dosovitskiy2021imageworth16x16words}
Alexey Dosovitskiy, Lucas Beyer, Alexander Kolesnikov, Dirk Weissenborn, Xiaohua Zhai, Thomas Unterthiner, Mostafa Dehghani, Matthias Minderer, Georg Heigold, Sylvain Gelly, Jakob Uszkoreit, and Neil Houlsby. 2021.
\newblock \href {https://arxiv.org/abs/2010.11929} {An image is worth 16x16 words: Transformers for image recognition at scale}.
\newblock \emph{Preprint}, arXiv:2010.11929.

\bibitem[{El-Nouby et~al.(2021)El-Nouby, Neverova, Laptev, and Jégou}]{elnouby2021trainingvisiontransformersimage}
Alaaeldin El-Nouby, Natalia Neverova, Ivan Laptev, and Hervé Jégou. 2021.
\newblock \href {https://arxiv.org/abs/2102.05644} {Training vision transformers for image retrieval}.
\newblock \emph{Preprint}, arXiv:2102.05644.

\bibitem[{EUROPOL(2020)}]{Europol}
EUROPOL. 2020.
\newblock \href {https://www.europol.europa.eu/publications-events/publications/challenges-of-countering-human-trafficking-in-digital-era} {The challenges of countering human trafficking in the digital era}.

\bibitem[{Fabien et~al.(2020)Fabien, Villatoro-Tello, Motlicek, and Parida}]{fabien-etal-2020-bertaa}
Ma{\"e}l Fabien, Esau Villatoro-Tello, Petr Motlicek, and Shantipriya Parida. 2020.
\newblock \href {https://aclanthology.org/2020.icon-main.16} {{B}ert{AA} : {BERT} fine-tuning for authorship attribution}.
\newblock In \emph{Proceedings of the 17th International Conference on Natural Language Processing (ICON)}, pages 127--137, Indian Institute of Technology Patna, Patna, India. NLP Association of India (NLPAI).

\bibitem[{Falcon and {The PyTorch Lightning team}(2019)}]{Falcon_PyTorch_Lightning_2019}
William Falcon and {The PyTorch Lightning team}. 2019.
\newblock \href {https://doi.org/10.5281/zenodo.3828935} {{PyTorch Lightning}}.

\bibitem[{Firmansyah et~al.(2023)Firmansyah, Kusumasari, and Alam}]{10127799}
Andrian Firmansyah, Tien~Fabrianti Kusumasari, and Ekky~Novriza Alam. 2023.
\newblock \href {https://doi.org/10.1109/ICCoSITE57641.2023.10127799} {Comparison of face recognition accuracy of arcface, facenet and facenet512 models on deepface framework}.
\newblock In \emph{2023 International Conference on Computer Science, Information Technology and Engineering (ICCoSITE)}, pages 535--539.

\bibitem[{Fobbe(2021)}]{Fobbe2021TextLinguisticAI}
Eilika Fobbe. 2021.
\newblock Text-linguistic analysis in forensic authorship attribution.

\bibitem[{Gallo et~al.(2018)Gallo, Calefati, Nawaz, and Janjua}]{8615789}
I.~Gallo, A.~Calefati, S.~Nawaz, and M.~K. Janjua. 2018.
\newblock \href {https://doi.org/10.1109/DICTA.2018.8615789} {Image and encoded text fusion for multi-modal classification}.
\newblock In \emph{2018 Digital Image Computing: Techniques and Applications (DICTA)}, pages 1--7.

\bibitem[{Gan et~al.(2024)Gan, Fu, Feng, Zhu, Cao, and Zhu}]{GAN2024122731}
Chenquan Gan, Xiang Fu, Qingdong Feng, Qingyi Zhu, Yang Cao, and Ye~Zhu. 2024.
\newblock \href {https://doi.org/10.1016/j.eswa.2023.122731} {A multimodal fusion network with attention mechanisms for visual–textual sentiment analysis}.
\newblock \emph{Expert Systems with Applications}, 242:122731.

\bibitem[{Gao et~al.(2021)Gao, Zhang, Han, and Callan}]{gao-etal-2021-scaling}
Luyu Gao, Yunyi Zhang, Jiawei Han, and Jamie Callan. 2021.
\newblock \href {https://doi.org/10.18653/v1/2021.repl4nlp-1.31} {Scaling deep contrastive learning batch size under memory limited setup}.
\newblock In \emph{Proceedings of the 6th Workshop on Representation Learning for NLP (RepL4NLP-2021)}, pages 316--321, Online. Association for Computational Linguistics.

\bibitem[{Gebru et~al.(2021)Gebru, Morgenstern, Vecchione, Vaughan, Wallach, au2, and Crawford}]{gebru2021datasheetsdatasets}
Timnit Gebru, Jamie Morgenstern, Briana Vecchione, Jennifer~Wortman Vaughan, Hanna Wallach, Hal Daumé~III au2, and Kate Crawford. 2021.
\newblock \href {https://arxiv.org/abs/1803.09010} {Datasheets for datasets}.
\newblock \emph{Preprint}, arXiv:1803.09010.

\bibitem[{Giorgi et~al.(2021)Giorgi, Nitski, Wang, and Bader}]{giorgi-etal-2021-declutr}
John Giorgi, Osvald Nitski, Bo~Wang, and Gary Bader. 2021.
\newblock \href {https://doi.org/10.18653/v1/2021.acl-long.72} {{D}e{CLUTR}: Deep contrastive learning for unsupervised textual representations}.
\newblock In \emph{Proceedings of the 59th Annual Meeting of the Association for Computational Linguistics and the 11th International Joint Conference on Natural Language Processing (Volume 1: Long Papers)}, pages 879--895, Online. Association for Computational Linguistics.

\bibitem[{Gkelios et~al.(2021)Gkelios, Boutalis, and Chatzichristofis}]{gkelios2021investigatingvisiontransformermodel}
Socratis Gkelios, Yiannis Boutalis, and Savvas~A. Chatzichristofis. 2021.
\newblock \href {https://arxiv.org/abs/2101.03771} {Investigating the vision transformer model for image retrieval tasks}.
\newblock \emph{Preprint}, arXiv:2101.03771.

\bibitem[{Hagberg et~al.(2008)Hagberg, Schult, and Swart}]{SciPyProceedings_11}
Aric~A. Hagberg, Daniel~A. Schult, and Pieter~J. Swart. 2008.
\newblock Exploring network structure, dynamics, and function using networkx.
\newblock In \emph{Proceedings of the 7th Python in Science Conference}, pages 11 -- 15, Pasadena, CA USA.

\bibitem[{He et~al.(2015)He, Zhang, Ren, and Sun}]{he2015deepresiduallearningimage}
Kaiming He, Xiangyu Zhang, Shaoqing Ren, and Jian Sun. 2015.
\newblock \href {https://arxiv.org/abs/1512.03385} {Deep residual learning for image recognition}.
\newblock \emph{Preprint}, arXiv:1512.03385.

\bibitem[{Hu et~al.(2020)Hu, Lee, Wang, Lim, and Dai}]{hu2020deepstyle}
Zhiqiang Hu, Roy Ka-Wei Lee, Lei Wang, Ee-peng Lim, and Bo~Dai. 2020.
\newblock Deepstyle: User style embedding for authorship attribution of short texts.
\newblock In \emph{Web and Big Data: 4th International Joint Conference, APWeb-WAIM 2020, Tianjin, China, September 18-20, 2020, Proceedings, Part II 4}, pages 221--229. Springer.

\bibitem[{Huang et~al.(2018)Huang, Liu, van~der Maaten, and Weinberger}]{huang2018denselyconnectedconvolutionalnetworks}
Gao Huang, Zhuang Liu, Laurens van~der Maaten, and Kilian~Q. Weinberger. 2018.
\newblock \href {https://arxiv.org/abs/1608.06993} {Densely connected convolutional networks}.
\newblock \emph{Preprint}, arXiv:1608.06993.

\bibitem[{Huertas-Tato et~al.(2024)Huertas-Tato, Mart{\'\i}n, and Camacho}]{huertas2024understanding}
Javier Huertas-Tato, Alejandro Mart{\'\i}n, and David Camacho. 2024.
\newblock Understanding writing style in social media with a supervised contrastively pre-trained transformer.
\newblock \emph{Knowledge-Based Systems}, 296:111867.

\bibitem[{Hunter(2007)}]{4160265}
John~D. Hunter. 2007.
\newblock \href {https://doi.org/10.1109/MCSE.2007.55} {Matplotlib: A 2d graphics environment}.
\newblock \emph{Computing in Science \& Engineering}, 9(3):90--95.

\bibitem[{Ibanez and Gazan(2016)}]{7752332}
Michelle Ibanez and Rich Gazan. 2016.
\newblock \href {https://doi.org/10.1109/ASONAM.2016.7752332} {Virtual indicators of sex trafficking to identify potential victims in online advertisements}.
\newblock In \emph{2016 IEEE/ACM International Conference on Advances in Social Networks Analysis and Mining (ASONAM)}, pages 818--824.

\bibitem[{Ibanez and Suthers(2014)}]{6758797}
Michelle Ibanez and Daniel~D. Suthers. 2014.
\newblock \href {https://doi.org/10.1109/HICSS.2014.200} {Detection of domestic human trafficking indicators and movement trends using content available on open internet sources}.
\newblock In \emph{2014 47th Hawaii International Conference on System Sciences}, pages 1556--1565.

\bibitem[{ILO(2012)}]{ILO}
ILO. 2012.
\newblock \href {http://www.ilo.org/wcmsp5/groups/public/---ed_norm/---declaration/documents/publication/wcms_182004.pdf} {Ilo global estimate of forced labour}.

\bibitem[{Inc.(2015)}]{plotly}
Plotly~Technologies Inc. 2015.
\newblock \href {https://plot.ly} {Collaborative data science}.

\bibitem[{Iqbal et~al.(2008)Iqbal, Hadjidj, Fung, and Debbabi}]{IQBAL2008S42}
Farkhund Iqbal, Rachid Hadjidj, Benjamin~C.M. Fung, and Mourad Debbabi. 2008.
\newblock \href {https://doi.org/10.1016/j.diin.2008.05.001} {A novel approach of mining write-prints for authorship attribution in e-mail forensics}.
\newblock \emph{Digital Investigation}, 5:S42--S51.
\newblock The Proceedings of the Eighth Annual DFRWS Conference.

\bibitem[{Johnson et~al.(2019)Johnson, Douze, and J{\'e}gou}]{johnson2019billion}
Jeff Johnson, Matthijs Douze, and Herv{\'e} J{\'e}gou. 2019.
\newblock Billion-scale similarity search with {GPUs}.
\newblock \emph{IEEE Transactions on Big Data}, 7(3):535--547.

\bibitem[{Juola and Baayen(2005)}]{juola2005controlled}
Patrick Juola and R~Harald Baayen. 2005.
\newblock A controlled-corpus experiment in authorship identification by cross-entropy.
\newblock \emph{Literary and Linguistic Computing}, 20(Suppl):59--67.

\bibitem[{Karkkainen and Joo(2021)}]{karkkainenfairface}
Kimmo Karkkainen and Jungseock Joo. 2021.
\newblock Fairface: Face attribute dataset for balanced race, gender, and age for bias measurement and mitigation.
\newblock In \emph{Proceedings of the IEEE/CVF Winter Conference on Applications of Computer Vision}, pages 1548--1558.

\bibitem[{Kaya and Bilge(2019)}]{kaya2019deep}
Mahmut Kaya and Hasan~{\c{S}}akir Bilge. 2019.
\newblock Deep metric learning: A survey.
\newblock \emph{Symmetry}, 11(9):1066.

\bibitem[{Kiela et~al.(2020)Kiela, Bhooshan, Firooz, Perez, and Testuggine}]{kiela2020supervisedmultimodalbitransformersclassifying}
Douwe Kiela, Suvrat Bhooshan, Hamed Firooz, Ethan Perez, and Davide Testuggine. 2020.
\newblock \href {https://arxiv.org/abs/1909.02950} {Supervised multimodal bitransformers for classifying images and text}.
\newblock \emph{Preprint}, arXiv:1909.02950.

\bibitem[{Kim et~al.(2021)Kim, Son, and Kim}]{kim2021viltvisionandlanguagetransformerconvolution}
Wonjae Kim, Bokyung Son, and Ildoo Kim. 2021.
\newblock \href {https://arxiv.org/abs/2102.03334} {Vilt: Vision-and-language transformer without convolution or region supervision}.
\newblock \emph{Preprint}, arXiv:2102.03334.

\bibitem[{Kokhlikyan et~al.(2020)Kokhlikyan, Miglani, Martin, Wang, Alsallakh, Reynolds, Melnikov, Kliushkina, Araya, Yan, and Reblitz-Richardson}]{kokhlikyan2020captumunifiedgenericmodel}
Narine Kokhlikyan, Vivek Miglani, Miguel Martin, Edward Wang, Bilal Alsallakh, Jonathan Reynolds, Alexander Melnikov, Natalia Kliushkina, Carlos Araya, Siqi Yan, and Orion Reblitz-Richardson. 2020.
\newblock \href {https://arxiv.org/abs/2009.07896} {Captum: A unified and generic model interpretability library for pytorch}.
\newblock \emph{Preprint}, arXiv:2009.07896.

\bibitem[{Kornblith et~al.(2019)Kornblith, Norouzi, Lee, and Hinton}]{kornblith2019similarityneuralnetworkrepresentations}
Simon Kornblith, Mohammad Norouzi, Honglak Lee, and Geoffrey Hinton. 2019.
\newblock \href {https://arxiv.org/abs/1905.00414} {Similarity of neural network representations revisited}.
\newblock \emph{Preprint}, arXiv:1905.00414.

\bibitem[{Krishna et~al.(2022)Krishna, Han, Gu, Pombra, Jabbari, Wu, and Lakkaraju}]{krishna2022disagreement}
Satyapriya Krishna, Tessa Han, Alex Gu, Javin Pombra, Shahin Jabbari, Steven Wu, and Himabindu Lakkaraju. 2022.
\newblock \href {https://arxiv.org/abs/2202.01602} {The disagreement problem in explainable machine learning: A practitioner's perspective}.
\newblock \emph{Preprint}, arXiv:2202.01602.

\bibitem[{Krotov et~al.(2020)Krotov, Johnson, and Silva}]{krotov2020tutorial}
Vlad Krotov, Leigh Johnson, and Leiser Silva. 2020.
\newblock Tutorial: Legality and ethics of web scraping.

\bibitem[{Lacoste et~al.(2019)Lacoste, Luccioni, Schmidt, and Dandres}]{lacoste2019quantifying}
Alexandre Lacoste, Alexandra Luccioni, Victor Schmidt, and Thomas Dandres. 2019.
\newblock Quantifying the carbon emissions of machine learning.
\newblock \emph{arXiv preprint arXiv:1910.09700}.

\bibitem[{Li et~al.(2023)Li, Li, Savarese, and Hoi}]{li2023blip2bootstrappinglanguageimagepretraining}
Junnan Li, Dongxu Li, Silvio Savarese, and Steven Hoi. 2023.
\newblock \href {https://arxiv.org/abs/2301.12597} {Blip-2: Bootstrapping language-image pre-training with frozen image encoders and large language models}.
\newblock \emph{Preprint}, arXiv:2301.12597.

\bibitem[{Li et~al.(2019)Li, Yatskar, Yin, Hsieh, and Chang}]{li2019visualbertsimpleperformantbaseline}
Liunian~Harold Li, Mark Yatskar, Da~Yin, Cho-Jui Hsieh, and Kai-Wei Chang. 2019.
\newblock \href {https://arxiv.org/abs/1908.03557} {Visualbert: A simple and performant baseline for vision and language}.
\newblock \emph{Preprint}, arXiv:1908.03557.

\bibitem[{Li et~al.(2022{\natexlab{a}})Li, Nair, Pelrine, and Rabbany}]{li-etal-2022-extracting}
Yifei Li, Pratheeksha Nair, Kellin Pelrine, and Reihaneh Rabbany. 2022{\natexlab{a}}.
\newblock \href {https://doi.org/10.18653/v1/2022.findings-acl.225} {Extracting person names from user generated text: Named-entity recognition for combating human trafficking}.
\newblock In \emph{Findings of the Association for Computational Linguistics: ACL 2022}, pages 2854--2868, Dublin, Ireland. Association for Computational Linguistics.

\bibitem[{Li et~al.(2022{\natexlab{b}})Li, Nair, Pelrine, and Rabbany}]{li-etal-2022-extracting-person-name}
Yifei Li, Pratheeksha Nair, Kellin Pelrine, and Reihaneh Rabbany. 2022{\natexlab{b}}.
\newblock \href {https://doi.org/10.18653/v1/2022.findings-acl.225} {Extracting person names from user generated text: Named-entity recognition for combating human trafficking}.
\newblock In \emph{Findings of the Association for Computational Linguistics: ACL 2022}, pages 2854--2868, Dublin, Ireland. Association for Computational Linguistics.

\bibitem[{Li et~al.(2024)Li, Zheng, Zhu, Mei, Chen, and Tao}]{li2024compact}
Yushi Li, Xin Zheng, Ming Zhu, Jie Mei, Ziwen Chen, and Yunfei Tao. 2024.
\newblock Compact bilinear pooling and multi-loss network for social media multimodal classification.
\newblock \emph{Signal, Image and Video Processing}, 18(11):8403--8412.

\bibitem[{Liu et~al.(2023)Liu, Yu, Sujaya, Nair, Pelrine, and Rabbany}]{liu-etal-2023-sweet}
Javin Liu, Hao Yu, Vidya Sujaya, Pratheeksha Nair, Kellin Pelrine, and Reihaneh Rabbany. 2023.
\newblock \href {https://doi.org/10.18653/v1/2023.findings-emnlp.219} {{SWEET} - weakly supervised person name extraction for fighting human trafficking}.
\newblock In \emph{Findings of the Association for Computational Linguistics: EMNLP 2023}, pages 3355--3367, Singapore. Association for Computational Linguistics.

\bibitem[{Lugo-Graulich()}]{GraulichSexTrafficking}
Kristina Lugo-Graulich.
\newblock Indicators of sex trafficking in online escort ads.
\newblock \url{https://www.ojp.gov/pdffiles1/nij/grants/305453.pdf}.

\bibitem[{Lugo-Graulich and Meyer(2021)}]{jrsa}
Kristina Lugo-Graulich and Leah~F. Meyer. 2021.
\newblock \href {https://www.jrsa.org/projects/escorts-resources/jrsa-le-guide-on-indicators-of-sex-trafficking-in-online-escort-ads.pdf} {Law enforcement guide on indicators of sex trafficking in online escort ads}.
\newblock \emph{Justice Research and Statistics Association}.

\bibitem[{Lundberg and Lee(2017)}]{lundberg2017unifiedapproachinterpretingmodel}
Scott Lundberg and Su-In Lee. 2017.
\newblock \href {https://arxiv.org/abs/1705.07874} {A unified approach to interpreting model predictions}.
\newblock \emph{Preprint}, arXiv:1705.07874.

\bibitem[{Manolache et~al.(2022)Manolache, Brad, Barbalau, Ionescu, and Popescu}]{manolache2022veridarklargescalebenchmarkauthorship}
Andrei Manolache, Florin Brad, Antonio Barbalau, Radu~Tudor Ionescu, and Marius Popescu. 2022.
\newblock \href {https://arxiv.org/abs/2207.03477} {Veridark: A large-scale benchmark for authorship verification on the dark web}.
\newblock \emph{Preprint}, arXiv:2207.03477.

\bibitem[{Nagpal et~al.(2017)Nagpal, Miller, Boecking, and Dubrawski}]{nagpal2017entity}
Chirag Nagpal, Kyle Miller, Benedikt Boecking, and Artur Dubrawski. 2017.
\newblock \href {https://arxiv.org/abs/1509.06659} {An entity resolution approach to isolate instances of human trafficking online}.
\newblock \emph{Preprint}, arXiv:1509.06659.

\bibitem[{Nagpal et~al.(2015)Nagpal, Miller, Boecking, and Dubrawski}]{Nagpal2015AnER}
Chirag Nagpal, Kyle Miller, Benedikt Boecking, and Artur~W. Dubrawski. 2015.
\newblock \href {https://api.semanticscholar.org/CorpusID:11047665} {An entity resolution approach to isolate instances of human trafficking online}.
\newblock In \emph{NUT@EMNLP}.

\bibitem[{Nair et~al.(2022)Nair, Li, Vajiac, Olligschlaeger, Lee, Park, Chau, Faloutsos, and Rabbany}]{10.1145/3487553.3524263}
Pratheeksha Nair, Yifei Li, Catalina Vajiac, Andreas Olligschlaeger, Meng-Chieh Lee, Namyong Park, Duen~Horng Chau, Christos Faloutsos, and Reihaneh Rabbany. 2022.
\newblock \href {https://doi.org/10.1145/3487553.3524263} {Vispad: Visualization and pattern discovery for fighting human trafficking}.
\newblock In \emph{Companion Proceedings of the Web Conference 2022}, WWW '22, page 273–277, New York, NY, USA. Association for Computing Machinery.

\bibitem[{Nirkhi and Dharaskar(2013)}]{Nirkhi2013}
Smita Nirkhi and Dr.~R.V. Dharaskar. 2013.
\newblock \href {https://doi.org/10.14569/IJACSA.2013.040505} {Comparative study of authorship identification techniques for cyber forensics analysis}.
\newblock \emph{International Journal of Advanced Computer Science and Applications}, 4(5).

\bibitem[{Paszke et~al.(2019)Paszke, Gross, Massa, Lerer, Bradbury, Chanan, Killeen, Lin, Gimelshein, Antiga, Desmaison, Kopf, Yang, DeVito, Raison, Tejani, Chilamkurthy, Steiner, Fang, Bai, and Chintala}]{NEURIPS2019_bdbca288}
Adam Paszke, Sam Gross, Francisco Massa, Adam Lerer, James Bradbury, Gregory Chanan, Trevor Killeen, Zeming Lin, Natalia Gimelshein, Luca Antiga, Alban Desmaison, Andreas Kopf, Edward Yang, Zachary DeVito, Martin Raison, Alykhan Tejani, Sasank Chilamkurthy, Benoit Steiner, Lu~Fang, Junjie Bai, and Soumith Chintala. 2019.
\newblock \href {https://proceedings.neurips.cc/paper_files/paper/2019/file/bdbca288fee7f92f2bfa9f7012727740-Paper.pdf} {Pytorch: An imperative style, high-performance deep learning library}.
\newblock In \emph{Advances in Neural Information Processing Systems}, volume~32. Curran Associates, Inc.

\bibitem[{Pedregosa et~al.(2011)Pedregosa, Varoquaux, Gramfort, Michel, Thirion, Grisel, Blondel, Prettenhofer, Weiss, Dubourg et~al.}]{pedregosa2011scikit}
Fabian Pedregosa, Ga{\"e}l Varoquaux, Alexandre Gramfort, Vincent Michel, Bertrand Thirion, Olivier Grisel, Mathieu Blondel, Peter Prettenhofer, Ron Weiss, Vincent Dubourg, et~al. 2011.
\newblock Scikit-learn: Machine learning in python.
\newblock \emph{Journal of machine learning research}, 12(Oct):2825--2830.

\bibitem[{POLARIS(2018)}]{Polaris_stats2018}
POLARIS. 2018.
\newblock \href {https://wiki.preventconnect.org/wp-content/uploads/2018/08/Human-Trafficking-Statistics-from-the-Polaris-Project.pdf} {Human trafficking statistics}.

\bibitem[{POLARIS(2020)}]{Polaris_stats2020}
POLARIS. 2020.
\newblock \href {https://polarisproject.org/wp-content/uploads/2022/01/Polaris-Analysis-of-2020-Data-from-the-National-Human-Trafficking-Hotline.pdf} {Polaris analysis of 2020 data from the national human trafficking hotline}.

\bibitem[{Portnoff et~al.(2017)Portnoff, Huang, Doerfler, Afroz, and McCoy}]{10.1145/3097983.3098082}
Rebecca~S. Portnoff, Danny~Yuxing Huang, Periwinkle Doerfler, Sadia Afroz, and Damon McCoy. 2017.
\newblock \href {https://doi.org/10.1145/3097983.3098082} {Backpage and bitcoin: Uncovering human traffickers}.
\newblock In \emph{Proceedings of the 23rd ACM SIGKDD International Conference on Knowledge Discovery and Data Mining}, KDD '17, page 1595–1604, New York, NY, USA. Association for Computing Machinery.

\bibitem[{Rabbany et~al.(2018)Rabbany, Bayani, and Dubrawski}]{rabbany2018active}
Reihaneh Rabbany, David Bayani, and Artur Dubrawski. 2018.
\newblock Active search of connections for case building and combating human trafficking.
\newblock In \emph{Proceedings of the 24th ACM SIGKDD International Conference on Knowledge Discovery \& Data Mining}, pages 2120--2129.

\bibitem[{Radford et~al.(2021)Radford, Kim, Hallacy, Ramesh, Goh, Agarwal, Sastry, Askell, Mishkin, Clark, Krueger, and Sutskever}]{radford2021learningtransferablevisualmodels}
Alec Radford, Jong~Wook Kim, Chris Hallacy, Aditya Ramesh, Gabriel Goh, Sandhini Agarwal, Girish Sastry, Amanda Askell, Pamela Mishkin, Jack Clark, Gretchen Krueger, and Ilya Sutskever. 2021.
\newblock \href {https://arxiv.org/abs/2103.00020} {Learning transferable visual models from natural language supervision}.
\newblock \emph{Preprint}, arXiv:2103.00020.

\bibitem[{Ramnial et~al.(2016)Ramnial, Panchoo, and Pudaruth}]{ramnial2016authorship}
Hoshiladevi Ramnial, Shireen Panchoo, and Sameerchand Pudaruth. 2016.
\newblock Authorship attribution using stylometry and machine learning techniques.
\newblock In \emph{Intelligent Systems Technologies and Applications: Volume 1}, pages 113--125. Springer.

\bibitem[{Ribeiro et~al.(2016)Ribeiro, Singh, and Guestrin}]{ribeiro2016whyitrustyou}
Marco~Tulio Ribeiro, Sameer Singh, and Carlos Guestrin. 2016.
\newblock \href {https://arxiv.org/abs/1602.04938} {"why should i trust you?": Explaining the predictions of any classifier}.
\newblock \emph{Preprint}, arXiv:1602.04938.

\bibitem[{Sahu and Vechtomova(2021)}]{sahu-vechtomova-2021-adaptive}
Gaurav Sahu and Olga Vechtomova. 2021.
\newblock \href {https://doi.org/10.18653/v1/2021.eacl-main.275} {Adaptive fusion techniques for multimodal data}.
\newblock In \emph{Proceedings of the 16th Conference of the European Chapter of the Association for Computational Linguistics: Main Volume}, pages 3156--3166, Online. Association for Computational Linguistics.

\bibitem[{Saxena et~al.(2023{\natexlab{a}})Saxena, Ashpole, van Dijck, and Spanakis}]{saxena-etal-2023-idtraffickers}
Vageesh Saxena, Benjamin Ashpole, Gijs van Dijck, and Gerasimos Spanakis. 2023{\natexlab{a}}.
\newblock \href {https://doi.org/10.18653/v1/2023.emnlp-main.524} {{IDT}raffickers: An authorship attribution dataset to link and connect potential human-trafficking operations on text escort advertisements}.
\newblock In \emph{Proceedings of the 2023 Conference on Empirical Methods in Natural Language Processing}, pages 8444--8464, Singapore. Association for Computational Linguistics.

\bibitem[{Saxena et~al.(2023{\natexlab{b}})Saxena, Rethmeier, van Dijck, and Spanakis}]{saxena-etal-2023-vendorlink}
Vageesh Saxena, Nils Rethmeier, Gijs van Dijck, and Gerasimos Spanakis. 2023{\natexlab{b}}.
\newblock \href {https://doi.org/10.18653/v1/2023.acl-long.481} {{V}endor{L}ink: An {NLP} approach for identifying {\&} linking vendor migrants {\&} potential aliases on {D}arknet markets}.
\newblock In \emph{Proceedings of the 61st Annual Meeting of the Association for Computational Linguistics (Volume 1: Long Papers)}, pages 8619--8639, Toronto, Canada. Association for Computational Linguistics.

\bibitem[{Serengil and Ozpinar(2023)}]{serengil2023db}
Sefik~Ilkin Serengil and Alper Ozpinar. 2023.
\newblock \href {https://doi.org/10.33774/coe-2023-18rcn} {An evaluation of sql and nosql databases for facial recognition pipelines}.

\bibitem[{Simonyan and Zisserman(2015)}]{simonyan2015deepconvolutionalnetworkslargescale}
Karen Simonyan and Andrew Zisserman. 2015.
\newblock \href {https://arxiv.org/abs/1409.1556} {Very deep convolutional networks for large-scale image recognition}.
\newblock \emph{Preprint}, arXiv:1409.1556.

\bibitem[{Sleeman et~al.(2022)Sleeman, Kapoor, and Ghosh}]{10.1145/3543848}
William~C. Sleeman, Rishabh Kapoor, and Preetam Ghosh. 2022.
\newblock \href {https://doi.org/10.1145/3543848} {Multimodal classification: Current landscape, taxonomy and future directions}.
\newblock \emph{ACM Comput. Surv.}, 55(7).

\bibitem[{Striebel et~al.(2024)Striebel, Edikala, Irby, Rosenfeld, Gage, Dakota, and K{\"u}bler}]{striebel-etal-2024-scaling}
Jacob Striebel, Abishek Edikala, Ethan Irby, Alex Rosenfeld, J.~Gage, Daniel Dakota, and Sandra K{\"u}bler. 2024.
\newblock \href {https://doi.org/10.18653/v1/2024.naacl-industry.24} {Scaling up authorship attribution}.
\newblock In \emph{Proceedings of the 2024 Conference of the North American Chapter of the Association for Computational Linguistics: Human Language Technologies (Volume 6: Industry Track)}, pages 295--302, Mexico City, Mexico. Association for Computational Linguistics.

\bibitem[{Szegedy et~al.(2015)Szegedy, Vanhoucke, Ioffe, Shlens, and Wojna}]{szegedy2015rethinkinginceptionarchitecturecomputer}
Christian Szegedy, Vincent Vanhoucke, Sergey Ioffe, Jonathon Shlens, and Zbigniew Wojna. 2015.
\newblock \href {https://arxiv.org/abs/1512.00567} {Rethinking the inception architecture for computer vision}.
\newblock \emph{Preprint}, arXiv:1512.00567.

\bibitem[{Tamás et~al.(2022)Tamás, Atzenhofer-Baumgartner, Florian, Aoun, Sandy, Nicolaou, Anguelos, Luger, Daniel, Decker, Franziska, Lamminger, Florian, Vogeler, and Georg}]{ercdidip2022}
Kovács Tamás, Atzenhofer-Baumgartner, Florian, Aoun, Sandy, Nicolaou, Anguelos, Luger, Daniel, Decker, Franziska, Lamminger, Florian, Vogeler, and Georg. 2022.
\newblock \href {https://doi.org/10.57967/hf/0135} {langdetect (revision 0215f72)}.

\bibitem[{Tan and Le(2021)}]{tan2021efficientnetv2smallermodelsfaster}
Mingxing Tan and Quoc~V. Le. 2021.
\newblock \href {https://arxiv.org/abs/2104.00298} {Efficientnetv2: Smaller models and faster training}.
\newblock \emph{Preprint}, arXiv:2104.00298.

\bibitem[{Tong et~al.(2017)Tong, Zadeh, Jones, and Morency}]{tong-etal-2017-combating}
Edmund Tong, Amir Zadeh, Cara Jones, and Louis-Philippe Morency. 2017.
\newblock \href {https://doi.org/10.18653/v1/P17-1142} {Combating human trafficking with multimodal deep models}.
\newblock In \emph{Proceedings of the 55th Annual Meeting of the Association for Computational Linguistics (Volume 1: Long Papers)}, pages 1547--1556, Vancouver, Canada. Association for Computational Linguistics.

\bibitem[{Tyo et~al.(2021)Tyo, Dhingra, and Lipton}]{tyo2021siamese}
Jacob Tyo, Bhuwan Dhingra, and Zachary~C Lipton. 2021.
\newblock Siamese bert for authorship verification.
\newblock In \emph{CLEF (Working Notes)}, pages 2169--2177.

\bibitem[{UNDOC(2020)}]{Undoc}
UNDOC. 2020.
\newblock \href {https://www.unodc.org/documents/data-and-analysis/tip/2021/GLOTiP_2020_15jan_web.pdf} {Global report on trafficking in persons}.

\bibitem[{Vaessen and van Leeuwen(2024)}]{vaessen2024effectbatchsizecontrastive}
Nik Vaessen and David~A. van Leeuwen. 2024.
\newblock \href {https://arxiv.org/abs/2402.13723} {The effect of batch size on contrastive self-supervised speech representation learning}.
\newblock \emph{Preprint}, arXiv:2402.13723.

\bibitem[{Vajiac et~al.(2023)Vajiac, Chau, Olligschlaeger, Mackenzie, Nair, Lee, Li, Park, Rabbany, and Faloutsos}]{9912645}
Catalina Vajiac, Duen~Horng Chau, Andreas Olligschlaeger, Rebecca Mackenzie, Pratheeksha Nair, Meng-Chieh Lee, Yifei Li, Namyong Park, Reihaneh Rabbany, and Christos Faloutsos. 2023.
\newblock \href {https://doi.org/10.1109/TVCG.2022.3209403} {Trafficvis: Visualizing organized activity and spatio-temporal patterns for detecting and labeling human trafficking}.
\newblock \emph{IEEE Transactions on Visualization and Computer Graphics}, 29(1):53--62.

\bibitem[{Villegas et~al.(2024)Villegas, Preoţiuc-Pietro, and Aletras}]{villegas2024improvingmultimodalclassificationsocial}
Danae~Sánchez Villegas, Daniel Preoţiuc-Pietro, and Nikolaos Aletras. 2024.
\newblock \href {https://arxiv.org/abs/2309.07794} {Improving multimodal classification of social media posts by leveraging image-text auxiliary tasks}.
\newblock \emph{Preprint}, arXiv:2309.07794.

\bibitem[{Wang et~al.(2018)Wang, Peng, Wang, and Wang}]{10.1145/3196494.3196529}
Xiangwen Wang, Peng Peng, Chun Wang, and Gang Wang. 2018.
\newblock \href {https://doi.org/10.1145/3196494.3196529} {You are your photographs: Detecting multiple identities of vendors in the darknet marketplaces}.
\newblock In \emph{Proceedings of the 2018 on Asia Conference on Computer and Communications Security}, ASIACCS '18, page 431–442, New York, NY, USA. Association for Computing Machinery.

\bibitem[{Wegmann et~al.(2022)Wegmann, Schraagen, and Nguyen}]{wegmann-etal-2022-author}
Anna Wegmann, Marijn Schraagen, and Dong Nguyen. 2022.
\newblock \href {https://doi.org/10.18653/v1/2022.repl4nlp-1.26} {Same author or just same topic? towards content-independent style representations}.
\newblock In \emph{Proceedings of the 7th Workshop on Representation Learning for NLP}, pages 249--268, Dublin, Ireland. Association for Computational Linguistics.

\bibitem[{Woo et~al.(2023)Woo, Debnath, Hu, Chen, Liu, Kweon, and Xie}]{woo2023convnextv2codesigningscaling}
Sanghyun Woo, Shoubhik Debnath, Ronghang Hu, Xinlei Chen, Zhuang Liu, In~So Kweon, and Saining Xie. 2023.
\newblock \href {https://arxiv.org/abs/2301.00808} {Convnext v2: Co-designing and scaling convnets with masked autoencoders}.
\newblock \emph{Preprint}, arXiv:2301.00808.

\bibitem[{Ye et~al.(2023)Ye, Zhong, Qi, and Han}]{ye2023supervised}
Zhanhong Ye, Changle Zhong, Haoliang Qi, and Yong Han. 2023.
\newblock Supervised contrastive learning for multi-author writing style analysis.
\newblock In \emph{CLEF (Working Notes)}, pages 2817--2822.

\bibitem[{Zhang et~al.(2019)Zhang, Fan, Song, Hou, Ye, Li, Zhao, Shi, Wang, and Xiong}]{10.1145/3308558.3313537}
Yiming Zhang, Yujie Fan, Wei Song, Shifu Hou, Yanfang Ye, Xin Li, Liang Zhao, Chuan Shi, Jiabin Wang, and Qi~Xiong. 2019.
\newblock \href {https://doi.org/10.1145/3308558.3313537} {Your style your identity: Leveraging writing and photography styles for drug trafficker identification in darknet markets over attributed heterogeneous information network}.
\newblock In \emph{The World Wide Web Conference}, WWW '19, page 3448–3454, New York, NY, USA. Association for Computing Machinery.

\end{thebibliography}

\appendix

\section{Appendix}
\label{sec:appendix}

\subsection{Responsible NLP Checklist}
\label{app:checklist}

\subsubsection{For every submission}
\paragraph{Did you describe the limitations of your work?} Yes, the limitations of our work are extensively described in Section \ref{sec:limitations}.

\paragraph{Did you discuss any potential risks of your work?} Yes, the potential privacy risks associated with our work are described in Section \ref{sec:risks}.

\subsubsection{Did you use or create scientific artifacts?}

\paragraph{Did you discuss the license or terms for use and / or distribution of any artifacts?} The dataset will be released under the license version CC0 1.0. Extensive details about the terms of use and / or distribution are mentioned in Appendix Section \ref{app:dataset}.

\paragraph{Did you discuss if your use of existing artifact(s) was consistent with their intended use, provided that it was specified? For the artifacts you create, do you specify intended use and whether that is compatible with the original access conditions (in particular, derivatives of data accessed for research purposes should not be used outside of research contexts)?} Given the sensitivity of our dataset, access will be provided under restricted conditions to ensure ethical use. Interested parties must sign a Non-Disclosure Agreement (NDA) and Data Transfer Agreement (DTA) with our institution and the ethics committee. To minimize risks to individuals represented in the dataset, we have implemented strong anonymization techniques to remove private and personally identifiable information. We strictly prohibit using this dataset for any commercial or unethical purposes beyond the intended scope of our research. Violations of these guidelines will be subject to legal repercussions as outlined by the institution’s policies and the ethics committee.

\paragraph{Did you discuss the steps taken to check whether the data that was collected / used contains any information that names or uniquely identifies individual people or offensive content, and the steps taken to protect / anonymize it?} Yes, we thoroughly detail the data collection and preprocessing steps, including the measures taken to identify and remove any private or personally identifiable information. Specifically, we anonymize sensitive content such as names, phone numbers, email addresses, advertisement IDs, dates, and ages of individuals to ensure privacy. These efforts and additional discussions are comprehensively reported in Appendix Section \ref{app:dataset}-\ref{app:datasheet}.

\paragraph{Did you provide documentation of the artifacts, e.g., coverage of domains, languages, and linguistic phenomena, demographic groups represented, etc.?} Yes, the details about the coverage of domains, languages, and geographical groups are presented in Section \ref{sec:dataset} and Appendix Sections \ref{app:dataset} - \ref{app:datasheet}.

\paragraph{Did you report relevant statistics like the number of examples, details of train / test / dev splits, etc. for the data that you used / created?} Yes, these details are mentioned in Section \ref{sec:dataset} and Appendix Section \ref{app:experimental_details}. 

\subsubsection{Did you run computational experiments?}

\paragraph{Did you report the number of parameters in the models used, the total computational budget (e.g., GPU hours), and computing infrastructure used?} Yes, these details are attached in Appendix Table \ref{tab:all_vendor_identification_results}.

\paragraph{Did you discuss the experimental setup, including hyperparameter search and best-found hyperparameter values? } Yes, these details are attached in Appendix Section \ref{app:experimental_details}.

\paragraph{Did you report descriptive statistics about your results (e.g., error bars around results, summary statistics from sets of experiments), and is it transparent whether you are reporting the max, mean, etc. or just a single run?} Details about the effects of random initialization for our best-performing model, the end-to-end multimodal DeCLUTR-ViT baseline, are attached in Appendix Section \ref{app:experimental_details}.

\paragraph{If you used existing packages (e.g., for preprocessing, for normalization, or for evaluation, such as NLTK, Spacy, ROUGE, etc.), did you report the implementation, model, and parameter settings used?} All relevant details are described in Appendix Section \ref{app:experimental_details}. 

\subsubsection{Did you use human annotators (e.g., crowdworkers) or research with human participants?}

\paragraph{Do you use any human annotators?} No.

\paragraph{Did you discuss whether and how consent was obtained from people whose data you’re using/curating?} Getting consent for our data is challenging due to the nature and timeline of our dataset. We have extensively described this problem in our Appendix Section \ref{app:datasheet}.

\paragraph{Was the data collection protocol approved (or determined exempt) by an ethics review board?} Yes, the approval was granted by our institutional's ethics board. We plan to attach the approval in our camera-ready version. 

\subsubsection{Did you use AI assistants (e.g., ChatGPT, Copilot) in your research, coding, or writing?}

While our research methodology, experiments, and results were developed independently without AI assistants, we utilized ChatGPT and Grammarly to improve our paper's readability, clarity, and flow. Importantly, we wrote the initial drafts, including all content. ChatGPT was used only to paraphrase sections for clarity and improve grammar. Additionally, for coding purposes, we employed ChatGPT solely to generate in-line comments for better code readability. Specifically, we passed hand-written functions and classes to ChatGPT and requested it to generate comments without altering any logic or structure in the code.

This information is transparently described here and is not included in the main manuscript because the AI assistance was limited to minor paraphrasing, grammar improvement, and in-line code comments, with no role in generating methodology, experiments, or results.

\subsection{Dataset}
\label{app:dataset}

\begin{figure*}[ht!]
    \centering
    \begin{subfigure}[b]{\textwidth}
    \centering
    \includegraphics[width=\textwidth]{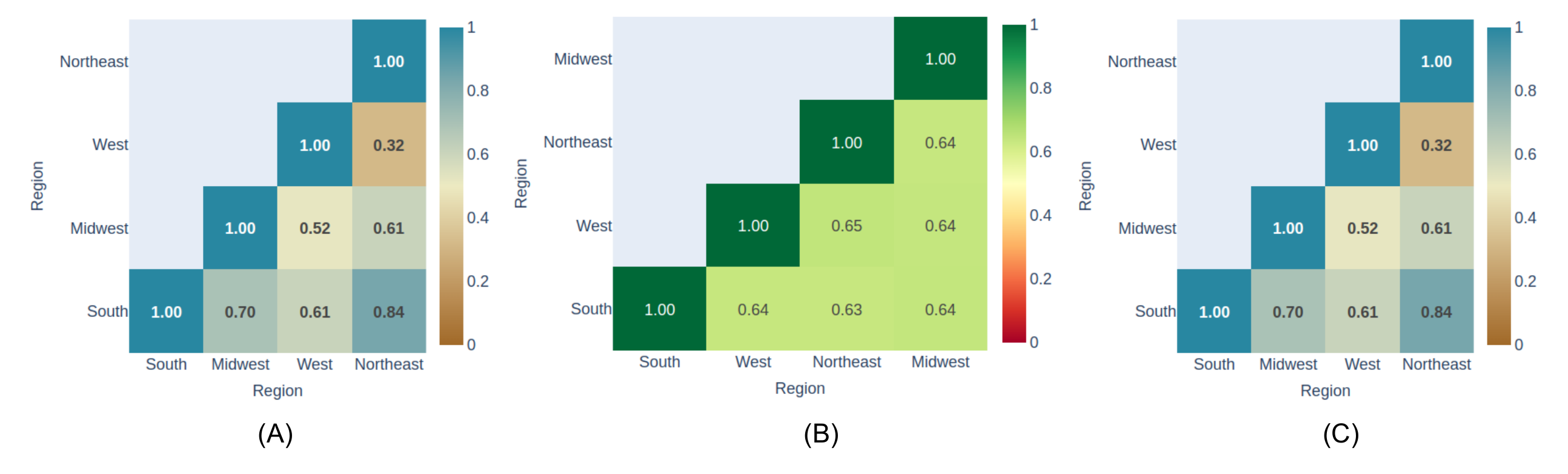}
    \caption{\label{fig:dataSimilarity} Figure (A) shows the \% of vendors shared between different datasets. Figures (B) and (C) show the average text-text and image-image cosine similarity between datasets computed on the ad representations from the pre-trained available checkpoints of DeCLUTR-small and ViT-base-patch16 backbones, respectively.}
    \end{subfigure}
    
    \vspace{0.5cm} 
    
    \begin{subfigure}[b]{\textwidth}
        \centering
        \includegraphics[width=\textwidth]{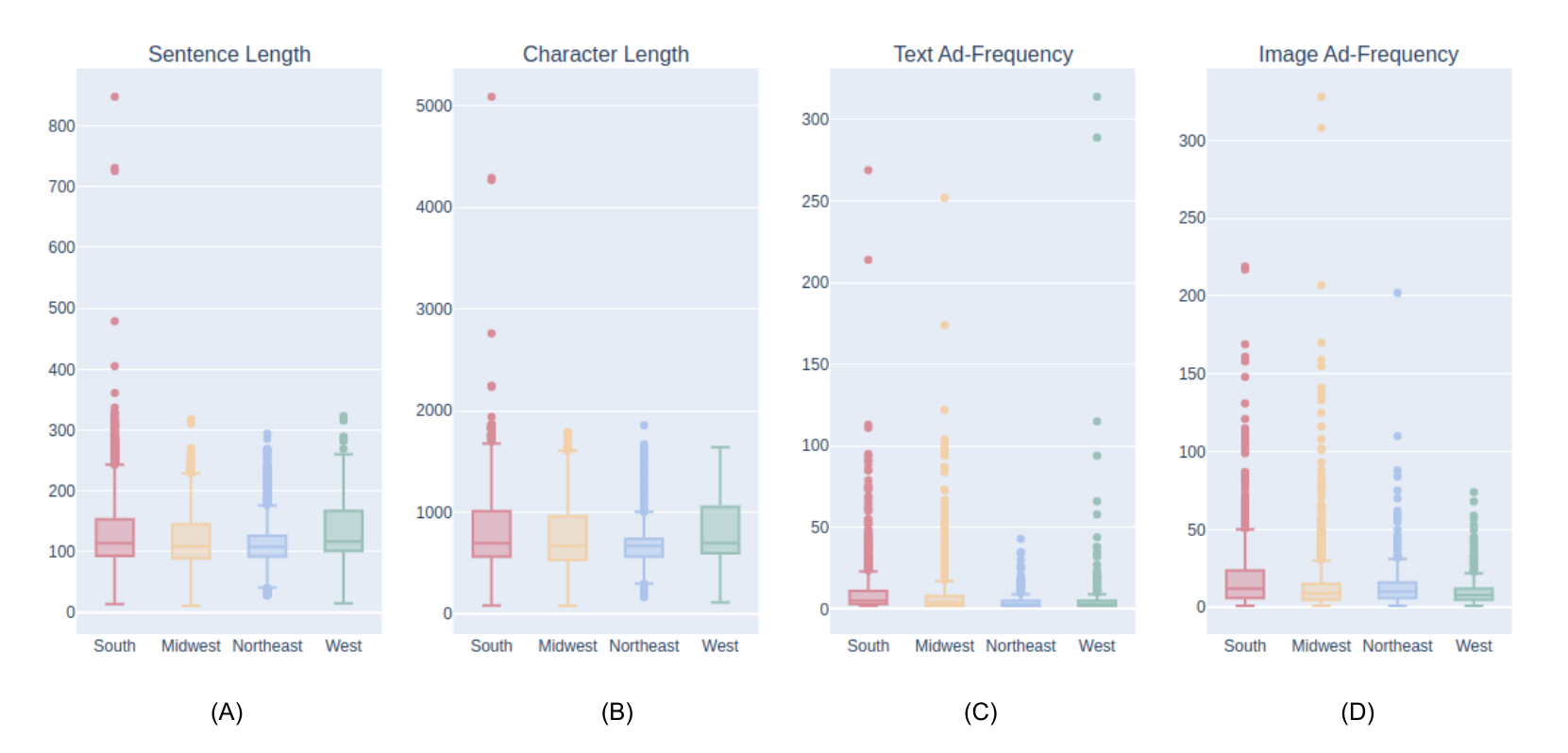}
        \caption{\label{fig:dataStats} Figures (A) and (B) showcase distributions of sentence and character length for text advertisements in the datasets. Figures (C) and (D) show a distribution of text-ad and image-ad frequency for each dataset, i.e., the number of text and image ads per vendor.}
    \end{subfigure}
    
    \vspace{0.5cm} 
    
    \begin{subfigure}[b]{\textwidth}
        \centering
        \includegraphics[width=\textwidth]{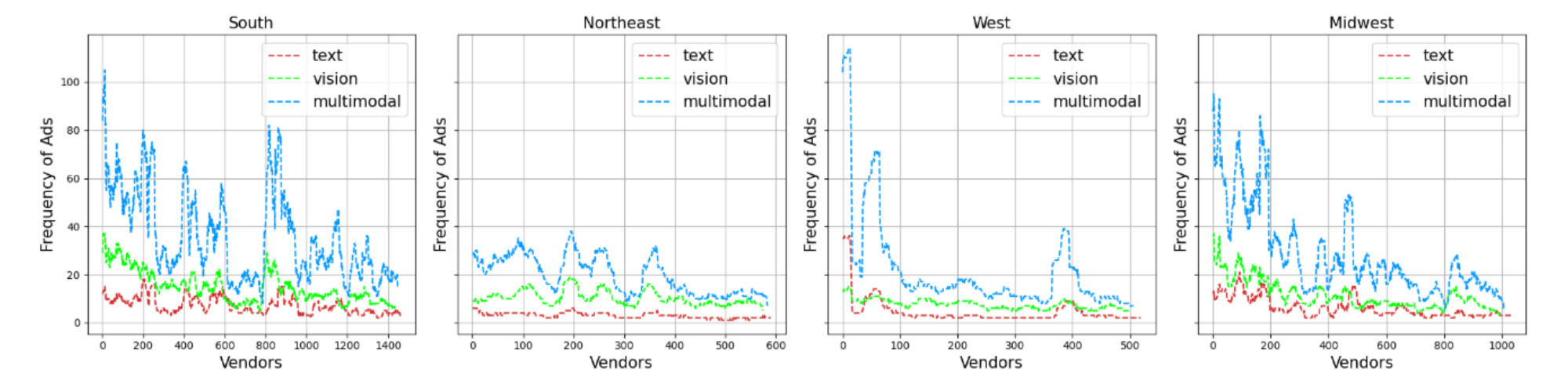}
        \caption{\label{fig:freq_distribution} Frequency of text, image, and multimodal ads in South, Northeast, West, and Midwest region datasets.}
    \end{subfigure}
    \caption{}
    \label{fig:data_stats_main}
\end{figure*}

\textbf{(i) Data Analysis:} Figure \ref{fig:dataSimilarity}(A) illustrates the \% of shared vendors across different datasets. As can be observed, many vendors post ads across multiple geographical regions, which aligns with existing findings that the Backpage escort marketplace was often flagged for HT activities, with vendors frequently advertising their services across various regions \citep{jrsa}. This cross-regional vendor activity also highlights a limitation in our OOD generalization experiments, which are designed to test the ability of our models to make predictions on data distribution that is different from the data it was trained on. These experiments may not fully capture real-world conditions. To properly assess true OOD generalization, future work would need to collect ads from an entirely separate escort platform to evaluate our models' adaptability to a new distribution of ads—an approach that lies outside the scope of this research.

\hspace{\parindent} Figure \ref{fig:dataSimilarity}(B) examines the average text-to-text similarity between ads from different datasets. Using a pre-trained DeCLUTR-small model, we compute the similarity by generating sentence embeddings for each ad and calculating the cosine similarity between pairs from different datasets. Given the high level of vendor overlap across regions, the text content is expected to exhibit considerable similarity. Similarly, figure \ref{fig:dataSimilarity}(C) shows the average image-to-image cosine similarity across ads from different datasets, calculated using representations from a pre-trained ViT-base-patch16 model. Compared to the relatively high text similarity, the image similarity is lower. This suggests that, while vendors often maintain consistent writing styles across regions, they tend to vary the images posted, potentially to depict different escorts. 

\hspace{\parindent} Figure \ref{fig:dataStats}(A) and (B) illustrate the sentence and character length distributions of text ads within our datasets. Sentence length is measured by counting the total number of tokens generated by the pre-trained DeCLUTR-small checkpoint after tokenization, while character length is the count of characters in each text ad. As shown, most text ads have a sentence length of fewer than 512 tokens. Therefore, we truncate all text ads to a maximum length of 512 tokens, also the maximum sequence length allowed by most transformers-based models. Figure \ref{fig:dataStats}(C) depicts the text-ad frequency, i.e., the number of text ads posted per vendor. As evident, most vendors post between 1 and 20 text ads. Unlike other authorship attribution (AA) approaches applied to criminal markets, which require a minimum of 5 \citep{saxena-etal-2023-idtraffickers} or 20 \citep{saxena-etal-2023-vendorlink} ads for effective AA implementation, our research explores the applicability of AA techniques for vendors with as few as two ads. This distribution of ad frequency highlights a class imbalance in our dataset, prompting us to prioritize Macro-F1 performance to ensure equal weighting across all classes in our classification tasks. Similarly, \ref{fig:dataStats}(D) depicts the image-ad frequency or the number of image ads posted per vendor. As evident, most vendors post between 5 and 24 image ads. A detailed analysis of the frequency of text, image, and multimodal ads per vendor is attached in Figure \ref{fig:freq_distribution} \footnote{Note that these line plots are plotted with a smoothing applied to window size of 30 for better readability.}. Finally, our language analysis using the \href{https://huggingface.co/ERCDiDip/langdetect}{LangDetect model} \citep{ercdidip2022} reveals that the vast majority of text ads are in English: 99.65\% in the South dataset, 99.98\% in the Midwest dataset, 99.88\% in the West dataset, and 99.85\% in the Northeast dataset. \\

\noindent \textbf{(ii) Data Pre-Processing:} \noindent As described in Section \ref{sec:dataset}, our dataset is sourced from Backpage escort ads posted across seven US cities between December 2015 and April 2016. We scrape titles, descriptions, and images for each ad. The text sequence for each entry is created by combining the title and description separated with a "[SEP]" token. Since ads may contain multiple images, we duplicate the text sequence for each associated image to prepare the dataset for multimodal training.

\hspace{\parindent} To establish ground truth, we follow \citet{saxena-etal-2023-idtraffickers} and utilize tools from \citet{nagpal2017entity, chambers-etal-2019-character} and \citet{SciPyProceedings_11} to extract phone numbers and form vendor communities, aka vendor labels. Consistent with \citet{saxena-etal-2023-idtraffickers}, we mask most personal information, including phone numbers, escort ages, measurements, ad IDs, and posting dates. Despite attempts to mask all identifiable information, existing Named Entity Recognizers (NER) \citep{li-etal-2022-extracting, liu-etal-2023-sweet} struggle to extract escort names from the ads reliably. However, since escorts generally use pseudonyms in these ads \citep{CARTER2021100065, GraulichSexTrafficking}, the potential for misuse of personal data is already minimal. 

\hspace{\parindent} For image anonymization, we initially considered blurring faces to protect escort identities. However, manual inspection revealed that many images with blurred or cropped faces are anonymously posted. To preserve these stylistic elements, we opted not to add artificial blurring, which could introduce visual biases. Similarly, we avoided other image augmentation techniques, as transformations such as flipping or rotating could alter stylistic cues linked to specific vendors. Some ads naturally feature mirrored or rotated images, which are retained to prevent misattribution. To further analyze model behavior, we categorized the image dataset into "Face" and "No Face" subsets for each of the four regions—South, West, Midwest, and Northeast—using a pre-trained FaceNet model \citep{10127799}. FaceNet detects and generates bounding boxes around faces in images, which are then assigned to that region's "Face" dataset. \\

\noindent \textbf{(iii) Language Distribution:} Our analysis reveals that approximately 99.84\% of our dataset's vocabulary is English. Given that only a small fraction of our dataset's vocabulary lies outside English, we anticipate that employing multilingual models would have a negligible effect on model performance. These statistics are obtained using the \href{https://huggingface.co/ERCDiDip/langdetect}{LangDetect} \citep{ercdidip2022} python model.

\subsection{Datasheet}
\label{app:datasheet}

Following \citet{gebru2021datasheetsdatasets}, we provide the datasheet for our MATCHED dataset below:

\subsubsection{Motivation}
\paragraph{For what purpose was the dataset created? Was there a specific task in mind? Was there a specific gap that needed to be filled? Please provide a description.} The MATCHED dataset was created to support LEAs, investigators, and researchers in identifying vendor connections within online escort ads. Traditional methods often rely on explicit personal identifiers such as phone numbers and email addresses. However, existing research shows that only a small fraction of ads include this information, limiting the effectiveness of these approaches. In response, \citet{saxena-etal-2023-idtraffickers} introduced AA methods to connect escort vendors through stylistic similarity in text, providing an alternative way to link ads without direct identifiers. Our dataset fills a critical gap by incorporating textual descriptions and images associated with escort ads, enabling researchers to move beyond text-only analysis. This multimodal dataset allows for the exploration of multimodal training strategies that integrate both text and images, aimed at improving the robustness and generalizability of AA in the context of HT detection. 



\subsubsection{Composition}

\paragraph{What do the instances that comprise the dataset represent (e.g., documents, photos, people, countries)? Are there multiple types of instances (e.g., movies, users, and ratings; people and interactions between
them; nodes and edges)? Please provide a description.} The instances in the MATCHED dataset represent individual ads from online escort services. Each ad instance comprises two main components: (1) a raw text sequence created by merging the title and description of the escort ad with a [SEP] token separating them, and (2) one or more images associated with the ad, typically depicting the escort being advertised. Each ad instance is then connected to a vendor ID, a unique identifier representing the individual or organization responsible for posting the ad. This vendor ID enables the grouping of ads by their source, supporting the AA task and facilitating the connection of ads linked to the same vendor.  

\paragraph{How many instances are there in total (of each type, if appropriate)?  What data does each instance consist of? “Raw” data (e.g., unprocessed text or images)or features? Is there a label or target associated with each instance?} The MATCHED dataset consists of 28,513 ad instances, including 27,619 unique text descriptions and 55,115 escort images linked to 3549 unique vendors. Each instance in the dataset comprises "raw" data from unprocessed text and images. The dataset is provided as a pandas DataFrame in a .csv format, with three main columns: "TEXT," "IMAGES," and "VENDOR." The "TEXT" column contains the input text sequence in string format, created by merging the title and description of the ad. The "IMAGES" column holds the local file path for each image associated with the ad. The "VENDOR" column includes the class labels, represented as integer IDs corresponding to specific vendors. Further details on dataset composition and split are outlined in Table \ref{tab:data-stats}.

\paragraph{Does the dataset contain all possible instances or is it a sample (not
necessarily random) of instances from a larger set? If the dataset is
a sample, then what is the larger set? Is the sample representative of the
larger set (e.g., geographic coverage)? If so, please describe how this
representativeness was validated/verified. If it is not representative of the
larger set, please describe why not (e.g., to cover a more diverse range of
instances, because instances were withheld or unavailable).} The MATCHED dataset represents a sample of the broader Backpage escort market data, with ads collected from seven cities across five U.S. states. To ensure a reliable ground truth for AA tasks, we filtered the ads to include only those with phone numbers (used to establish vendor labels) and at least one image. This filtering process resulted in a final set of 28,513 ads. Consequently, while the dataset does not fully represent the entire Backpage escort market, it focuses on instances where both text and image modalities are available, which is essential for exploring MAA.


\paragraph{Is any information missing from individual instances? If so, please
provide a description, explaining why this information is missing (e.g., because it was unavailable). This does not include intentionally removed
information, but might include, e.g., redacted text.}
No

\paragraph{Are relationships between individual instances made explicit (e.g.,
users’ movie ratings, social network links)? If so, please describe
how these relationships are made explicit.} Relationships between instances in our dataset are established by extracting and grouping phone numbers found within ads. Using the TJBatchExtractor \citep{nagpal2017entity} and CNN-LSTM-CRF classifier \citep{chambers-etal-2019-character}, we identify phone numbers that act as identifiers for vendors. These identifiers are then used to construct vendor communities via NetworkX \citep{SciPyProceedings_11}, where each community corresponds to a unique vendor label. This approach links ads to individual or organizational entities (vendors) by grouping ads associated with the same phone number, creating a structured relationship among instances in the dataset.

\paragraph{Are there recommended data splits (e.g., training, development/validation, testing)? If so, please describe these
splits, explaining the rationale behind them.} We split our dataset into training, validation, and test sets using a 0.75:0.05:0.20 split ratio. This allocation is intended to provide a substantial training set (75\%) for effective model learning, a validation set (5\%) for tuning model hyperparameters and avoiding overfitting, and a test set (20\%) to assess model generalization and in-distribution performance.

\paragraph{Are there any errors, sources of noise, or redundancies in the
dataset? If so, please provide a description.} As indicated by \citet{saxena-etal-2023-idtraffickers}, a considerable amount of noise is present in the Backpage escort ads. In the text data, vendors often add extra punctuation, emojis, irregular white spaces, and random characters, likely as a tactic to circumvent automated detection systems. These irregularities can impact text processing and add complexity to data-cleaning efforts. Our manual inspection of the image data also reveals visual noise, including intentionally blurred areas and white noise, which further complicates the analysis. However, quantifying the extent of this noise in images remains challenging. Despite these issues, the noise and irregularities reflect the original conditions in which the data was originally posted, providing a realistic foundation for developing robust AA models that can handle similar situations in real-world applications. 

\paragraph{Is the dataset self-contained, or does it link to or otherwise rely on
external resources (e.g., websites, tweets, other datasets)? If it links to or relies on external resources, a) are there guarantees that they will exist, and remain constant, over time; b) are there official archival versions of
the complete dataset (i.e., including the external resources as they existed
at the time the dataset was created); c) are there any restrictions (e.g.,
licenses, fees) associated with any of the external resources that might
apply to a dataset consumer? Please provide descriptions of all external
resources and any restrictions associated with them, as well as links or
other access points, as appropriate} No. The dataset is self-contained. 

\paragraph{Does the dataset contain data that might be considered confidential
(e.g., data that is protected by legal privilege or by doctor–patient
confidentiality, data that includes the content of individuals’ nonpublic communications)? If so, please provide a description.}  Building on the guidelines by \citet{saxena-etal-2023-idtraffickers}, we also include measures to minimize privacy risks and mitigate data misuse. We anonymize sensitive details in text by replacing digits with the letter "N" and substituting email addresses with $<EMAIL_ID>$, post IDs with $POST_ID: NNNNN$, dates with $<DATES>$, and links with $<LINK>$. Attempts were made to mask escort names and locations using NER models \citep{li-etal-2022-extracting, liu-etal-2023-sweet}, but noise in the data led to inaccurate predictions. Nevertheless, as previous studies suggest that escorts often use pseudonyms \citep{CARTER2021100065, GraulichSexTrafficking}, the potential for misuse of personal details in text ads is low. 

That said, we recognize that identities could still be inferred from images. Initially, we considered blurring faces to enhance anonymity. However, manual inspection showed that many images already had faces blurred or cropped by the posters. To retain these natural stylistic cues, we decided against additional blurring, as it could interfere with AA tasks and introduce unintended biases in the visual data. Additionally, a sanity check using the FairFace \citep{karkkainenfairface} and DeepFace \citep{serengil2023db} models demonstrated that these tools, when applied to our noisy dataset, were unable to extract any ethnicity or age-related information from the dataset's images.

\paragraph{Does the dataset contain data that, if viewed directly, might be offensive, insulting, threatening, or might otherwise cause anxiety? If so,
please describe why.} Yes, the dataset comprises text and (semi-nude) images from escort advertisements that contain sexual descriptions. 

\paragraph{Does the dataset identify any subpopulations (e.g., by age, gender)?
If so, please describe how these subpopulations are identified and provide
a description of their respective distributions within the dataset.} Our dataset does not explicitly identify subpopulations by age, as all age information has been masked in the text ads. However, some ads include descriptions of the escorts' ethnicities, which remain unmasked to preserve the original stylometric features for AA tasks. Additionally, most ads in our dataset correspond to women-based escort services. It is important to note that while we have not provided age or ethnicity labels, malicious users could potentially infer such details by applying automated systems to the images. This potential for inference underscores the importance of responsible dataset usage and adherence to ethical guidelines to prevent misuse.

\paragraph{Is it possible to identify individuals (i.e., one or more natural persons), either directly or indirectly (i.e., in combination with other
data) from the dataset? If so, please describe how.} While we cannot entirely rule out the possibility of identifying individuals through our dataset, we have followed extensive privacy measures pointed out by \citep{saxena-etal-2023-idtraffickers} to minimize this risk. In the text ads, we have masked private identifiers, such as phone numbers, email addresses, and other personal information, to protect the identities of individuals. The dataset comprises ads from the Backpage escort market collected between December 2015 and April 2016, a period for which there are no longer public records since the website was seized. However, there remains a risk associated with the images in our dataset, as they may still allow for indirect identification of individuals.

To mitigate this risk, we will restrict access to the MATCHED dataset, allowing only approved researchers or agencies with legitimate research objectives—specifically those focused on combating HT or conducting academic (non-commercial) research related to AA. Access will be granted through a data portal, \href{https://doi.org/10.34894/UR3RVE}{DataverseNL}, subject to approval from our ethics review board, which ensures that the dataset is used solely for its intended purposes. Unauthorized use of the dataset, particularly for purposes beyond AA or HT research, is strictly prohibited under our ethical guidelines and will have legal repercussions.

\paragraph{Does the dataset contain data that might be considered sensitive in
any way (e.g., data that reveals race or ethnic origins, sexual orientations, religious beliefs, political opinions or union memberships, or
locations; financial or health data; biometric or genetic data; forms of
government identification, such as social security numbers; criminal
history)? If so, please provide a description.} Despite our masking efforts, our dataset still contains sensitive information. While we have successfully masked certain private identifiers, such as phone numbers and email addresses, challenges remain in masking other potentially sensitive details, including escort names, ad locations, ethnicities, and sexual orientations. These details are present in the ads' text descriptions and could be extracted from the images using automated systems. The inherent noise in the data further complicates the accurate masking of these elements. As a result, while we have taken significant precautions, there remains a possibility that sensitive information could be inferred from the dataset.

\subsubsection{Collection Process}

\paragraph{How was the data associated with each instance acquired? Was the
data directly observable (e.g., raw text, movie ratings), reported by subjects (e.g., survey responses), or indirectly inferred/derived from other data
(e.g., part-of-speech tags, model-based guesses for age or language)? If
the data was reported by subjects or indirectly inferred/derived from other
data, was the data validated/verified? If so, please describe how.} The data for each instance was acquired from raw text and images associated with escort ads posted on the Backpage market. Following \citet{saxena-etal-2023-idtraffickers}, we utilized the TJBatchExtractor \citep{nagpal2017entity} and a CNN-LSTM-CRF classifier \citep{chambers-etal-2019-character} to extract phone numbers from these ads, which serve as identifiers to group ads into vendor communities. NetworkX \citep{SciPyProceedings_11} was subsequently used to build these communities, assigning a unique label ID to each vendor. 

\paragraph{What mechanisms or procedures were used to collect the data (e.g.,
hardware apparatuses or sensors, manual human curation, software programs, software APIs)? How were these mechanisms or procedures
validated?} The raw data is provided to us from \href{https://bashpolesoftware.com/bashpole-intelligence-suite/}{Bashpole Software, Inc.}. 



\paragraph{Did you collect the data from the individuals in question directly, or
obtain it via third parties or other sources (e.g., websites)? Over what timeframe was the data collected? Does this timeframe
match the creation timeframe of the data associated with the instances
(e.g., recent crawl of old news articles)? If not, please describe the timeframe in which the data associated with the instances was created.}  The MATCHED dataset contains ads from seven US cities and is scraped from online posted ads between December 2015 and April 2016 on the Backpage Escort Markets. The raw data is provided to us from \href{https://bashpolesoftware.com/bashpole-intelligence-suite/}{Bashpole Software, Inc.}.


\paragraph{Were the individuals in question notified about the data collection?
If so, please describe (or show with screenshots or other information) how
notice was provided, and provide a link or other access point to, or otherwise reproduce, the exact language of the notification itself.} The individuals in our ads were not notified about the data collection. Given that the ads were posted on Backpage between December 2015 and April 2016, obtaining consent from these individuals is infeasible. Since the Backpage escort market was seized and shut down, reconnecting with these individuals—many of whom used pseudonyms and transient contact information like phone numbers or email addresses—is impractical after such a long period. Additionally, as Backpage no longer exists as a platform, contacting the original poster would be challenging and unlikely to yield responses.

\paragraph{Did the individuals in question consent to the collection and use of
their data? If so, please describe (or show with screenshots or other
information) how consent was requested and provided, and provide a link
or other access point to, or otherwise reproduce, the exact language to
which the individuals consented.} No.

\paragraph{If consent was obtained, were the consenting individuals provided
with a mechanism to revoke their consent in the future or for certain
uses? If so, please provide a description, as well as a link or other access
point to the mechanism (if appropriate).} NA


\subsubsection{Preprocessing/cleaning/labeling}

\paragraph{Was any preprocessing/cleaning/labeling of the data done (e.g., discretization or bucketing, tokenization, part-of-speech tagging, SIFT
feature extraction, removal of instances, processing of missing values)? If so, please provide a description. If not, you may skip the remaining questions in this section.} To prioritize privacy and reduce the risk of misuse, we implemented extensive preprocessing and cleaning procedures to protect sensitive information within the text descriptions in our dataset. This involved masking identifiable elements, including phone numbers, email addresses, age details, post IDs, dates, and links mentioned in the ads. The images are not processed or cleaned to maintain their original stylometric cues. Finally, since the goal of our research is MAA, we removed all instances that did not contain phone numbers or images.

\paragraph{Was the “raw” data saved in addition to the preprocessed/cleaned/labeled data (e.g., to support unanticipated
future uses)? If so, please provide a link or other access point to the
“raw” data.} No.

\paragraph{Is the software that was used to preprocess/clean/label the data available? If so, please provide a link or other access point.} No.

\subsubsection{Uses}

\paragraph{Has the dataset been used for any tasks already? If so, please provide
a description.} This research introduces MATCHED, a novel dataset comprising text descriptions and images from Backpage escort markets, specifically developed for MAA. While MATCHED has not been utilized in any previous studies, several works have reportedly used text descriptions or images from Backpage escort marketplaces for similar analyses \citep{7745456, 10.1145/3097983.3098082, alvari2017semisupervisedlearningdetectinghuman, saxena-etal-2023-idtraffickers}, etc. However, due to the unavailability of these datasets, we could not verify whether any ads overlap with those in MATCHED. 


\paragraph{What (other) tasks could the dataset be used for?} The MATCHED dataset is strictly intended for use in AA tasks related to combating human trafficking or conducting academic research within ethical boundaries. Our ethics review board has implemented strict guidelines prohibiting using this dataset beyond these purposes. Consequently, we discourage any other applications, as they could risk potential misuse or ethical concerns that are not aligned with the dataset’s purpose and ethical considerations. 

\paragraph{Is there anything about the composition of the dataset or the way it
was collected and preprocessed/cleaned/labeled that might impact
future uses? For example, is there anything that a dataset consumer
might need to know to avoid uses that could result in unfair treatment of
individuals or groups (e.g., stereotyping, quality of service issues) or other
risks or harms (e.g., legal risks, financial harms)? If so, please provide
a description. Is there anything a dataset consumer could do to mitigate
these risks or harms?} Although we have taken extensive precautions to mask sensitive information, our dataset still includes details like escort pseudonyms, posted locations, ethnicity, and sexual preferences, which could be potentially sensitive. While these details are unlikely to be used to harm individuals directly, we strongly caution against any unethical applications, particularly those that could lead to re-identifying individuals or otherwise compromising their privacy. This includes any research or commercial use aimed at profiling, targeting, or stereotyping. To mitigate these risks, we advise dataset consumers to strictly adhere to ethical guidelines, focusing solely on the dataset's intended purpose of combating human trafficking through academic research. Additionally, we encourage the users to implement further anonymization techniques, especially if using images, and avoid practices that could unintentionally expose or unfairly represent individuals or groups in the dataset.


\subsubsection{Distribution}

\paragraph{Will the dataset be distributed to third parties outside of the entity (e.g., company, institution, organization) on behalf of which the
dataset was created? If so, please provide a description.} Yes, we plan to make our dataset accessible to third parties via the \href{https://doi.org/10.34894/UR3RVE}{DataverseNL} data repository. To mitigate risks of illegal or unethical use, access will be granted under specific conditions, including mandatory signing of a non-disclosure agreement (NDA) and data protection agreements. Each application for access will be evaluated by our ethics committee to ensure alignment with the dataset’s intended purpose. These agreements will prohibit data redistribution and restrict its use exclusively to ethical, non-commercial research, especially in contexts that support combating HT.

\paragraph{How will the dataset will be distributed (e.g., tarball on website, API,
GitHub)? Does the dataset have a digital object identifier (DOI)?} Yes

\paragraph{When will the dataset be distributed?} The MATCHED dataset will be released after the final decision from the ACL committee, along with the camera-ready version. 


\paragraph{Have any third parties imposed IP-based or other restrictions on the
data associated with the instances? If so, please describe these restrictions, and provide a link or other access point to, or otherwise reproduce,
any relevant licensing terms, as well as any fees associated with these
restrictions.} No


\subsubsection{Maintenance}




\paragraph{Will the dataset be updated (e.g., to correct labeling errors, add new
instances, delete instances)? If so, please describe how often, by
whom, and how updates will be communicated to dataset consumers (e.g.,
mailing list, GitHub)?} We are committed to enhancing the dataset by exploring advanced NLP-based entity extraction techniques to protect individual privacy further. Specifically, we aim to implement more effective methods for masking escort pseudonyms, posted locations, and ethnicities. Additionally, we plan to expand the dataset by including ads from multiple escort platforms, enabling us to evaluate our models' generalization on real-to-close-world OOD datasets. These updates aim to improve the dataset’s privacy measures and its utility for robust, cross-platform AA tasks. Progress and updates will be communicated through research publications, and detailed updates will be made to the dataset’s description on the \href{https://doi.org/10.34894/UR3RVE}{DataverseNL} portal.


\paragraph{If others want to extend/augment/build on/contribute to the dataset,
is there a mechanism for them to do so? If so, please provide a description. Will these contributions be validated/verified? If so, please describe
how. If not, why not? Is there a process for communicating/distributing
these contributions to dataset consumers? If so, please provide a description.} We encourage researchers to collaborate with us to extend and improve the dataset through extensions, augmentations, or related enhancements. To safeguard the privacy and well-being of individuals in the dataset, we have restricted sharing rights, meaning contributors cannot freely distribute the dataset. However, we invite researchers to work with us directly, and we are open to reviewing and integrating validated contributions to improve the dataset’s utility responsibly. We ensure that all validated contributions and updates will be acknowledged and communicated to the research community.

\subsection{Infrastructure \& Schedule}
\label{app:experimental_details}

\paragraph{Split Ratio:} We split the dataset into training, validation, and test sets using a standard ratio of 0.75:0.05:0.20 for our experiments. During this process, we set the seed parameter to 1111 for reproducibility.

\paragraph{Training:} We conduct model training and evaluation on an NVIDIA H100 GPU with 80 GB of memory. For optimization, we use the Adam optimizer configured with $\beta$1 and $\beta$2 values of 0.9 and 0.999, respectively, along with an L2 weight decay of 0.01. We experiment with learning rates of 0.01, 0.001, and 0.0001, ultimately finding the best performance at a learning rate of 0.001. Additionally, we apply a warm-up strategy for the first 100 steps, followed by a linear decay schedule.

\paragraph{Architectures \& Hyperparameters:} Considering our computational constraints, we initialize text baselines using pre-trained model checkpoints from \href{https://huggingface.co/johngiorgi/declutr-small}{DeCLUTR-small} and \href{https://huggingface.co/AnnaWegmann/Style-Embedding}{Style-Embedding} architectures. Similarly, vision baselines are initialized using pre-trained checkpoints from \href{https://huggingface.co/timm/vgg16.tv_in1k}{VGG-16}, \href{https://huggingface.co/timm/resnet50.a1_in1k}{ResNet-50}, \href{https://huggingface.co/timm/densenet121.tv_in1k}{DenseNet-121}, \href{https://huggingface.co/docs/timm/en/models/inception-v3}{InceptionNetV3}, \href{https://huggingface.co/timm/efficientnetv2_rw_m.agc_in1k}{EfficientNetV2}, \href{https://huggingface.co/timm/convnext_small.fb_in22k}{ConvNext-small}, and \href{https://huggingface.co/google/vit-base-patch16-224}{ViT-base-patch16-244} architectures. We also explore face recognition models such as VGG-Face2 \citep{cao2018vggface2datasetrecognisingfaces}, ArcFace, FaceNet512 \citep{10127799}, and GhostFaceNet \citep{10098610} from DeepFace \citep{serengil2023db} for the vision baselines. However, these models struggle with vendor identification and verification tasks, likely because they focus solely on facial features, making it challenging to connect multiple faces to a single vendor. We further experimented by training these face recognition models on the face (images with faces) and no face (images without faces) subsets of our dataset. However, the results remained consistent, confirming their unsuitability for these tasks. Finally, the multimodal baselines are initialized by combining the DeCLUTR-small and ViT-base-patch-244 baselines to process text and vision modalities. Each model is equipped with a sequence classification head to perform classification tasks. Due to resource limitations, all models are trained with a batch size of 32, the maximum feasible size, and training continues until convergence.

During model training, we use five in-batch negatives for contrastive objectives such as Triplet, SupCon, CE+Triplet, and CE+SupCon. Increasing the number of in-batch negatives did not improve performance, likely constrained by the fixed batch size of 32 for the classification task. For the text-image alignment pre-training task, we employ the Normalized Temperature-Scaled Cross-Entropy (NT-XENT) loss \citep{chen2020simpleframeworkcontrastivelearning} for the Image-Text Contrastive (ITC) objective, sampling negatives from regions outside the training dataset. In all multimodal experiments, negatives are strictly non-associated, ensuring text-image pairs are unrelated ads. We also experiment with temperature coefficient values of 0.01, 0.1, and 0.3 for the NT-XENT loss, finding the best performance at 0.1.

The experiments are implemented in \href{https://www.python.org/downloads/release/python-3100/}{Python 3.10} using frameworks such as \href{https://scikit-learn.org/}{scikit-learn} \citep{pedregosa2011scikit}, \href{https://pytorch.org/}{PyTorch} \citep{NEURIPS2019_bdbca288}, \href{https://huggingface.co/}{Hugging Face}, \href{https://huggingface.co/timm}{timm}, and \href{https://github.com/Lightning-AI/lightning}{Lightning 2.0} \citep{Falcon_PyTorch_Lightning_2019}. The plots in the research are developed using \href{https://matplotlib.org/}{Matplotlib} \citep{4160265} and \href{https://plotly.com/python/}{Plotly} \citep{plotly}.

\paragraph{Computational Details:} Table \ref{tab:all_vendor_identification_results} provides an overview of the number of trainable parameters, training time, and convergence epochs for all the classifiers evaluated in our experiments. Additionally, we dedicated 8 hours 21 minutes and 51 seconds, 1 hour 51 minutes and 6 seconds, and 3 hours 52 minutes and 12 seconds to pre-train our text-image alignment models using ITC (CLIP), ITC+ITM (Image Text Matching loss), and ITC+ITM+Text Generation Loss (BLIP2) training strategies, respectively.

\begin{table}[h]
\resizebox{\columnwidth}{!}{%
\begin{tabular}{lllll}
\hline
\textbf{Seed} & \textbf{Acc.} & \textbf{Weighted-F1} & \textbf{Micro-F1} & \textbf{Macro-F1} \\ \hline \hline
\textbf{100}  & 0.9670        &  0.9862      & 0.9878            & 0.9630            \\
\textbf{500}  & 0.9761        & 0.9914               & 0.9921            & 0.9755            \\
\textbf{1111} & 0.9823        & 0.9911               & 0.9916            & 0.9802            \\ \hline
\textbf{Mean} & 0.9751        & 0.9896               & 0.9905            & 0.9729            \\
\textbf{Std.} & 0.0077        & 0.0029               & 0.0024            & 0.0089            \\ \hline
\end{tabular}
}
\caption{Influence of random initialization on DeCLUTR-ViT classifier's performance}
\label{tab:seed}
\end{table}

\paragraph{Random Initialization:} Due to limited resources, we only examine the effects of different initializations on our model's performance for the established DeCLUTR-ViT benchmark with the CE+SupCon objective. Table \ref{tab:seed} displays the mean and standard deviation in the model's performance against balanced accuracy, Micro-F1, Weighted-F1, and Macro-F1 scores. The results indicate minimal to no effects on these scores across different initializations.

\subsection{Model Performance}
\label{app:model_performance}

This section provides detailed insights into our experiments' training and evaluation results, as summarized in the appendix tables. Table \ref{tab:all_vendor_identification_results} outlines the performance of text-only, vision-only, and multimodal classifiers on the vendor identification task. These classifiers were trained on the South region dataset and evaluated using Balanced Accuracy, Weighted-F1, Micro-F1, and Macro-F1 metrics. Given the class imbalance in our datasets, we emphasize Macro-F1 as the primary metric to assess model performance effectively. The models were trained with various objectives, including CE, Triplet, SupCon, CE+Triplet, and CE+SupCon, allowing a comprehensive comparison of their capabilities.

For retrieval tasks, results are detailed in Tables \ref{tab:text_retrieval_results}, \ref{tab:image-baselines-retrieval}, \ref{tab:vit-retrieval-results}, \ref{tab:multimodal_text_retrieval}, \ref{tab:multimodal_vision_retrieval}, and \ref{tab:multimodal_retrieval}, covering text-to-text, image-to-image, and multimodal retrieval scenarios. While we analyze all three retrieval metrics—MRR@10, R-Precision, and Macro-F1@X—our emphasis is on R-Precision. This metric reflects the model's ability to retrieve all relevant ads linked to a query ad from the same vendor, offering a direct measure of retrieval effectiveness.

As explained in the main manuscript, the Zero-Shot (ZS) performance refers to the capability of pre-trained models to perform retrieval tasks without prior AA training. Pre-trained text-only model is represented in Table \ref{tab:text_retrieval_results}, vision-only models in Table \ref{tab:vit-retrieval-results}, and text-image alignment models, as Aligned DeCLUTR-ViT, in Tables \ref{tab:multimodal_text_retrieval}, \ref{tab:multimodal_vision_retrieval}, and \ref{tab:multimodal_retrieval}. These models are evaluated on the South, Midwest, West, and Northeast region datasets without specific AA task training, making them ideal for understanding baseline performance in unseen contexts. Conversely, the Out-of-Distribution (OOD) average performance measures how well AA models trained for vendor identification or verification tasks generalize to unseen datasets from the Midwest, West, and Northeast regions. This evaluation highlights the models' robustness in handling diverse, previously unseen ads and vendors, offering critical insights into their cross-region generalization capabilities. By contrasting ZS and OOD performance, we assess both the initial adaptability of pre-trained models and the impact of AA-specific training. All the vendor verification metrics are represented in x ± y format, where x and y represent the mean and standard deviation of performance across all vendor classes. 

\subsubsection{Text-only Modality}
The text-baseline results presented in Table \ref{tab:all_vendor_identification_results} demonstrate that the DeCLUTR-small architecture significantly outperforms the Style-Embedding model in terms of Macro-F1 score for the vendor identification task. As a result, the DeCLUTR-small architecture is exclusively used for further experiments involving joint objectives. Among all text-only baselines, the DeCLUTR backbone trained with the CE+SupCon objective achieves the highest performance across all vendor identification metrics, showcasing its effectiveness. For the vendor verification task, retrieval results in Table \ref{tab:text_retrieval_results} reveal that the DeCLUTR backbone trained with the CE+SupCon objective consistently outperforms the CE objective and performs comparably to the SupCon-only objective. Additionally, the smaller standard deviation in performance between the CE+SupCon and CE objectives highlights the model's enhanced consistency across all vendor classes, further underscoring the robustness of the CE+SupCon objective for text-only baselines.

\begin{table*}[h]
\centering
\resizebox{\linewidth}{!}{%
\begin{tabular}{|cccccccccc|}
\hline
\multicolumn{1}{|c|}{\textbf{Model}}                                                     & \multicolumn{1}{c|}{\textbf{Param}}                                         & \multicolumn{1}{c|}{\textbf{Loss}}                             & \multicolumn{1}{c|}{\textbf{Fusion}}                                        & \multicolumn{1}{c|}{\textbf{Epochs}}                    & \multicolumn{1}{c|}{\textbf{Accuracy}}                      & \multicolumn{1}{c|}{\textbf{\begin{tabular}[c]{@{}c@{}}Weighted\\ F1\end{tabular}}} & \multicolumn{1}{c|}{\textbf{\begin{tabular}[c]{@{}c@{}}Micro\\ F1\end{tabular}}} & \multicolumn{1}{c|}{\textbf{\begin{tabular}[c]{@{}c@{}}Macro\\ F1\end{tabular}}} & \textbf{\begin{tabular}[c]{@{}c@{}}Time \\ (hrs.)\end{tabular}} \\ \hline
\multicolumn{10}{|c|}{\textbf{Text-Baselines}}                                                                                                                                                                                                                                                                                                                                                                                                                                                                                                                                                                                                                                                                                                                              \\ \hline
\multicolumn{1}{|c|}{\textbf{Style-Embedding}}                                           & \multicolumn{1}{c|}{128M}                                                   & \multicolumn{1}{c|}{CE}                                        & \multicolumn{1}{c|}{}                                                       & \multicolumn{1}{c|}{28}                                 & \multicolumn{1}{c|}{0.6582}                                 & \multicolumn{1}{c|}{0.6883}                                                         & \multicolumn{1}{c|}{0.6897}                                                      & \multicolumn{1}{c|}{0.5210}                                                      & 01:07:12                                                        \\ \cline{1-3} \cline{5-10} 
\multicolumn{1}{|c|}{{\color[HTML]{340096} }}                                            & \multicolumn{1}{c|}{{\color[HTML]{340096} }}                                & \multicolumn{1}{c|}{CE}                                        & \multicolumn{1}{c|}{}                                                       & \multicolumn{1}{c|}{21}                                 & \multicolumn{1}{c|}{0.7647}                                 & \multicolumn{1}{c|}{0.7772}                                                         & \multicolumn{1}{c|}{0.7777}                                                      & \multicolumn{1}{c|}{0.6379}                                                      & 0:12:19                                                         \\ \cline{3-3} \cline{5-10} 
\multicolumn{1}{|c|}{{\color[HTML]{340096} }}                                            & \multicolumn{1}{c|}{{\color[HTML]{340096} }}                                & \multicolumn{1}{c|}{CE+Triplet}                                & \multicolumn{1}{c|}{}                                                       & \multicolumn{1}{c|}{10}                                 & \multicolumn{1}{c|}{0.6905}                                 & \multicolumn{1}{c|}{0.7068}                                                         & \multicolumn{1}{c|}{0.7074}                                                      & \multicolumn{1}{c|}{0.5503}                                                      & 0:07:32                                                         \\ \cline{3-3} \cline{5-10} 
\multicolumn{1}{|c|}{\multirow{-3}{*}{{\color[HTML]{340096} \textbf{DeCLUTR-small}}}}    & \multicolumn{1}{c|}{\multirow{-3}{*}{{\color[HTML]{340096} \textbf{86M}}}}  & \multicolumn{1}{c|}{{\color[HTML]{340096} \textbf{CE+SupCon}}} & \multicolumn{1}{c|}{\multirow{-4}{*}{-}}                                    & \multicolumn{1}{c|}{{\color[HTML]{340096} \textbf{15}}} & \multicolumn{1}{c|}{{\color[HTML]{340096} \textbf{0.7786}}} & \multicolumn{1}{c|}{{\color[HTML]{340096} \textbf{0.7891}}}                         & \multicolumn{1}{c|}{{\color[HTML]{340096} \textbf{0.7898}}}                      & \multicolumn{1}{c|}{{\color[HTML]{340096} \textbf{0.6540}}}                      & {\color[HTML]{340096} \textbf{0:06:33}}                         \\ \hline
\multicolumn{10}{|c|}{\textbf{Vision-Baselines}}                                                                                                                                                                                                                                                                                                                                                                                                                                                                                                                                                                                                                                                                                                                            \\ \hline
\multicolumn{1}{|c|}{\textbf{VGG-16}}                                                    & \multicolumn{1}{c|}{138M}                                                   & \multicolumn{1}{c|}{}                                          & \multicolumn{1}{c|}{}                                                       & \multicolumn{1}{c|}{9}                                  & \multicolumn{1}{c|}{0.6823}                                 & \multicolumn{1}{c|}{0.6873}                                                         & \multicolumn{1}{c|}{0.6884}                                                      & \multicolumn{1}{c|}{0.5262}                                                      & 0:15:33                                                         \\ \cline{1-2} \cline{5-10} 
\multicolumn{1}{|c|}{\textbf{ResNet-50}}                                                 & \multicolumn{1}{c|}{25M}                                                    & \multicolumn{1}{c|}{}                                          & \multicolumn{1}{c|}{}                                                       & \multicolumn{1}{c|}{19}                                 & \multicolumn{1}{c|}{0.7741}                                 & \multicolumn{1}{c|}{0.7777}                                                         & \multicolumn{1}{c|}{0.7789}                                                      & \multicolumn{1}{c|}{0.6394}                                                      & 0:23:14                                                         \\ \cline{1-2} \cline{5-10} 
\multicolumn{1}{|c|}{\textbf{DenseNet-121}}                                              & \multicolumn{1}{c|}{7M}                                                     & \multicolumn{1}{c|}{}                                          & \multicolumn{1}{c|}{}                                                       & \multicolumn{1}{c|}{13}                                 & \multicolumn{1}{c|}{0.7624}                                 & \multicolumn{1}{c|}{0.7656}                                                         & \multicolumn{1}{c|}{0.7673}                                                      & \multicolumn{1}{c|}{0.6262}                                                      & 0:27:01                                                         \\ \cline{1-2} \cline{5-10} 
\multicolumn{1}{|c|}{\textbf{InceptionNetV3}}                                            & \multicolumn{1}{c|}{23M}                                                    & \multicolumn{1}{c|}{}                                          & \multicolumn{1}{c|}{}                                                       & \multicolumn{1}{c|}{12}                                 & \multicolumn{1}{c|}{0.7471}                                 & \multicolumn{1}{c|}{0.7510}                                                         & \multicolumn{1}{c|}{0.7524}                                                      & \multicolumn{1}{c|}{0.6047}                                                      & 0:20:26                                                         \\ \cline{1-2} \cline{5-10} 
\multicolumn{1}{|c|}{\textbf{EfficientNetV2}}                                            & \multicolumn{1}{c|}{23M}                                                    & \multicolumn{1}{c|}{}                                          & \multicolumn{1}{c|}{}                                                       & \multicolumn{1}{c|}{12}                                 & \multicolumn{1}{c|}{0.7652}                                 & \multicolumn{1}{c|}{0.7690}                                                         & \multicolumn{1}{c|}{0.7703}                                                      & \multicolumn{1}{c|}{0.6285}                                                      & 0:29:29                                                         \\ \cline{1-2} \cline{5-10} 
\multicolumn{1}{|c|}{\textbf{ConvNeXT-small}}                                            & \multicolumn{1}{c|}{50M}                                                    & \multicolumn{1}{c|}{\multirow{-6}{*}{CE}}                      & \multicolumn{1}{c|}{}                                                       & \multicolumn{1}{c|}{7}                                  & \multicolumn{1}{c|}{0.7593}                                 & \multicolumn{1}{c|}{0.7625}                                                         & \multicolumn{1}{c|}{0.7646}                                                      & \multicolumn{1}{c|}{0.6215}                                                      & 0:16:52                                                         \\ \cline{1-3} \cline{5-10} 
\multicolumn{1}{|c|}{{\color[HTML]{010066} }}                                            & \multicolumn{1}{c|}{{\color[HTML]{340096} }}                                & \multicolumn{1}{c|}{CE}                                        & \multicolumn{1}{c|}{}                                                       & \multicolumn{1}{c|}{8}                                  & \multicolumn{1}{c|}{0.7559}                                 & \multicolumn{1}{c|}{0.7593}                                                         & \multicolumn{1}{c|}{0.7606}                                                      & \multicolumn{1}{c|}{0.6142}                                                      & 0:13:16                                                         \\ \cline{3-3} \cline{5-10} 
\multicolumn{1}{|c|}{{\color[HTML]{010066} }}                                            & \multicolumn{1}{c|}{{\color[HTML]{340096} }}                                & \multicolumn{1}{c|}{CE+Triplet}                                & \multicolumn{1}{c|}{}                                                       & \multicolumn{1}{c|}{13}                                 & \multicolumn{1}{c|}{0.7729}                                 & \multicolumn{1}{c|}{0.7765}                                                         & \multicolumn{1}{c|}{0.7771}                                                      & \multicolumn{1}{c|}{0.6378}                                                      & 0:30:35                                                         \\ \cline{3-3} \cline{5-10} 
\multicolumn{1}{|c|}{\multirow{-3}{*}{{\color[HTML]{010066} \textbf{ViT-base-patch16}}}} & \multicolumn{1}{c|}{\multirow{-3}{*}{{\color[HTML]{340096} \textbf{86M}}}}  & \multicolumn{1}{c|}{{\color[HTML]{340096} \textbf{CE+SupCon}}} & \multicolumn{1}{c|}{\multirow{-9}{*}{-}}                                    & \multicolumn{1}{c|}{{\color[HTML]{340096} \textbf{13}}} & \multicolumn{1}{c|}{{\color[HTML]{340096} \textbf{0.7711}}} & \multicolumn{1}{c|}{{\color[HTML]{340096} \textbf{0.7709}}}                         & \multicolumn{1}{c|}{{\color[HTML]{340096} \textbf{0.7716}}}                      & \multicolumn{1}{c|}{{\color[HTML]{340096} \textbf{0.6294}}}                      & {\color[HTML]{340096} \textbf{0:31:41}}                         \\ \hline
\multicolumn{10}{|c|}{\textbf{Multimodal-Baselines}}                                                                                                                                                                                                                                                                                                                                                                                                                                                                                                                                                                                                                                                                                                                        \\ \hline
\multicolumn{1}{|c|}{\textbf{ViLT}}                                                      & \multicolumn{1}{c|}{112M}                                                   & \multicolumn{1}{c|}{}                                          & \multicolumn{1}{c|}{}                                                       & \multicolumn{1}{c|}{12}                                 & \multicolumn{1}{c|}{0.8454}                                 & \multicolumn{1}{c|}{0.8327}                                                         & \multicolumn{1}{c|}{0.8291}                                                      & \multicolumn{1}{c|}{0.7369}                                                      & 01:18:00                                                        \\ \cline{1-2} \cline{5-10} 
\multicolumn{1}{|c|}{\textbf{VisualBERT}}                                                & \multicolumn{1}{c|}{197M}                                                   & \multicolumn{1}{c|}{\multirow{-2}{*}{CE}}                      & \multicolumn{1}{c|}{\multirow{-2}{*}{-}}                                    & \multicolumn{1}{c|}{11}                                 & \multicolumn{1}{c|}{0.9652}                                 & \multicolumn{1}{c|}{0.9637}                                                         & \multicolumn{1}{c|}{0.9641}                                                      & \multicolumn{1}{c|}{0.9355}                                                      & 01:10:17                                                        \\ \hline
\multicolumn{1}{|c|}{{\color[HTML]{340096} }}                                            & \multicolumn{1}{c|}{}                                                       & \multicolumn{1}{c|}{}                                          & \multicolumn{1}{c|}{auto}                                                & \multicolumn{1}{c|}{11}                                 & \multicolumn{1}{c|}{0.9344}                                 & \multicolumn{1}{c|}{0.9578}                                                         & \multicolumn{1}{c|}{0.9565}                                                      & \multicolumn{1}{c|}{0.9121}                                                      & 03:41:44                                                        \\ \cline{4-10} 
\multicolumn{1}{|c|}{{\color[HTML]{340096} }}                                            & \multicolumn{1}{c|}{\multirow{-2}{*}{171M}}                                 & \multicolumn{1}{c|}{}                                          & \multicolumn{1}{c|}{attention}                                              & \multicolumn{1}{c|}{14}                                 & \multicolumn{1}{c|}{0.8774}                                 & \multicolumn{1}{c|}{0.9184}                                                         & \multicolumn{1}{c|}{0.9217}                                                      & \multicolumn{1}{c|}{0.8451}                                                      & 03:45:15                                                        \\ \cline{2-2} \cline{4-10} 
\multicolumn{1}{|c|}{{\color[HTML]{340096} }}                                            & \multicolumn{1}{c|}{{\color[HTML]{340096} }}                                & \multicolumn{1}{c|}{}                                          & \multicolumn{1}{c|}{concat}                                                 & \multicolumn{1}{c|}{15}                                 & \multicolumn{1}{c|}{0.9422}                                 & \multicolumn{1}{c|}{0.9762}                                                         & \multicolumn{1}{c|}{0.9781}                                                      & \multicolumn{1}{c|}{0.9411}                                                      & 03:52:36                                                        \\ \cline{4-10} 
\multicolumn{1}{|c|}{{\color[HTML]{340096} }}                                            & \multicolumn{1}{c|}{{\color[HTML]{340096} }}                                & \multicolumn{1}{c|}{\multirow{-4}{*}{CE}}                      & \multicolumn{1}{c|}{mean}                                                   & \multicolumn{1}{c|}{16}                                 & \multicolumn{1}{c|}{0.9713}                                 & \multicolumn{1}{c|}{0.9857}                                                         & \multicolumn{1}{c|}{0.9861}                                                      & \multicolumn{1}{c|}{0.9670}                                                      & 01:02:16                                                        \\ \cline{3-10} 
\multicolumn{1}{|c|}{{\color[HTML]{340096} }}                                            & \multicolumn{1}{c|}{{\color[HTML]{340096} }}                                & \multicolumn{1}{c|}{{\color[HTML]{340096} \textbf{CE+SupCon}}} & \multicolumn{1}{c|}{{\color[HTML]{340096} }}                                & \multicolumn{1}{c|}{{\color[HTML]{340096} \textbf{17}}} & \multicolumn{1}{c|}{{\color[HTML]{340096} \textbf{0.9823}}} & \multicolumn{1}{c|}{{\color[HTML]{340096} \textbf{0.9911}}}                         & \multicolumn{1}{c|}{{\color[HTML]{340096} \textbf{0.9916}}}                      & \multicolumn{1}{c|}{{\color[HTML]{340096} \textbf{0.9802}}}                      & {\color[HTML]{340096} \textbf{01:15:56}}                        \\ \cline{3-3} \cline{5-10} 
\multicolumn{1}{|c|}{{\color[HTML]{340096} }}                                            & \multicolumn{1}{c|}{\multirow{-4}{*}{{\color[HTML]{340096} \textbf{169M}}}} & \multicolumn{1}{c|}{ITC+CE}                                   & \multicolumn{1}{c|}{{\color[HTML]{340096} }}                                & \multicolumn{1}{c|}{18}                                 & \multicolumn{1}{c|}{0.9463}                                 & \multicolumn{1}{c|}{0.9744}                                                         & \multicolumn{1}{c|}{0.9760}                                                      & \multicolumn{1}{c|}{0.9466}                                                      & 01:17:20                                                        \\ \cline{2-3} \cline{5-10} 
\multicolumn{1}{|c|}{{\color[HTML]{340096} }}                                            & \multicolumn{1}{c|}{}                                                       & \multicolumn{1}{c|}{ITC+ITM+CE}                               & \multicolumn{1}{c|}{{\color[HTML]{340096} }}                                & \multicolumn{1}{c|}{10}                                 & \multicolumn{1}{c|}{0.8456}                                 & \multicolumn{1}{c|}{0.9010}                                                         & \multicolumn{1}{c|}{0.8995}                                                      & \multicolumn{1}{c|}{0.8443}                                                      & 01:07:17                                                        \\ \cline{3-3} \cline{5-10} 
\multicolumn{1}{|c|}{{\color[HTML]{340096} }}                                            & \multicolumn{1}{c|}{}                                                       & \multicolumn{1}{c|}{BLIP2+CE}                                  & \multicolumn{1}{c|}{{\color[HTML]{340096} }}                                & \multicolumn{1}{c|}{11}                                 & \multicolumn{1}{c|}{0.9101}                                 & \multicolumn{1}{c|}{0.9620}                                                         & \multicolumn{1}{c|}{0.9644}                                                      & \multicolumn{1}{c|}{0.9128}                                                      & 01:14:19                                                        \\ \cline{3-3} \cline{5-10} 
\multicolumn{1}{|c|}{\multirow{-9}{*}{{\color[HTML]{340096} \textbf{DeCLUTR-ViT}}}}      & \multicolumn{1}{c|}{\multirow{-3}{*}{307M}}                                 & \multicolumn{1}{c|}{BLIP2+CE+SupCon}                              & \multicolumn{1}{c|}{\multirow{-5}{*}{{\color[HTML]{340096} \textbf{mean}}}} & \multicolumn{1}{c|}{13}                                 & \multicolumn{1}{c|}{0.9450}                                 & \multicolumn{1}{c|}{0.9702}                                                         & \multicolumn{1}{c|}{0.9722}                                                      & \multicolumn{1}{c|}{0.9420}                                                      & 01:30:57                                                        \\ \hline
\end{tabular}
}
\caption{Performance metrics (Balanced Accuracy, Weighted-F1, Micro-F1, and Macro-F1) and computational details for text, vision, and multimodal classifier baselines trained on the South region dataset. Pre-training strategies—ITC, ITC+ITM, and BLIP2—are applied to DeCLUTR-small and ViT-base-patch16 models to align text and images from the same advertisement. Fine-tuning is then conducted for the vendor identification task on the South region dataset, with classifiers optimized using CE, CE+Triplet, and CE+SupCon loss objectives.}
\label{tab:all_vendor_identification_results}
\end{table*}

\begin{table*}
\centering
\resizebox{\linewidth}{!}{%
\begin{tabular}{|c cccccc|}
\hline
\multicolumn{1}{|c|}{\textbf{Loss}}                             & \multicolumn{1}{c|}{\textbf{South}}        & \multicolumn{1}{c|}{\textbf{Midwest}}      & \multicolumn{1}{c|}{\textbf{West}}         & \multicolumn{1}{c|}{\textbf{Northeast}}    & \multicolumn{1}{c|}{\textbf{OOD Avg.}}     & \textbf{ZS Avg.} \\ \hline
\multicolumn{7}{|c|}{\textbf{MRR@10}}                                                                                                                                                                                                                                                                                                                                                                                                                               \\ \hline
\multicolumn{1}{|c|}{\textbf{Pre-trained}}                      & \multicolumn{1}{c|}{0.2248 ± 0.30}                                 & \multicolumn{1}{c|}{0.2866 ± 0.36}                                 & \multicolumn{1}{c|}{0.3479 ± 0.41}                                 & \multicolumn{1}{c|}{0.3385 ± 0.38}                                 & \multicolumn{1}{c|}{-}                                             & 0.2995 ± 0.36                            \\ \hline
\multicolumn{1}{|c|}{\textbf{CE}}                               & \multicolumn{1}{c|}{0.7445 ± 0.39}                                 & \multicolumn{1}{c|}{0.5703 ± 0.46}                                 & \multicolumn{1}{c|}{0.6394 ± 0.45}                                 & \multicolumn{1}{c|}{0.5862 ± 0.48}                                 & \multicolumn{1}{c|}{0.5986 ± 0.46}                                 & -                                        \\ \hline
\multicolumn{1}{|c|}{\textbf{Triplet}}                          & \multicolumn{1}{c|}{0.4282 ± 0.45}                                 & \multicolumn{1}{c|}{0.3200 ± 0.43}                                 & \multicolumn{1}{c|}{0.4074 ± 0.46}                                 & \multicolumn{1}{c|}{0.3503 ± 0.45}                                 & \multicolumn{1}{c|}{0.3592 ± 0.45}                                 & -                                        \\ \hline
\multicolumn{1}{|c|}{\textbf{SupCon}}                           & \multicolumn{1}{c|}{0.8829 ± 0.29}                                 & \multicolumn{1}{c|}{0.7636 ± 0.39}                                 & \multicolumn{1}{c|}{0.8331 ± 0.35}                                 & \multicolumn{1}{c|}{0.7520 ± 0.42}                                 & \multicolumn{1}{c|}{0.7829 ± 0.39}                                 & -                                        \\ \hline
\multicolumn{1}{|c|}{\textbf{CE+Triplet}}                       & \multicolumn{1}{c|}{0.8891 ± 0.28}                                 & \multicolumn{1}{c|}{0.6410 ± 0.45}                                 & \multicolumn{1}{c|}{0.6969 ± 0.43}                                 & \multicolumn{1}{c|}{0.6561 ± 0.45}                                 & \multicolumn{1}{c|}{0.6647 ± 0.44}                                 & -                                        \\ \hline
\multicolumn{1}{|c|}{{\color[HTML]{340096} \textbf{CE+SupCon}}} & \multicolumn{1}{c|}{{\color[HTML]{340096} \textbf{0.9290 ± 0.23}}} & \multicolumn{1}{c|}{{\color[HTML]{340096} \textbf{0.7716 ± 0.38}}} & \multicolumn{1}{c|}{{\color[HTML]{340096} \textbf{0.8145 ± 0.36}}} & \multicolumn{1}{c|}{{\color[HTML]{340096} \textbf{0.7449 ± 0.42}}} & \multicolumn{1}{c|}{{\color[HTML]{340096} \textbf{0.7770 ± 0.39}}} & {\color[HTML]{340096} -}                 \\ \hline
\multicolumn{7}{|c|}{\textbf{R-Precision@X}}                                                                                                                                                                                                                                                                                                                                                                                                                        \\ \hline
\multicolumn{1}{|c|}{\textbf{Pre-trained}}                      & \multicolumn{1}{c|}{0.3265 ± 0.47}                                 & \multicolumn{1}{c|}{0.3943 ± 0.49}                                 & \multicolumn{1}{c|}{0.3139 ± 0.46}                                 & \multicolumn{1}{c|}{0.4037 ± 0.49}                                 & \multicolumn{1}{c|}{-}                                             & 0.3596 ± 0.48                            \\ \hline
\multicolumn{1}{|c|}{\textbf{CE}}                               & \multicolumn{1}{c|}{0.5557 ± 0.36}                                 & \multicolumn{1}{c|}{0.4596 ± 0.40}                                 & \multicolumn{1}{c|}{0.5842 ± 0.41}                                 & \multicolumn{1}{c|}{0.4944 ± 0.43}                                 & \multicolumn{1}{c|}{0.5127 ± 0.41}                                 & -                                        \\ \hline
\multicolumn{1}{|c|}{\textbf{Triplet}}                          & \multicolumn{1}{c|}{0.3200 ± 0.34}                                 & \multicolumn{1}{c|}{0.2443 ± 0.33}                                 & \multicolumn{1}{c|}{0.3365 ± 0.38}                                 & \multicolumn{1}{c|}{0.3032 ± 0.38}                                 & \multicolumn{1}{c|}{0.2947 ± 0.36}                                 & -                                        \\ \hline
\multicolumn{1}{|c|}{\textbf{SupCon}}                           & \multicolumn{1}{c|}{0.7673 ± 0.29}                                 & \multicolumn{1}{c|}{0.6346 ± 0.37}                                 & \multicolumn{1}{c|}{0.7612 ± 0.35}                                 & \multicolumn{1}{c|}{0.6707 ± 0.41}                                 & \multicolumn{1}{c|}{0.6888 ± 0.38}                                 & -                                        \\ \hline
\multicolumn{1}{|c|}{\textbf{CE+Triplet}}                       & \multicolumn{1}{c|}{0.8055 ± 0.30}                                 & \multicolumn{1}{c|}{0.5000 ± 0.40}                                 & \multicolumn{1}{c|}{0.5890 ± 0.4}                                  & \multicolumn{1}{c|}{0.5410 ± 0.42}                                 & \multicolumn{1}{c|}{0.5433 ± 0.41}                                 & -                                        \\ \hline
\multicolumn{1}{|c|}{{\color[HTML]{340096} \textbf{CE+SupCon}}} & \multicolumn{1}{c|}{{\color[HTML]{340096} \textbf{0.8706 ± 0.24}}} & \multicolumn{1}{c|}{{\color[HTML]{340096} \textbf{0.6264 ± 0.38}}} & \multicolumn{1}{c|}{{\color[HTML]{340096} \textbf{0.7339 ± 0.37}}} & \multicolumn{1}{c|}{{\color[HTML]{340096} \textbf{0.6699 ± 0.41}}} & \multicolumn{1}{c|}{{\color[HTML]{340096} \textbf{0.6767 ± 0.39}}} & {\color[HTML]{340096} -}                 \\ \hline
\multicolumn{7}{|c|}{\textbf{Macro-F1@X}}                                                                                                                                                                                                                                                                                                                                                                                                                           \\ \hline
\multicolumn{1}{|c|}{\textbf{Pre-trained}}                      & \multicolumn{1}{c|}{0.2224 ± 0.30}                                 & \multicolumn{1}{c|}{0.2804 ± 0.36}                                 & \multicolumn{1}{c|}{0.2731 ± 0.36}                                 & \multicolumn{1}{c|}{0.3801 ± 0.39}                                 & \multicolumn{1}{c|}{-}                                             & 0.2890 ± 0.37                            \\ \hline
\multicolumn{1}{|c|}{\textbf{CE}}                               & \multicolumn{1}{c|}{0.6098 ± 0.35}                                 & \multicolumn{1}{c|}{0.4760 ± 0.38}                                 & \multicolumn{1}{c|}{0.6123 ± 0.35}                                 & \multicolumn{1}{c|}{0.5042 ± 0.42}                                 & \multicolumn{1}{c|}{0.5308 ± 0.38}                                 & -                                        \\ \hline
\multicolumn{1}{|c|}{\textbf{Triplet}}                          & \multicolumn{1}{c|}{0.4135 ± 0.37}                                 & \multicolumn{1}{c|}{0.2892 ± 0.35}                                 & \multicolumn{1}{c|}{0.4337 ± 0.35}                                 & \multicolumn{1}{c|}{0.3121 ± 0.39}                                 & \multicolumn{1}{c|}{0.3450 ± 0.36}                                 & -                                        \\ \hline
\multicolumn{1}{|c|}{\textbf{SupCon}}                           & \multicolumn{1}{c|}{0.8157 ± 0.27}                                 & \multicolumn{1}{c|}{0.6333 ± 0.36}                                 & \multicolumn{1}{c|}{0.7408 ± 0.31}                                 & \multicolumn{1}{c|}{0.6950 ± 0.39}                                 & \multicolumn{1}{c|}{0.6897 ± 0.35}                                 & -                                        \\ \hline
\multicolumn{1}{|c|}{\textbf{CE+Triplet}}                       & \multicolumn{1}{c|}{0.8680 ± 0.26}                                 & \multicolumn{1}{c|}{0.5198 ± 0.39}                                 & \multicolumn{1}{c|}{0.5789 ± 0.35}                                 & \multicolumn{1}{c|}{0.5612 ± 0.41}                                 & \multicolumn{1}{c|}{0.5533 ± 0.38}                                 & -                                        \\ \hline
\multicolumn{1}{|c|}{{\color[HTML]{340096} \textbf{CE+SupCon}}} & \multicolumn{1}{c|}{{\color[HTML]{340096} \textbf{0.9102 ± 0.21}}} & \multicolumn{1}{c|}{{\color[HTML]{340096} \textbf{0.6162 ± 0.37}}} & \multicolumn{1}{c|}{{\color[HTML]{340096} \textbf{0.7169 ± 0.33}}} & \multicolumn{1}{c|}{{\color[HTML]{340096} \textbf{0.6879 ± 0.40}}} & \multicolumn{1}{c|}{{\color[HTML]{340096} \textbf{0.6737 ± 0.37}}} & {\color[HTML]{340096} -}                 \\ \hline
\end{tabular}
}
\caption{Comparison of text-to-text retrieval performance for the text-only benchmark, DeCLUTR-small backbone, with different objectives (losses), evaluated across MRR@10, R-Precision@X, and Macro-F1@X metrics.
}
\label{tab:text_retrieval_results}
\end{table*}

\subsubsection{Vision-only Modality}
The vision baselines in Table \ref{tab:all_vendor_identification_results} highlight that ResNet-50 with CE loss achieves the highest performance among classifiers for the vendor identification task. However, retrieval results in Table \ref{tab:image-baselines-retrieval} show that, despite slightly lower classification performance, the ViT-base-patch16 backbone consistently outperforms other models on both training and OOD datasets for the image-to-image retrieval task. Given our research's dual objectives of vendor identification and verification, we establish the ViT-base-patch16 backbone as the most suitable choice for further experiments. Consistent with the text-only modality findings, Table \ref{tab:vit-retrieval-results} indicates that using a joint objective with CE+SupCon loss delivers the best results across all vision-only baselines, reinforcing its effectiveness in both classification and retrieval tasks.

\subsubsection{Multimodal Modality}
The multimodal baselines in Table \ref{tab:all_vendor_identification_results} consistently outperform their text-only and vision-only counterparts on the classification task. Among the fusion techniques explored, mean pooling proves to be the most effective for merging text and vision representations. However, despite pre-training on text-image alignment tasks, the fine-tuned multimodal baselines show limited vendor identification and verification performance. Table \ref{tab:alignment_retreival_results} highlights the text-to-image retrieval performance of these pre-trained baselines, where, given a query text ad, the goal is to retrieve its associated images from the original ad. The underperformance of these models stems from the lack of semantic alignment in escort ads, as the visual content often fails to correspond meaningfully to the accompanying text. In contrast, as demonstrated in Tables \ref{tab:all_vendor_identification_results}, \ref{tab:multimodal_text_retrieval}, \ref{tab:multimodal_vision_retrieval}, and \ref{tab:multimodal_retrieval}, the DeCLUTR-ViT backbone trained end-to-end with the CE+SupCon objective (without pre-training) achieves superior performance across all tasks, reinforcing the effectiveness of end-to-end training for multimodal AA in this domain.

\begin{table*}[h]
\centering
\resizebox{0.90\linewidth}{!}{%
\begin{tabular}{|c ccccl|}
\hline
\multicolumn{1}{|c|}{\textbf{Loss}}                                    & \multicolumn{1}{c|}{\textbf{South}}        & \multicolumn{1}{c|}{\textbf{Midwest}}      & \multicolumn{1}{c|}{\textbf{West}}         & \multicolumn{1}{c|}{\textbf{Northeast}}    & \multicolumn{1}{c|}{\textbf{OOD Avg.}} \\ \hline
\multicolumn{6}{|c|}{\textbf{MRR@10}}                                                                                                                                                                                                                                                                                                                                                                                       \\ \hline
\multicolumn{1}{|c|}{\textbf{VGG16}}                                   & \multicolumn{1}{c|}{0.0069 ± 0.05}                                 & \multicolumn{1}{c|}{0.0098 ± 0.07}                                 & \multicolumn{1}{c|}{0.0491 ± 0.19}                                 & \multicolumn{1}{c|}{0.0172 ± 0.1}                                  & 0.0254 ± 0.12                                                  \\ \hline
\multicolumn{1}{|c|}{\textbf{ResNet50}}                                & \multicolumn{1}{c|}{0.1026 ± 0.22}                                 & \multicolumn{1}{c|}{0.1569 ± 0.29}                                 & \multicolumn{1}{c|}{0.221 ± 0.35}                                  & \multicolumn{1}{c|}{0.125 ± 0.26}                                  & 0.1676 ± 0.30                                                  \\ \hline
\multicolumn{1}{|c|}{\textbf{Densenet121}}                             & \multicolumn{1}{c|}{0.218 ± 0.32}                                  & \multicolumn{1}{c|}{0.2465 ± 0.35}                                 & \multicolumn{1}{c|}{0.2669 ± 0.37}                                 & \multicolumn{1}{c|}{0.1889 ± 0.32}                                 & 0.2341 ± 0.35                                                  \\ \hline
\multicolumn{1}{|c|}{\textbf{InceptionNetV3}}                          & \multicolumn{1}{l|}{0.0477 ± 0.15}                                 & \multicolumn{1}{l|}{0.0583 ± 0.19}                                 & \multicolumn{1}{l|}{0.0684 ± 0.2}                                  & \multicolumn{1}{l|}{0.0625 ± 0.19}                                 & 0.0631 ± 0.19                                                  \\ \hline
\multicolumn{1}{|l|}{\textbf{EfficientNetV2}}                          & \multicolumn{1}{c|}{0.2305 ± 0.32}                                 & \multicolumn{1}{c|}{0.2468 ± 0.35}                                 & \multicolumn{1}{c|}{0.2523 ± 0.36}                                 & \multicolumn{1}{c|}{0.2276 ± 0.35}                                 & 0.2422 ± 0.35                                                  \\ \hline
\multicolumn{1}{|l|}{\textbf{ConvNext-small}}                          & \multicolumn{1}{l|}{0.0588 ± 0.17}                                 & \multicolumn{1}{l|}{0.0851 ± 0.22}                                 & \multicolumn{1}{l|}{0.0854 ± 0.23}                                 & \multicolumn{1}{l|}{0.0917 ± 0.24}                                 & 0.0874 ± 0.23                                                  \\ \hline
\multicolumn{1}{|c|}{{\color[HTML]{340096} \textbf{ViT-base-patch16}}} & \multicolumn{1}{l|}{{\color[HTML]{340096} \textbf{0.2587 ± 0.33}}} & \multicolumn{1}{l|}{{\color[HTML]{340096} \textbf{0.2854 ± 0.37}}} & \multicolumn{1}{l|}{{\color[HTML]{340096} \textbf{0.3019 ± 0.39}}} & \multicolumn{1}{l|}{{\color[HTML]{340096} \textbf{0.2597 ± 0.36}}} & {\color[HTML]{340096} \textbf{0.2823 ± 0.37}}                  \\ \hline
\multicolumn{6}{|c|}{\textbf{R-Precision@X}}                                                                                                                                                                                                                                                                                                                                                                                \\ \hline
\multicolumn{1}{|c|}{\textbf{VGG16}}                                   & \multicolumn{1}{c|}{0.0063 ± 0.03}                                 & \multicolumn{1}{c|}{0.0074 ± 0.03}                                 & \multicolumn{1}{c|}{0.0165 ± 0.05}                                 & \multicolumn{1}{c|}{0.0139 ± 0.06}                                 & 0.0126 ± 0.05                                                  \\ \hline
\multicolumn{1}{|c|}{\textbf{ResNet50}}                                & \multicolumn{1}{c|}{0.0267 ± 0.05}                                 & \multicolumn{1}{c|}{0.0415 ± 0.09}                                 & \multicolumn{1}{c|}{0.0599 ± 0.1}                                  & \multicolumn{1}{c|}{0.0452 ± 0.09}                                 & 0.0489 ± 0.09                                                  \\ \hline
\multicolumn{1}{|c|}{\textbf{Densenet121}}                             & \multicolumn{1}{c|}{0.0413 ± 0.08}                                 & \multicolumn{1}{c|}{0.0618 ± 0.11}                                 & \multicolumn{1}{c|}{0.0849 ± 0.11}                                 & \multicolumn{1}{c|}{0.0671 ± 0.11}                                 & 0.0713 ± 0.11                                                  \\ \hline
\multicolumn{1}{|c|}{\textbf{InceptionNetV3}}                          & \multicolumn{1}{l|}{0.0084 ± 0.02}                                 & \multicolumn{1}{l|}{0.0176 ± 0.07}                                 & \multicolumn{1}{l|}{0.0224 ± 0.06}                                 & \multicolumn{1}{l|}{0.0143 ± 0.04}                                 & 0.0181 ± 0.06                                                  \\ \hline
\multicolumn{1}{|l|}{\textbf{EfficientNetV2}}                          & \multicolumn{1}{c|}{0.0417 ± 0.07}                                 & \multicolumn{1}{c|}{0.0609 ± 0.1}                                  & \multicolumn{1}{c|}{0.0752 ± 0.11}                                 & \multicolumn{1}{c|}{0.0692 ± 0.11}                                 & 0.0684 ± 0.11                                                  \\ \hline
\multicolumn{1}{|l|}{\textbf{ConvNext-small}}                          & \multicolumn{1}{l|}{0.0157 ± 0.04}                                 & \multicolumn{1}{c|}{0.026 ± 0.06}                                  & \multicolumn{1}{l|}{0.0299 ± 0.07}                                 & \multicolumn{1}{l|}{0.0291 ± 0.06}                                 & 0.0283 ± 0.06                                                  \\ \hline
\multicolumn{1}{|c|}{{\color[HTML]{340096} \textbf{ViT-base-patch16}}} & \multicolumn{1}{l|}{{\color[HTML]{340096} \textbf{0.0459 ± 0.07}}} & \multicolumn{1}{l|}{{\color[HTML]{340096} \textbf{0.0645 ± 0.11}}} & \multicolumn{1}{l|}{{\color[HTML]{340096} \textbf{0.0781 ± 0.11}}} & \multicolumn{1}{l|}{{\color[HTML]{340096} \textbf{0.078 ± 0.13}}}  & {\color[HTML]{340096} \textbf{0.0735 ± 0.12}}                  \\ \hline
\multicolumn{6}{|c|}{\textbf{Macro-F1@X}}                                                                                                                                                                                                                                                                                                                                                                                   \\ \hline
\multicolumn{1}{|c|}{\textbf{VGG16}}                                   & \multicolumn{1}{c|}{0.0091 ± 0.03}                                 & \multicolumn{1}{c|}{0.0151 ± 0.04}                                 & \multicolumn{1}{c|}{0.0171 ± 0.06}                                 & \multicolumn{1}{c|}{0.0158 ± 0.05}                                 & 0.0160 ± 0.05                                                  \\ \hline
\multicolumn{1}{|c|}{\textbf{ResNet50}}                                & \multicolumn{1}{l|}{0.0276 ± 0.06}                                 & \multicolumn{1}{l|}{0.0407 ± 0.08}                                 & \multicolumn{1}{l|}{0.0565 ± 0.1}                                  & \multicolumn{1}{l|}{0.0468 ± 0.09}                                 & 0.0479 ± 0.09                                                  \\ \hline
\multicolumn{1}{|c|}{\textbf{Densenet121}}                             & \multicolumn{1}{l|}{0.04 ± 0.07}                                   & \multicolumn{1}{l|}{0.0535 ± 0.09}                                 & \multicolumn{1}{l|}{0.0823 ± 0.12}                                 & \multicolumn{1}{l|}{0.0641 ± 0.11}                                 & 0.0666 ± 0.11                                                  \\ \hline
\multicolumn{1}{|c|}{\textbf{InceptionNetV3}}                          & \multicolumn{1}{c|}{0.0083 ± 0.02}                                 & \multicolumn{1}{c|}{0.0154 ± 0.05}                                 & \multicolumn{1}{c|}{0.0215 ± 0.06}                                 & \multicolumn{1}{c|}{0.0147 ± 0.04}                                 & 0.0172 ± 0.05                                                  \\ \hline
\multicolumn{1}{|l|}{\textbf{EfficientNetV2}}                          & \multicolumn{1}{c|}{0.042 ± 0.07}                                  & \multicolumn{1}{c|}{0.0546 ± 0.09}                                 & \multicolumn{1}{c|}{0.0764 ± 0.12}                                 & \multicolumn{1}{c|}{0.0648 ± 0.1}                                  & 0.0653 ± 0.10                                                  \\ \hline
\multicolumn{1}{|l|}{\textbf{ConvNext-small}}                          & \multicolumn{1}{c|}{0.0159 ± 0.04}                                 & \multicolumn{1}{c|}{0.028 ± 0.06}                                  & \multicolumn{1}{c|}{\textit{0.0312 ± 0.07}}                        & \multicolumn{1}{c|}{0.0301 ± 0.06}                                 & 0.0298 ± 0.06                                                  \\ \hline
\multicolumn{1}{|c|}{{\color[HTML]{340096} \textbf{ViT-base-patch16}}} & \multicolumn{1}{c|}{{\color[HTML]{340096} \textbf{0.0436 ± 0.07}}} & \multicolumn{1}{c|}{{\color[HTML]{340096} \textbf{0.0574 ± 0.09}}} & \multicolumn{1}{c|}{{\color[HTML]{340096} \textbf{0.077 ± 0.12}}}  & \multicolumn{1}{c|}{{\color[HTML]{340096} \textbf{0.0727 ± 0.11}}} & {\color[HTML]{340096} \textbf{0.0690 ± 0.11}}                  \\ \hline
\end{tabular}
}
\caption{Comparison of image-to-image retrieval performance for the vision-baselines trained on south region image ads with CE loss, evaluated on MRR@10, R-Precision@X, and Macro-F1@X metrics}
\label{tab:image-baselines-retrieval}
\end{table*}

\begin{table*}[h]
\resizebox{\linewidth}{!}{%
\centering
\begin{tabular}{|c cccccc|}
\hline
\multicolumn{1}{|c|}{\textbf{Loss}}                             & \multicolumn{1}{c|}{\textbf{South}}        & \multicolumn{1}{c|}{\textbf{Midwest}}      & \multicolumn{1}{c|}{\textbf{West}}         & \multicolumn{1}{c|}{\textbf{Northeast}}    & \multicolumn{1}{c|}{\textbf{OOD Avg.}}     & \textbf{ZS Avg.} \\ \hline
\multicolumn{7}{|c|}{\textbf{MRR@10}}                                                                                                                                                                                                                                                                                                                                                                                                                               \\ \hline
\multicolumn{1}{|c|}{\textbf{Pre-trained}}                      & \multicolumn{1}{c|}{0.2286 ± 0.32}                                 & \multicolumn{1}{c|}{0.2432 ± 0.35}                                 & \multicolumn{1}{c|}{0.2517 ± 0.36}                                 & \multicolumn{1}{c|}{0.2242 ± 0.35}                                 & \multicolumn{1}{c|}{-}                                             & 0.2369 ± 0.35                            \\ \hline
\multicolumn{1}{|c|}{\textbf{CE}}                               & \multicolumn{1}{c|}{0.2587 ± 0.33}                                 & \multicolumn{1}{c|}{0.2854 ± 0.37}                                 & \multicolumn{1}{c|}{0.3019 ± 0.39}                                 & \multicolumn{1}{c|}{0.2597 ± 0.36}                                 & \multicolumn{1}{c|}{0.2823 ± 0.37}                                 & -                                        \\ \hline
\multicolumn{1}{|c|}{\textbf{SupCon}}                           & \multicolumn{1}{c|}{0.0010 ± 0.03}                                 & \multicolumn{1}{c|}{0.0013 ± 0.03}                                 & \multicolumn{1}{c|}{0.0031 ± 0.03}                                 & \multicolumn{1}{c|}{0.0079 ± 0.08}                                 & \multicolumn{1}{c|}{0.0041 ± 0.05}                                 & -                                        \\ \hline
\multicolumn{1}{|c|}{\textbf{Triplet}}                          & \multicolumn{1}{c|}{0.0010 ± 0.03}                                 & \multicolumn{1}{c|}{0.0016 ± 0.04}                                 & \multicolumn{1}{c|}{0.0035 ± 0.06}                                 & \multicolumn{1}{c|}{0.0054 ± 0.07}                                 & \multicolumn{1}{c|}{0.0035 ± 0.06}                                 & -                                        \\ \hline
\multicolumn{1}{|c|}{\textbf{CE+Triplet}}                       & \multicolumn{1}{c|}{0.2760 ± 0.35}                                 & \multicolumn{1}{c|}{0.3242 ± 0.39}                                 & \multicolumn{1}{c|}{0.366 ± 0.41}                                  & \multicolumn{1}{c|}{0.3322 ± 0.39}                                 & \multicolumn{1}{c|}{0.3408 ± 0.40}                                 & -                                        \\ \hline
\multicolumn{1}{|c|}{{\color[HTML]{340096} \textbf{CE+SupCon}}} & \multicolumn{1}{c|}{{\color[HTML]{340096} \textbf{0.3464 ± 0.37}}} & \multicolumn{1}{c|}{{\color[HTML]{340096} \textbf{0.3749 ± 0.40}}} & \multicolumn{1}{c|}{{\color[HTML]{340096} \textbf{0.4049 ± 0.42}}} & \multicolumn{1}{c|}{{\color[HTML]{340096} \textbf{0.4330 ± 0.42}}} & \multicolumn{1}{c|}{{\color[HTML]{340096} \textbf{0.4041 ± 0.41}}} & {\color[HTML]{340096} \textbf{-}}        \\ \hline
\multicolumn{7}{|c|}{\textbf{R-Precision@X}}                                                                                                                                                                                                                                                                                                                                                                                                                        \\ \hline
\multicolumn{1}{|c|}{\textbf{Pre-trained}}                      & \multicolumn{1}{c|}{0.0420 ± 0.07}                                 & \multicolumn{1}{c|}{0.0593 ± 0.10}                                 & \multicolumn{1}{c|}{0.0754 ± 0.11}                                 & \multicolumn{1}{c|}{0.0691 ± 0.11}                                 & \multicolumn{1}{c|}{-}                                             & 0.0615 ± 0.10                            \\ \hline
\multicolumn{1}{|c|}{\textbf{CE}}                               & \multicolumn{1}{c|}{0.0459 ± 0.07}                                 & \multicolumn{1}{c|}{0.0645 ± 0.11}                                 & \multicolumn{1}{c|}{0.0781 ± 0.11}                                 & \multicolumn{1}{c|}{0.078 ± 0.13}                                  & \multicolumn{1}{c|}{0.0735 ± 0.12}                                 & -                                        \\ \hline
\multicolumn{1}{|c|}{\textbf{SupCon}}                           & \multicolumn{1}{c|}{0.0010 ± 0.01}                                 & \multicolumn{1}{c|}{0.0018 ± 0.01}                                 & \multicolumn{1}{c|}{0.0028 ± 0.01}                                 & \multicolumn{1}{c|}{0.0028 ± 0.02}                                 & \multicolumn{1}{c|}{0.0025 ± 0.01}                                 & -                                        \\ \hline
\multicolumn{1}{|c|}{\textbf{Triplet}}                          & \multicolumn{1}{c|}{0.0009 ± 0.01}                                 & \multicolumn{1}{c|}{0.0007 ± 0.01}                                 & \multicolumn{1}{c|}{0.0017 ± 0.02}                                 & \multicolumn{1}{c|}{0.003 ± 0.02}                                  & \multicolumn{1}{l|}{0.0018 ± 0.02}                                 & -                                        \\ \hline
\multicolumn{1}{|c|}{\textbf{CE+Triplet}}                       & \multicolumn{1}{c|}{0.0824 ± 0.14}                                 & \multicolumn{1}{c|}{0.0963 ± 0.15}                                 & \multicolumn{1}{c|}{0.139 ± 0.19}                                  & \multicolumn{1}{c|}{0.1281 ± 0.17}                                 & \multicolumn{1}{c|}{0.1211 ± 0.17}                                 & -                                        \\ \hline
\multicolumn{1}{|c|}{{\color[HTML]{340096} \textbf{CE+SupCon}}} & \multicolumn{1}{c|}{{\color[HTML]{340096} \textbf{0.1064 ± 0.16}}} & \multicolumn{1}{c|}{{\color[HTML]{340096} \textbf{0.1095 ± 0.16}}} & \multicolumn{1}{c|}{{\color[HTML]{340096} \textbf{0.1519 ± 0.20}}} & \multicolumn{1}{c|}{{\color[HTML]{340096} \textbf{0.1685 ± 0.21}}} & \multicolumn{1}{c|}{{\color[HTML]{340096} \textbf{0.1433 ± 0.19}}} & {\color[HTML]{340096} \textbf{-}}        \\ \hline
\multicolumn{7}{|c|}{\textbf{Macro-F1@X}}                                                                                                                                                                                                                                                                                                                                                                                                                           \\ \hline
\multicolumn{1}{|c|}{\textbf{Pre-trained}}                      & \multicolumn{1}{c|}{0.0421 ± 0.07}                                 & \multicolumn{1}{c|}{0.0539 ± 0.09}                                 & \multicolumn{1}{c|}{0.0767 ± 0.12}                                 & \multicolumn{1}{c|}{0.0647 ± 0.1}                                  & \multicolumn{1}{c|}{-}                                             & 0.0594 ± 0.10                            \\ \hline
\multicolumn{1}{|c|}{\textbf{CE}}                               & \multicolumn{1}{c|}{0.0436 ± 0.07}                                 & \multicolumn{1}{c|}{0.0574 ± 0.09}                                 & \multicolumn{1}{c|}{0.077 ± 0.12}                                  & \multicolumn{1}{c|}{0.0727 ± 0.11}                                 & \multicolumn{1}{c|}{0.0690 ± 0.11}                                 & -                                        \\ \hline
\multicolumn{1}{|c|}{\textbf{SupCon}}                           & \multicolumn{1}{c|}{0.0015 ± 0.01}                                 & \multicolumn{1}{c|}{0.0041 ± 0.01}                                 & \multicolumn{1}{c|}{0.0043 ± 0.02}                                 & \multicolumn{1}{c|}{0.0034 ± 0.02}                                 & \multicolumn{1}{c|}{0.0039 ± 0.02}                                 & -                                        \\ \hline
\multicolumn{1}{|c|}{\textbf{Triplet}}                          & \multicolumn{1}{c|}{0.0011 ± 0.01}                                 & \multicolumn{1}{c|}{0.0031 ± 0.01}                                 & \multicolumn{1}{c|}{0.0028 ± 0.01}                                 & \multicolumn{1}{c|}{0.0026 ± 0.01}                                 & \multicolumn{1}{c|}{0.0028 ± 0.01}                                 & -                                        \\ \hline
\multicolumn{1}{|c|}{\textbf{CE+Triplet}}                       & \multicolumn{1}{c|}{0.1091 ± 0.20}                                 & \multicolumn{1}{c|}{0.0842 ± 0.14}                                 & \multicolumn{1}{c|}{0.1413 ± 0.2}                                  & \multicolumn{1}{c|}{0.1143 ± 0.17}                                 & \multicolumn{1}{c|}{0.1133 ± 0.17}                                 & -                                        \\ \hline
\multicolumn{1}{|c|}{{\color[HTML]{340096} \textbf{CE+SupCon}}} & \multicolumn{1}{c|}{{\color[HTML]{340096} \textbf{0.1296 ± 0.21}}} & \multicolumn{1}{c|}{{\color[HTML]{340096} \textbf{0.0948 ± 0.14}}} & \multicolumn{1}{c|}{{\color[HTML]{340096} \textbf{0.1460 ± 0.20}}} & \multicolumn{1}{c|}{{\color[HTML]{340096} \textbf{0.1497 ± 0.20}}} & \multicolumn{1}{c|}{{\color[HTML]{340096} \textbf{0.1302 ± 0.18}}} & {\color[HTML]{340096} \textbf{-}}        \\ \hline
\end{tabular}
}
\caption{Comparison of image-to-image retrieval performance for the vision-only benchmark, ViT-base-patch16 backbone, with different objectives (losses), evaluated on MRR@10, R-Precision@X, and Macro-F1@X metrics.}
\label{tab:vit-retrieval-results}
\end{table*}

\begin{table*}[h]
\centering
\resizebox{0.80\linewidth}{!}{%
\begin{tabular}{|c ccccc|}
\hline
\multicolumn{1}{|c|}{\textbf{Loss}}     & \multicolumn{1}{c|}{\textbf{South}} & \multicolumn{1}{c|}{\textbf{Midwest}} & \multicolumn{1}{c|}{\textbf{West}} & \multicolumn{1}{c|}{\textbf{Northeast}} & \textbf{Avg.} \\ \hline
\multicolumn{6}{|c|}{\textbf{Alignment MRR@10}}                                                                                                                                                                                                                                                                                              \\ \hline
\multicolumn{1}{|c|}{\textbf{ITC}}     & \multicolumn{1}{c|}{0.0001 ± 0.01}                          & \multicolumn{1}{c|}{0.0001 ± 0.01}                            & \multicolumn{1}{c|}{0.0003 ± 0.02}                         & \multicolumn{1}{c|}{0.0004 ± 0.02}                              & 0.0002 ± 0.01                         \\ \hline
\multicolumn{1}{|c|}{\textbf{ITC+ITM}} & \multicolumn{1}{c|}{0.0001 ± 0.01}                          & \multicolumn{1}{c|}{0.0001 ± 0.01}                            & \multicolumn{1}{c|}{0.0003 ± 0.02}                         & \multicolumn{1}{c|}{0.0008 ± 0.03}                              & 0.0003 ± 0.02                         \\ \hline
\multicolumn{1}{|c|}{\textbf{BLIP2}}    & \multicolumn{1}{c|}{0.001 ± 0.03}                           & \multicolumn{1}{c|}{0.0027 ± 0.05}                            & \multicolumn{1}{c|}{0.0063 ± 0.08}                         & \multicolumn{1}{c|}{0.0098 ± 0.10}                              & 0.0050 ± 0.07                         \\ \hline
\multicolumn{6}{|c|}{\textbf{Alignment R-Precision@X}}                                                                                                                                                                                                                                                                                       \\ \hline
\multicolumn{1}{|c|}{\textbf{ITC}}     & \multicolumn{1}{c|}{0.0001 ± 0.01}                          & \multicolumn{1}{c|}{0.0002 ± 0.01}                            & \multicolumn{1}{c|}{0.0013 ± 0.03}                         & \multicolumn{1}{c|}{0.0005 ± 0.01}                              & 0.0005 ± 0.01                         \\ \hline
\multicolumn{1}{|c|}{\textbf{ITC+ITM}} & \multicolumn{1}{c|}{0.0002 ± 0.01}                          & \multicolumn{1}{c|}{0.0002 ± 0.01}                            & \multicolumn{1}{c|}{0.0006 ± 0.01}                         & \multicolumn{1}{c|}{0.0007 ± 0.01}                              & 0.0004 ± 0.01                         \\ \hline
\multicolumn{1}{|c|}{\textbf{BLIP2}}    & \multicolumn{1}{c|}{0.0017 ± 0.02}                          & \multicolumn{1}{c|}{0.0049 ± 0.04}                            & \multicolumn{1}{c|}{0.0103 ± 0.06}                         & \multicolumn{1}{c|}{0.0104 ± 0.06}                              & 0.0068 ± 0.05                         \\ \hline
\multicolumn{6}{|c|}{\textbf{Alignment Macro-F1@X}}                                                                                                                                                                                                                                                                                          \\ \hline
\multicolumn{1}{|c|}{\textbf{ITC}}     & \multicolumn{1}{c|}{0.0001 ± 0.01}                          & \multicolumn{1}{c|}{0.0002 ± 0.01}                            & \multicolumn{1}{c|}{0.0013 ± 0.03}                         & \multicolumn{1}{c|}{0.0005 ± 0.01}                              & 0.0005 ± 0.02                         \\ \hline
\multicolumn{1}{|c|}{\textbf{ITC+ITM}} & \multicolumn{1}{c|}{0.0002 ± 0.01}                          & \multicolumn{1}{c|}{0.0002 ± 0.01}                            & \multicolumn{1}{c|}{0.0006 ± 0.01}                         & \multicolumn{1}{c|}{0.0007 ± 0.01}                              & 0.0004 ± 0.01                         \\ \hline
\multicolumn{1}{|c|}{\textbf{BLIP2}}    & \multicolumn{1}{c|}{0.0017 ± 0.02}                          & \multicolumn{1}{c|}{0.0049 ± 0.04}                            & \multicolumn{1}{c|}{0.0103 ± 0.06}                         & \multicolumn{1}{c|}{0.0104 ± 0.06}                              & 0.0068 ± 0.05                         \\ \hline
\end{tabular}
}
\caption{Text-to-Image retrieval results from the multimodal DeCLUTR-ViT backbone pre-trained on the text-image alignment task using CLIP (ITC), ITC+ITM (Image text matching loss), BLIP2 (ITC+ITM+Text generation loss).}
\label{tab:alignment_retreival_results}
\end{table*}

\begin{table*}[h]
\centering
\resizebox{\linewidth}{!}{%
\begin{tabular}{|c c cccccc|}
\hline
\multicolumn{1}{|c|}{\textbf{Backbone}}                                                                                               & \multicolumn{1}{c|}{\textbf{Loss}}                             & \multicolumn{1}{c|}{\textbf{South}}        & \multicolumn{1}{c|}{\textbf{Midwest}}      & \multicolumn{1}{c|}{\textbf{West}}         & \multicolumn{1}{c|}{\textbf{Northeast}}    & \multicolumn{1}{c|}{\textbf{OOD Avg.}}     & \textbf{ZS Avg.} \\ \hline
\multicolumn{8}{|c|}{\textbf{Text MRR@10}}                                                                                                                                                                                                                                                                                                                                                                                                                                                                                                                                                                                         \\ \hline
\multicolumn{1}{|c|}{\textbf{DeCLUTR}}                                                                                                & \multicolumn{1}{c|}{\textbf{CE+SupCon}}                        & \multicolumn{1}{c|}{0.9290 ± 0.23}                                 & \multicolumn{1}{c|}{0.7716 ± 0.38}                                 & \multicolumn{1}{c|}{0.8145 ± 0.36}                                 & \multicolumn{1}{c|}{0.7449 ± 042}                                  & \multicolumn{1}{c|}{0.7770 ± 0.39}                                 &                                          \\ \hline
\multicolumn{1}{|c|}{{\color[HTML]{340096} }}                                                                                         & \multicolumn{1}{c|}{\textbf{CE}}                               & \multicolumn{1}{c|}{0.9850 ± 0.10}                                 & \multicolumn{1}{c|}{0.9693 ± 0.14}                                 & \multicolumn{1}{c|}{0.9900 ± 0.07}                                 & \multicolumn{1}{c|}{0.9778 ± 0.12}                                 & \multicolumn{1}{c|}{0.9790 ± 0.11}                                 & -                                        \\ \cline{2-8} 
\multicolumn{1}{|c|}{\multirow{-2}{*}{{\color[HTML]{340096} \textbf{\begin{tabular}[c]{@{}c@{}}End2End\\ DeCLUTR-ViT\end{tabular}}}}} & \multicolumn{1}{c|}{{\color[HTML]{340096} \textbf{CE+SupCon}}} & \multicolumn{1}{c|}{{\color[HTML]{340096} \textbf{0.9866 ± 0.09}}} & \multicolumn{1}{c|}{{\color[HTML]{340096} \textbf{0.9704 ± 0.14}}} & \multicolumn{1}{c|}{{\color[HTML]{340096} \textbf{0.9932 ± 0.07}}} & \multicolumn{1}{c|}{{\color[HTML]{340096} \textbf{0.9821 ± 0.12}}} & \multicolumn{1}{c|}{{\color[HTML]{340096} \textbf{0.9819 ± 0.11}}} & -                                        \\ \hline
\multicolumn{1}{|c|}{}                                                                                                                & \multicolumn{1}{c|}{\textbf{ITC}}                             & \multicolumn{1}{c|}{0.4097 ± 0.43}                                 & \multicolumn{1}{c|}{0.4289 ± 0.45}                                 & \multicolumn{1}{c|}{0.5404 ± 0.47}                                 & \multicolumn{1}{c|}{0.5034 ± 0.47}                                 & \multicolumn{1}{c|}{-}                                             & 0.4909 ± 0.46                            \\ \cline{2-8} 
\multicolumn{1}{|c|}{}                                                                                                                & \multicolumn{1}{c|}{\textbf{ITC+ITM}}                         & \multicolumn{1}{c|}{0.8192 ± 0.37}                                 & \multicolumn{1}{c|}{0.7990 ± 0.39}                                 & \multicolumn{1}{c|}{0.8600 ± 0.35}                                 & \multicolumn{1}{c|}{0.5914 ± 0.48}                                 & \multicolumn{1}{c|}{-}                                             & 0.7674 ± 0.40                            \\ \cline{2-8} 
\multicolumn{1}{|c|}{}                                                                                                                & \multicolumn{1}{c|}{\textbf{BLIP2}}                            & \multicolumn{1}{c|}{0.7551 ± 0.41}                                 & \multicolumn{1}{c|}{0.7226 ± 0.44}                                 & \multicolumn{1}{c|}{0.8400 ± 0.37}                                 & \multicolumn{1}{c|}{0.5376 ± 0.49}                                 & \multicolumn{1}{c|}{-}                                             & 0.7140 ± 0.43                            \\ \cline{2-8} 
\multicolumn{1}{|c|}{\multirow{-4}{*}{\textbf{\begin{tabular}[c]{@{}c@{}}Aligned\\ DeCLUTR-ViT\end{tabular}}}}                    & \multicolumn{1}{c|}{\textbf{BLIP2-Cond}}                       & \multicolumn{1}{c|}{0.7672 ± 0.41}                                 & \multicolumn{1}{c|}{0.7203 ± 0.44}                                 & \multicolumn{1}{c|}{0.8400 ± 0.37}                                 & \multicolumn{1}{c|}{0.4946 ± 0.49}                                 & \multicolumn{1}{c|}{-}                                             & 0.7055 ± 0.43                            \\ \hline
\multicolumn{1}{|c|}{}                                                                                                                & \multicolumn{1}{c|}{\textbf{ITC+CE}}                          & \multicolumn{1}{c|}{0.8613 ± 0.34}                                 & \multicolumn{1}{c|}{0.6623 ± 0.46}                                 & \multicolumn{1}{c|}{0.8600 ± 0.35}                                 & \multicolumn{1}{c|}{0.6263 ± 0.48}                                 & \multicolumn{1}{c|}{0.7162 ± 0.43}                                 & -                                        \\ \cline{2-8} 
\multicolumn{1}{|c|}{}                                                                                                                & \multicolumn{1}{c|}{\textbf{ITC+ITM+CE}}                      & \multicolumn{1}{c|}{0.4239 ± 0.39}                                 & \multicolumn{1}{c|}{0.2851 ± 0.37}                                 & \multicolumn{1}{c|}{0.3417 ± 0.42}                                 & \multicolumn{1}{c|}{0.3600 ± 0.41}                                 & \multicolumn{1}{c|}{0.3289 ± 0.40}                                 & -                                        \\ \cline{2-8} 
\multicolumn{1}{|c|}{}                                                                                                                & \multicolumn{1}{c|}{\textbf{BLIP2+CE}}                         & \multicolumn{1}{c|}{0.8866 ± 0.30}                                 & \multicolumn{1}{c|}{0.7226 ± 0.44}                                 & \multicolumn{1}{c|}{0.8400 ± 0.37}                                 & \multicolumn{1}{c|}{0.7292 ± 0.44}                                 & \multicolumn{1}{c|}{0.7639 ± 0.42}                                 & -                                        \\ \cline{2-8} 
\multicolumn{1}{|c|}{\multirow{-4}{*}{\textbf{\begin{tabular}[c]{@{}c@{}}Fine-tuned\\ DeCLUTR-ViT\end{tabular}}}}                     & \multicolumn{1}{c|}{\textbf{BLIP2+CE+SupCon}}                  & \multicolumn{1}{c|}{0.8886 ± 0.31}                                 & \multicolumn{1}{c|}{0.7397 ± 0.43}                                 & \multicolumn{1}{c|}{0.8600 ± 0.35}                                 & \multicolumn{1}{c|}{0.7604 ± 0.42}                                 & \multicolumn{1}{c|}{0.7867 ± 0.40}                                 & -                                        \\ \hline
\multicolumn{8}{|c|}{\textbf{Text R-Precision@X}}                                                                                                                                                                                                                                                                                                                                                                                                                                                                                                                                                                                  \\ \hline
\multicolumn{1}{|c|}{\textbf{DeCLUTR}}                                                                                                & \multicolumn{1}{c|}{\textbf{CE+SupCon}}                        & \multicolumn{1}{c|}{0.8706 ± 0.24}                                 & \multicolumn{1}{c|}{0.6264 ± 0.38}                                 & \multicolumn{1}{c|}{0.7339 ± 0.37}                                 & \multicolumn{1}{c|}{0.6699 ± 0.41}                                 & \multicolumn{1}{c|}{0.6767 ± 0.39}                                 &                                          \\ \hline
\multicolumn{1}{|c|}{{\color[HTML]{340096} }}                                                                                         & \multicolumn{1}{c|}{\textbf{CE}}                               & \multicolumn{1}{c|}{0.8687 ± 0.19}                                 & \multicolumn{1}{c|}{0.6500 ± 0.30}                                 & \multicolumn{1}{c|}{0.7934 ± 0.24}                                 & \multicolumn{1}{c|}{0.7300 ± 0.28}                                 & \multicolumn{1}{c|}{0.7245 ± 0.27}                                 & -                                        \\ \cline{2-8} 
\multicolumn{1}{|c|}{\multirow{-2}{*}{{\color[HTML]{340096} \textbf{\begin{tabular}[c]{@{}c@{}}End2End\\ DeCLUTR-ViT\end{tabular}}}}} & \multicolumn{1}{c|}{{\color[HTML]{340096} \textbf{CE+SupCon}}} & \multicolumn{1}{c|}{{\color[HTML]{340096} \textbf{0.9193 ± 0.16}}} & \multicolumn{1}{c|}{{\color[HTML]{340096} \textbf{0.6612 ± 0.31}}} & \multicolumn{1}{c|}{{\color[HTML]{340096} \textbf{0.8008 ± 0.25}}} & \multicolumn{1}{c|}{{\color[HTML]{340096} \textbf{0.7365 ± 0.28}}} & \multicolumn{1}{c|}{{\color[HTML]{340096} \textbf{0.7418 ± 0.28}}} & -                                        \\ \hline
\multicolumn{1}{|c|}{}                                                                                                                & \multicolumn{1}{c|}{\textbf{ITC}}                             & \multicolumn{1}{c|}{0.2337 ± 0.28}                                 & \multicolumn{1}{c|}{0.2936 ± 0.34}                                 & \multicolumn{1}{c|}{0.4035 ± 0.37}                                 & \multicolumn{1}{c|}{0.3779 ± 0.38}                                 & \multicolumn{1}{c|}{-}                                             & 0.3583 ± 0.36                            \\ \cline{2-8} 
\multicolumn{1}{|c|}{}                                                                                                                & \multicolumn{1}{c|}{\textbf{ITC+ITM}}                         & \multicolumn{1}{c|}{0.4964 ± 0.34}                                 & \multicolumn{1}{c|}{0.5679 ± 0.38}                                 & \multicolumn{1}{c|}{0.7093 ± 0.33}                                 & \multicolumn{1}{c|}{0.4818 ± 0.45}                                 & \multicolumn{1}{c|}{-}                                             & 0.5639 ± 0.38                            \\ \cline{2-8} 
\multicolumn{1}{|c|}{}                                                                                                                & \multicolumn{1}{c|}{\textbf{BLIP2}}                            & \multicolumn{1}{c|}{0.4230 ± 0.34}                                 & \multicolumn{1}{c|}{0.5094 ± 0.39}                                 & \multicolumn{1}{c|}{0.6354 ± 0.37}                                 & \multicolumn{1}{c|}{0.3913 ± 0.41}                                 & \multicolumn{1}{c|}{-}                                             & 0.4898 ± 0.38                            \\ \cline{2-8} 
\multicolumn{1}{|c|}{\multirow{-4}{*}{\textbf{\begin{tabular}[c]{@{}c@{}}Aligned\\ DeCLUTR-ViT\end{tabular}}}}                    & \multicolumn{1}{c|}{\textbf{BLIP2-Cond}}                       & \multicolumn{1}{c|}{0.4341 ± 0.35}                                 & \multicolumn{1}{c|}{0.5142 ± 0.39}                                 & \multicolumn{1}{c|}{0.6644 ± 0.36}                                 & \multicolumn{1}{c|}{0.3729 ± 0.42}                                 & \multicolumn{1}{c|}{-}                                             & 0.4964 ± 0.38                            \\ \hline
\multicolumn{1}{|c|}{}                                                                                                                & \multicolumn{1}{c|}{\textbf{ITC+CE}}                          & \multicolumn{1}{c|}{0.6378 ± 0.33}                                 & \multicolumn{1}{c|}{0.4885 ± 0.37}                                 & \multicolumn{1}{c|}{0.6825 ± 0.35}                                 & \multicolumn{1}{c|}{0.3770 ± 0.35}                                 & \multicolumn{1}{c|}{0.5160 ± 0.36}                                 & -                                        \\ \cline{2-8} 
\multicolumn{1}{|c|}{}                                                                                                                & \multicolumn{1}{c|}{\textbf{ITC+ITM+CE}}                      & \multicolumn{1}{c|}{0.1462 ± 0.19}                                 & \multicolumn{1}{c|}{0.0818 ± 0.14}                                 & \multicolumn{1}{c|}{0.1292 ± 0.18}                                 & \multicolumn{1}{c|}{0.1359 ± 0.19}                                 & \multicolumn{1}{c|}{0.1156 ± 0.17}                                 & -                                        \\ \cline{2-8} 
\multicolumn{1}{|c|}{}                                                                                                                & \multicolumn{1}{c|}{\textbf{BLIP2+CE}}                         & \multicolumn{1}{c|}{0.7131 ± 0.32}                                 & \multicolumn{1}{c|}{0.5569 ± 0.39}                                 & \multicolumn{1}{c|}{0.7280 ± 0.36}                                 & \multicolumn{1}{c|}{0.5627 ± 0.41}                                 & \multicolumn{1}{c|}{0.6159 ± 0.39}                                 & -                                        \\ \cline{2-8} 
\multicolumn{1}{|c|}{\multirow{-4}{*}{\textbf{\begin{tabular}[c]{@{}c@{}}Fine-tuned\\ DeCLUTR-ViT\end{tabular}}}}                     & \multicolumn{1}{c|}{\textbf{BLIP2+CE+SupCon}}                  & \multicolumn{1}{c|}{0.7632 ± 0.32}                                 & \multicolumn{1}{c|}{0.5666 ± 0.40}                                 & \multicolumn{1}{c|}{0.7652 ± 0.31}                                 & \multicolumn{1}{c|}{0.5869 ± 0.40}                                 & \multicolumn{1}{c|}{0.6362 ± 0.37}                                 & -                                        \\ \hline
\multicolumn{8}{|c|}{\textbf{Text Macro-F1@X}}                                                                                                                                                                                                                                                                                                                                                                                                                                                                                                                                                                                     \\ \hline
\multicolumn{1}{|c|}{\textbf{DeCLUTR}}                                                                                                & \multicolumn{1}{c|}{\textbf{CE+SupCon}}                        & \multicolumn{1}{c|}{0.9102 ± 0.21}                                 & \multicolumn{1}{c|}{0.6162 ± 0.37}                                 & \multicolumn{1}{c|}{0.7169 ± 0.33}                                 & \multicolumn{1}{c|}{0.6879 ± 0.40}                                 & \multicolumn{1}{c|}{0.6737 ± 0.37}                                 &                                          \\ \hline
\multicolumn{1}{|c|}{{\color[HTML]{340096} }}                                                                                         & \multicolumn{1}{c|}{\textbf{CE}}                               & \multicolumn{1}{c|}{0.8726 ± 0.20}                                 & \multicolumn{1}{c|}{0.5653 ± 0.33}                                 & \multicolumn{1}{c|}{0.7374 ± 0.26}                                 & \multicolumn{1}{c|}{0.7261 ± 0.31}                                 & \multicolumn{1}{c|}{0.6763 ± 0.30}                                 & -                                        \\ \cline{2-8} 
\multicolumn{1}{|c|}{\multirow{-2}{*}{{\color[HTML]{340096} \textbf{\begin{tabular}[c]{@{}c@{}}End2End\\ DeCLUTR-ViT\end{tabular}}}}} & \multicolumn{1}{c|}{{\color[HTML]{340096} \textbf{CE+SupCon}}} & \multicolumn{1}{c|}{{\color[HTML]{340096} \textbf{0.9433 ± 0.16}}} & \multicolumn{1}{c|}{{\color[HTML]{340096} \textbf{0.5819 ± 0.34}}} & \multicolumn{1}{c|}{{\color[HTML]{340096} \textbf{0.7466 ± 0.26}}} & \multicolumn{1}{c|}{{\color[HTML]{340096} \textbf{0.7242 ± 0.31}}} & \multicolumn{1}{c|}{{\color[HTML]{340096} \textbf{0.6841 ± 0.30}}} & -                                        \\ \hline
\multicolumn{1}{|c|}{}                                                                                                                & \multicolumn{1}{c|}{\textbf{ITC}}                             & \multicolumn{1}{c|}{0.3039 ± 0.31}                                 & \multicolumn{1}{c|}{0.3756 ± 0.35}                                 & \multicolumn{1}{c|}{0.4887 ± 0.33}                                 & \multicolumn{1}{c|}{0.4173 ± 0.39}                                 & \multicolumn{1}{c|}{-}                                             & 0.4272 ± 0.36                            \\ \cline{2-8} 
\multicolumn{1}{|c|}{}                                                                                                                & \multicolumn{1}{c|}{\textbf{ITC+ITM}}                         & \multicolumn{1}{c|}{0.5079 ± 0.32}                                 & \multicolumn{1}{c|}{0.5659 ± 0.36}                                 & \multicolumn{1}{c|}{0.7281 ± 0.29}                                 & \multicolumn{1}{c|}{0.5136 ± 0.44}                                 & \multicolumn{1}{c|}{-}                                             & 0.5946 ± 0.35                            \\ \cline{2-8} 
\multicolumn{1}{|c|}{}                                                                                                                & \multicolumn{1}{c|}{\textbf{BLIP2}}                            & \multicolumn{1}{c|}{0.4283 ± 0.33}                                 & \multicolumn{1}{c|}{0.5279 ± 0.38}                                 & \multicolumn{1}{c|}{\textit{0.6552 ± 0.34}}                        & \multicolumn{1}{c|}{0.4216 ± 0.39}                                 & \multicolumn{1}{c|}{-}                                             & 0.5605 ± 0.38                            \\ \cline{2-8} 
\multicolumn{1}{|c|}{\multirow{-4}{*}{\textbf{\begin{tabular}[c]{@{}c@{}}Aligned\\ DeCLUTR-ViT\end{tabular}}}}                    & \multicolumn{1}{c|}{\textbf{BLIP2-Cond}}                       & \multicolumn{1}{c|}{0.4356 ± 0.34}                                 & \multicolumn{1}{c|}{0.5251 ± 0.38}                                 & \multicolumn{1}{c|}{0.6720 ± 0.34}                                 & \multicolumn{1}{c|}{0.4249 ± 0.42}                                 & \multicolumn{1}{c|}{-}                                             & 0.5125 ± 0.38                            \\ \hline
\multicolumn{1}{|c|}{}                                                                                                                & \multicolumn{1}{c|}{\textbf{ITC+CE}}                          & \multicolumn{1}{c|}{0.6805 ± 0.32}                                 & \multicolumn{1}{c|}{0.5054 ± 0.37}                                 & \multicolumn{1}{c|}{0.6877 ± 0.32}                                 & \multicolumn{1}{c|}{0.3790 ± 0.34}                                 & \multicolumn{1}{c|}{0.5240 ± 0.34}                                 & -                                        \\ \cline{2-8} 
\multicolumn{1}{|c|}{}                                                                                                                & \multicolumn{1}{c|}{\textbf{ITC+ITM+CE}}                      & \multicolumn{1}{c|}{0.1438 ± 0.20}                                 & \multicolumn{1}{c|}{0.0748 ± 0.12}                                 & \multicolumn{1}{c|}{0.1218 ± 0.18}                                 & \multicolumn{1}{c|}{0.1214 ± 0.17}                                 & \multicolumn{1}{c|}{0.1060 ± 0.16}                                 & -                                        \\ \cline{2-8} 
\multicolumn{1}{|c|}{}                                                                                                                & \multicolumn{1}{c|}{\textbf{BLIP2+CE}}                         & \multicolumn{1}{c|}{0.7215 ± 0.31}                                 & \multicolumn{1}{c|}{0.5774 ± 0.38}                                 & \multicolumn{1}{c|}{0.7391 ± 0.32}                                 & \multicolumn{1}{c|}{0.5499 ± 0.39}                                 & \multicolumn{1}{c|}{0.6221 ± 0.36}                                 & -                                        \\ \cline{2-8} 
\multicolumn{1}{|c|}{\multirow{-4}{*}{\textbf{\begin{tabular}[c]{@{}c@{}}Fine-tuned\\ DeCLUTR-ViT\end{tabular}}}}                     & \multicolumn{1}{c|}{\textbf{BLIP2+CE+SupCon}}                  & \multicolumn{1}{c|}{0.7879 ± 0.29}                                 & \multicolumn{1}{c|}{0.5762 ± 0.39}                                 & \multicolumn{1}{c|}{0.7482 ± 0.29}                                 & \multicolumn{1}{c|}{0.5912 ± 0.38}                                 & \multicolumn{1}{c|}{0.6385 ± 0.35}                                 & -                                        \\ \hline
\end{tabular}
}
\caption{Comparison of text-to-text retrieval performance for the multimodal, DeCLUTR-ViT backbone, evaluated on the text-only modality using MRR@10, R-Precision@X, and Macro-F1@X metrics. The DeCLUTR-small model serves as the text-only baseline. End2End baselines denote DeCLUTR-ViT models trained directly for vendor identification tasks, while Aligned baselines represent DeCLUTR-ViT backbone pre-trained for text-image alignment tasks using ITC, ITC+ITM, and BLIP2 objectives. Fine-tuned baselines build upon pre-trained aligned models by fine-tuning them for vendor identification tasks on the South region ads.}
\label{tab:multimodal_text_retrieval}
\end{table*}

\begin{table*}[h]
\centering
\resizebox{\linewidth}{!}{%
\begin{tabular}{|c c cccccc|}
\hline
\multicolumn{1}{|c|}{\textbf{Backbone}}                                                                                               & \multicolumn{1}{c|}{\textbf{Loss}}                             & \multicolumn{1}{c|}{\textbf{South}}        & \multicolumn{1}{c|}{\textbf{Midwest}}      & \multicolumn{1}{c|}{\textbf{West}}         & \multicolumn{1}{c|}{\textbf{Northeast}}    & \multicolumn{1}{c|}{\textbf{OOD Avg.}}     & \textbf{ZS Avg.} \\ \hline
\multicolumn{8}{|c|}{\textbf{Vision MRR@10}}                                                                                                                                                                                                                                                                                                                                                                                                                                                                                                                                                                                       \\ \hline
\multicolumn{1}{|c|}{\textbf{ViT}}                                                                                                    & \multicolumn{1}{c|}{\textbf{CE+SupCon}}                        & \multicolumn{1}{c|}{0.3464 ± 0.37}                                 & \multicolumn{1}{c|}{0.3749 ± 0.40}                                 & \multicolumn{1}{c|}{0.4049 ± 0.42}                                 & \multicolumn{1}{c|}{0.4330 ± 0.42}                                 & \multicolumn{1}{c|}{0.4041 ± 0.41}                                 & -                                        \\ \hline
\multicolumn{1}{|c|}{{\color[HTML]{340096} }}                                                                                         & \multicolumn{1}{c|}{\textbf{CE}}                               & \multicolumn{1}{c|}{0.2257 ± 0.33}                                 & \multicolumn{1}{c|}{0.1716 ± 0.32}                                 & \multicolumn{1}{c|}{0.2142 ± 0.35}                                 & \multicolumn{1}{c|}{0.1866 ± 0.32}                                 & \multicolumn{1}{c|}{0.2575 ± 0.33}                                 & -                                        \\ \cline{2-8} 
\multicolumn{1}{|c|}{\multirow{-2}{*}{{\color[HTML]{340096} \textbf{\begin{tabular}[c]{@{}c@{}}End2End\\ DeCLUTR-ViT\end{tabular}}}}} & \multicolumn{1}{c|}{{\color[HTML]{340096} \textbf{CE+SupCon}}} & \multicolumn{1}{c|}{{\color[HTML]{340096} \textbf{0.4045 ± 0.38}}} & \multicolumn{1}{c|}{{\color[HTML]{340096} \textbf{0.3905 ± 0.40}}} & \multicolumn{1}{c|}{{\color[HTML]{340096} \textbf{0.4603 ± 0.45}}} & \multicolumn{1}{c|}{{\color[HTML]{340096} \textbf{0.4521 ± 0.42}}} & \multicolumn{1}{c|}{{\color[HTML]{340096} \textbf{0.4343 ± 0.42}}} & -                                        \\ \hline
\multicolumn{1}{|c|}{}                                                                                                                & \multicolumn{1}{c|}{\textbf{ITC}}                             & \multicolumn{1}{c|}{0.2329 ± 0.30}                                 & \multicolumn{1}{c|}{0.2336 ± 0.33}                                 & \multicolumn{1}{c|}{0.2984 ± 0.39}                                 & \multicolumn{1}{c|}{0.2964 ± 0.37}                                 & \multicolumn{1}{c|}{-}                                             & 0.2761 ± 0.36                            \\ \cline{2-8} 
\multicolumn{1}{|c|}{}                                                                                                                & \multicolumn{1}{c|}{\textbf{ITC+ITM}}                         & \multicolumn{1}{c|}{0.3281 ± 0.37}                                 & \multicolumn{1}{c|}{0.3434 ± 0.39}                                 & \multicolumn{1}{c|}{0.3683 ± 0.43}                                 & \multicolumn{1}{c|}{0.3442 ± 0.40}                                 & \multicolumn{1}{c|}{-}                                             & 0.3324 ± 0.38                            \\ \cline{2-8} 
\multicolumn{1}{|c|}{}                                                                                                                & \multicolumn{1}{c|}{\textbf{BLIP2}}                            & \multicolumn{1}{c|}{0.2119 ± 0.32}                                 & \multicolumn{1}{c|}{0.2055 ± 0.33}                                 & \multicolumn{1}{c|}{0.2674 ± 0.40}                                 & \multicolumn{1}{c|}{0.2858 ± 0.39}                                 & \multicolumn{1}{c|}{-}                                             & 0.2425 ± 0.36                            \\ \cline{2-8} 
\multicolumn{1}{|c|}{\multirow{-4}{*}{\textbf{\begin{tabular}[c]{@{}c@{}}Aligned\\ DeCLUTR-ViT\end{tabular}}}}                    & \multicolumn{1}{c|}{\textbf{BLIP2-Cond}}                       & \multicolumn{1}{c|}{0.2049 ± 0.32}                                 & \multicolumn{1}{c|}{0.1855 ± 0.31}                                 & \multicolumn{1}{c|}{0.2488 ± 0.39}                                 & \multicolumn{1}{c|}{0.2450 ± 0.36}                                 & \multicolumn{1}{c|}{-}                                             & 0.2211 ± 0.35                            \\ \hline
\multicolumn{1}{|c|}{}                                                                                                                & \multicolumn{1}{c|}{\textbf{ITC}}                             & \multicolumn{1}{c|}{0.4157 ± 0.38}                                 & \multicolumn{1}{c|}{0.3512 ± 0.39}                                 & \multicolumn{1}{c|}{0.3818 ± 0.43}                                 & \multicolumn{1}{c|}{0.3792 ± 0.41}                                 & \multicolumn{1}{c|}{0.3707 ± 0.41}                                 & -                                        \\ \cline{2-8} 
\multicolumn{1}{|c|}{}                                                                                                                & \multicolumn{1}{c|}{\textbf{ITC+ITM}}                         & \multicolumn{1}{c|}{0.4239 ± 0.39}                                 & \multicolumn{1}{c|}{0.2851 ± 0.37}                                 & \multicolumn{1}{c|}{0.3417 ± 0.42}                                 & \multicolumn{1}{c|}{0.3600 ± 0.41}                                 & \multicolumn{1}{c|}{0.3289 ± 0.40}                                 & -                                        \\ \cline{2-8} 
\multicolumn{1}{|c|}{}                                                                                                                & \multicolumn{1}{c|}{\textbf{BLIP2}}                            & \multicolumn{1}{c|}{0.3677 ± 0.38}                                 & \multicolumn{1}{c|}{0.2629 ± 0.36}                                 & \multicolumn{1}{c|}{0.3229 ± 0.41}                                 & \multicolumn{1}{c|}{0.3128 ± 0.39}                                 & \multicolumn{1}{c|}{0.2995 ± 0.39}                                 & -                                        \\ \cline{2-8} 
\multicolumn{1}{|c|}{\multirow{-4}{*}{\textbf{\begin{tabular}[c]{@{}c@{}}Fine-tuned\\ DeCLUTR-ViT\end{tabular}}}}                     & \multicolumn{1}{c|}{\textbf{BLIP2-CE+SupCon}}                  & \multicolumn{1}{c|}{0.3470 ± 0.38}                                 & \multicolumn{1}{c|}{0.2542 ± 0.35}                                 & \multicolumn{1}{c|}{0.3026 ± 0.41}                                 & \multicolumn{1}{c|}{0.3312 ± 0.39}                                 & \multicolumn{1}{c|}{0.2960 ± 0.39}                                 & -                                        \\ \hline
\multicolumn{8}{|c|}{\textbf{Vision R-Precision@X}}                                                                                                                                                                                                                                                                                                                                                                                                                                                                                                                                                                                \\ \hline
\multicolumn{1}{|c|}{\textbf{ViT}}                                                                                                    & \multicolumn{1}{c|}{\textbf{CE+SupCon}}                        & \multicolumn{1}{c|}{0.1064 ± 0.16}                                 & \multicolumn{1}{c|}{0.1095 ± 0.16}                                 & \multicolumn{1}{c|}{0.1519 ± 0.20}                                 & \multicolumn{1}{c|}{0.1685 ± 0.21}                                 & \multicolumn{1}{c|}{0.1433 ± 0.19}                                 & -                                        \\ \hline
\multicolumn{1}{|c|}{{\color[HTML]{340096} }}                                                                                         & \multicolumn{1}{c|}{\textbf{CE}}                               & \multicolumn{1}{c|}{0.0862 ± 0.16}                                 & \multicolumn{1}{c|}{0.0567 ± 0.12}                                 & \multicolumn{1}{c|}{0.0915 ± 0.14}                                 & \multicolumn{1}{c|}{0.0676 ± 0.11}                                 & \multicolumn{1}{c|}{0.0719 ± 0.12}                                 & -                                        \\ \cline{2-8} 
\multicolumn{1}{|c|}{\multirow{-2}{*}{{\color[HTML]{340096} \textbf{\begin{tabular}[c]{@{}c@{}}End2End\\ DeCLUTR-ViT\end{tabular}}}}} & \multicolumn{1}{c|}{{\color[HTML]{340096} \textbf{CE+SupCon}}} & \multicolumn{1}{c|}{{\color[HTML]{340096} \textbf{0.1115 ± 0.15}}} & \multicolumn{1}{c|}{{\color[HTML]{340096} \textbf{0.1141 ± 0.16}}} & \multicolumn{1}{c|}{{\color[HTML]{340096} \textbf{0.1768 ± 0.21}}} & \multicolumn{1}{c|}{{\color[HTML]{340096} \textbf{0.1646 ± 0.19}}} & \multicolumn{1}{c|}{{\color[HTML]{340096} \textbf{0.1518 ± 0.19}}} & -                                        \\ \hline
\multicolumn{1}{|c|}{}                                                                                                                & \multicolumn{1}{c|}{\textbf{ITC}}                             & \multicolumn{1}{c|}{0.0537 ± 0.09}                                 & \multicolumn{1}{c|}{0.0752 ± 0.13}                                 & \multicolumn{1}{c|}{0.1275 ± 0.17}                                 & \multicolumn{1}{c|}{0.1143 ± 0.16}                                 & \multicolumn{1}{c|}{-}                                             & 0.1057 ± 0.16                            \\ \cline{2-8} 
\multicolumn{1}{|c|}{}                                                                                                                & \multicolumn{1}{c|}{\textbf{ITC+ITM}}                         & \multicolumn{1}{c|}{0.0650 ± 0.10}                                 & \multicolumn{1}{c|}{0.0826 ± 0.14}                                 & \multicolumn{1}{c|}{0.1218 ± 0.17}                                 & \multicolumn{1}{c|}{0.1003 ± 0.14}                                 & \multicolumn{1}{c|}{-}                                             & 0.0924 ± 0.14                            \\ \cline{2-8} 
\multicolumn{1}{|c|}{}                                                                                                                & \multicolumn{1}{c|}{\textbf{BLIP2}}                            & \multicolumn{1}{c|}{0.0645 ± 0.15}                                 & \multicolumn{1}{c|}{0.0641 ± 0.13}                                 & \multicolumn{1}{c|}{0.1197 ± 0.20}                                 & \multicolumn{1}{c|}{0.1492 ± 0.24}                                 & \multicolumn{1}{c|}{-}                                             & 0.0994 ± 0.18                            \\ \cline{2-8} 
\multicolumn{1}{|c|}{\multirow{-4}{*}{\textbf{\begin{tabular}[c]{@{}c@{}}Aligned\\ DeCLUTR-ViT\end{tabular}}}}                    & \multicolumn{1}{c|}{\textbf{BLIP2-Cond}}                       & \multicolumn{1}{c|}{0.0563 ± 0.13}                                 & \multicolumn{1}{c|}{0.0569 ± 0.13}                                 & \multicolumn{1}{c|}{0.1001 ± 0.18}                                 & \multicolumn{1}{c|}{0.1115 ± 0.20}                                 & \multicolumn{1}{c|}{-}                                             & 0.0812 ± 0.16                            \\ \hline
\multicolumn{1}{|c|}{}                                                                                                                & \multicolumn{1}{c|}{\textbf{ITC}}                             & \multicolumn{1}{c|}{0.1247 ± 0.17}                                 & \multicolumn{1}{c|}{0.0957 ± 0.15}                                 & \multicolumn{1}{c|}{0.1461 ± 0.18}                                 & \multicolumn{1}{c|}{0.1383 ± 0.17}                                 & \multicolumn{1}{c|}{0.1267 ± 0.17}                                 & -                                        \\ \cline{2-8} 
\multicolumn{1}{|c|}{}                                                                                                                & \multicolumn{1}{c|}{\textbf{ITC+ITM}}                         & \multicolumn{1}{c|}{0.1462 ± 0.19}                                 & \multicolumn{1}{c|}{0.0818 ± 0.14}                                 & \multicolumn{1}{c|}{0.1292 ± 0.18}                                 & \multicolumn{1}{c|}{0.1359 ± 0.19}                                 & \multicolumn{1}{c|}{0.1156 ± 0.17}                                 & -                                        \\ \cline{2-8} 
\multicolumn{1}{|c|}{}                                                                                                                & \multicolumn{1}{c|}{\textbf{BLIP2}}                            & \multicolumn{1}{c|}{0.1370 ± 0.19}                                 & \multicolumn{1}{c|}{0.0775 ± 0.14}                                 & \multicolumn{1}{c|}{0.1217 ± 0.18}                                 & \multicolumn{1}{c|}{0.1393 ± 0.21}                                 & \multicolumn{1}{c|}{0.1128 ± 0.18}                                 & -                                        \\ \cline{2-8} 
\multicolumn{1}{|c|}{\multirow{-4}{*}{\textbf{\begin{tabular}[c]{@{}c@{}}Fine-tuned\\ DeCLUTR-ViT\end{tabular}}}}                     & \multicolumn{1}{c|}{\textbf{BLIP2-CE+SupCon}}                  & \multicolumn{1}{c|}{0.1256 ± 0.19}                                 & \multicolumn{1}{c|}{0.0777 ± 0.14}                                 & \multicolumn{1}{c|}{0.1228 ± 0.17}                                 & \multicolumn{1}{c|}{0.1414 ± 0.20}                                 & \multicolumn{1}{c|}{0.1140 ± 0.17}                                 & -                                        \\ \hline
\multicolumn{8}{|c|}{\textbf{Vision Macro-F1@X}}                                                                                                                                                                                                                                                                                                                                                                                                                                                                                                                                                                                   \\ \hline
\multicolumn{1}{|c|}{\textbf{ViT}}                                                                                                    & \multicolumn{1}{c|}{\textbf{CE+SupCon}}                        & \multicolumn{1}{c|}{0.1296 ± 0.21}                                 & \multicolumn{1}{c|}{0.0948 ± 0.14}                                 & \multicolumn{1}{c|}{0.1460 ± 0.20}                                 & \multicolumn{1}{c|}{0.1497 ± 0.20}                                 & \multicolumn{1}{c|}{0.1302 ± 0.18}                                 & -                                        \\ \hline
\multicolumn{1}{|c|}{{\color[HTML]{340096} }}                                                                                         & \multicolumn{1}{c|}{\textbf{CE}}                               & \multicolumn{1}{c|}{0.1028 ± 0.21}                                 & \multicolumn{1}{c|}{0.0600 ± 0.11}                                 & \multicolumn{1}{c|}{0.0960 ± 0.15}                                 & \multicolumn{1}{c|}{0.0657 ± 0.11}                                 & \multicolumn{1}{c|}{0.0859 ± 0.14}                                 & -                                        \\ \cline{2-8} 
\multicolumn{1}{|c|}{\multirow{-2}{*}{{\color[HTML]{340096} \textbf{\begin{tabular}[c]{@{}c@{}}End2End\\ DeCLUTR-ViT\end{tabular}}}}} & \multicolumn{1}{c|}{{\color[HTML]{340096} \textbf{CE+SupCon}}} & \multicolumn{1}{c|}{{\color[HTML]{340096} \textbf{0.1152 ± 0.17}}} & \multicolumn{1}{c|}{{\color[HTML]{340096} \textbf{0.1049 ± 0.14}}} & \multicolumn{1}{c|}{{\color[HTML]{340096} \textbf{0.1739 ± 0.21}}} & \multicolumn{1}{c|}{{\color[HTML]{340096} \textbf{0.1493 ± 0.18}}} & \multicolumn{1}{c|}{{\color[HTML]{340096} \textbf{0.1427 ± 0.19}}} & -                                        \\ \hline
\multicolumn{1}{|c|}{}                                                                                                                & \multicolumn{1}{c|}{\textbf{ITC}}                             & \multicolumn{1}{c|}{0.0689 ± 0.11}                                 & \multicolumn{1}{c|}{0.0892 ± 0.14}                                 & \multicolumn{1}{c|}{0.1415 ± 0.19}                                 & \multicolumn{1}{c|}{0.1072 ± 0.15}                                 & \multicolumn{1}{c|}{-}                                             & 0.1118 ± 0.18                            \\ \cline{2-8} 
\multicolumn{1}{|c|}{}                                                                                                                & \multicolumn{1}{c|}{\textbf{ITC+ITM}}                         & \multicolumn{1}{c|}{0.0614 ± 0.10}                                 & \multicolumn{1}{c|}{0.0675 ± 0.11}                                 & \multicolumn{1}{c|}{0.1070 ± 0.15}                                 & \multicolumn{1}{c|}{0.0933 ± 0.13}                                 & \multicolumn{1}{c|}{-}                                             & 0.0837 ± 0.13                            \\ \cline{2-8} 
\multicolumn{1}{|c|}{}                                                                                                                & \multicolumn{1}{c|}{\textbf{BLIP2}}                            & \multicolumn{1}{c|}{0.0938 ± 0.20}                                 & \multicolumn{1}{c|}{0.0908 ± 0.17}                                 & \multicolumn{1}{c|}{0.1281 ± 0.22}                                 & \multicolumn{1}{c|}{0.1458 ± 0.24}                                 & \multicolumn{1}{c|}{-}                                             & 0.1146 ± 0.21                            \\ \cline{2-8} 
\multicolumn{1}{|c|}{\multirow{-4}{*}{\textbf{\begin{tabular}[c]{@{}c@{}}Aligned\\ DeCLUTR-ViT\end{tabular}}}}                    & \multicolumn{1}{c|}{\textbf{BLIP2-Cond}}                       & \multicolumn{1}{c|}{0.0805 ± 0.18}                                 & \multicolumn{1}{c|}{0.0776 ± 0.16}                                 & \multicolumn{1}{c|}{0.1074 ± 0.20}                                 & \multicolumn{1}{c|}{0.1088 ± 0.19}                                 & \multicolumn{1}{c|}{-}                                             & 0.0936 ± 0.18                            \\ \hline
\multicolumn{1}{|c|}{}                                                                                                                & \multicolumn{1}{c|}{\textbf{ITC}}                             & \multicolumn{1}{c|}{0.1319 ± 0.18}                                 & \multicolumn{1}{c|}{0.0914 ± 0.14}                                 & \multicolumn{1}{c|}{0.1485 ± 0.20}                                 & \multicolumn{1}{c|}{0.1333 ± 0.17}                                 & \multicolumn{1}{c|}{0.1244 ± 0.17}                                 & -                                        \\ \cline{2-8} 
\multicolumn{1}{|c|}{}                                                                                                                & \multicolumn{1}{c|}{\textbf{ITC+ITM}}                         & \multicolumn{1}{c|}{0.1438 ± 0.20}                                 & \multicolumn{1}{c|}{0.0748 ± 0.12}                                 & \multicolumn{1}{c|}{0.1218 ± 0.18}                                 & \multicolumn{1}{c|}{0.1214 ± 0.17}                                 & \multicolumn{1}{c|}{0.1060 ± 0.16}                                 & -                                        \\ \cline{2-8} 
\multicolumn{1}{|c|}{}                                                                                                                & \multicolumn{1}{c|}{\textbf{BLIP2}}                            & \multicolumn{1}{c|}{0.1517 ± 0.23}                                 & \multicolumn{1}{c|}{0.0837 ± 0.14}                                 & \multicolumn{1}{c|}{0.1277 ± 0.19}                                 & \multicolumn{1}{c|}{0.1367 ± 0.20}                                 & \multicolumn{1}{c|}{0.1160 ± 0.18}                                 & -                                        \\ \cline{2-8} 
\multicolumn{1}{|c|}{\multirow{-4}{*}{\textbf{\begin{tabular}[c]{@{}c@{}}Fine-tuned\\ DeCLUTR-ViT\end{tabular}}}}                     & \multicolumn{1}{c|}{\textbf{BLIP2-CE+SupCon}}                  & \multicolumn{1}{c|}{0.1526 ± 0.24}                                 & \multicolumn{1}{c|}{0.0799 ± 0.14}                                 & \multicolumn{1}{c|}{0.1276 ± 0.19}                                 & \multicolumn{1}{c|}{0.1335 ± 0.20}                                 & \multicolumn{1}{c|}{0.1137 ± 0.18}                                 & -                                        \\ \hline
\end{tabular}
}
\caption{Comparison of image-to-image retrieval performance for the multimodal, DeCLUTR-ViT backbone, evaluated on the vision-only modality using MRR@10, R-Precision@X, and Macro-F1@X metrics. The ViT-base-patch16-244 model serves as the vision-only baseline. End2End baselines denote DeCLUTR-ViT models trained directly for vendor identification tasks, while Aligned baselines represent DeCLUTR-ViT backbone pre-trained for text-image alignment tasks using ITC, ITC+ITM, and BLIP2 objectives. Fine-tuned baselines build upon pre-trained aligned models by fine-tuning them for vendor identification tasks on the South region ads.}
\label{tab:multimodal_vision_retrieval}
\end{table*}

\begin{table*}[h]
\centering
\resizebox{\linewidth}{!}{%
\begin{tabular}{|c c cccccc|}
\hline
\multicolumn{1}{|c|}{\textbf{Backbone}}                                                                                               & \multicolumn{1}{c|}{\textbf{Loss}}                             & \multicolumn{1}{c|}{\textbf{South}}                                & \multicolumn{1}{c|}{\textbf{Midwest}}                              & \multicolumn{1}{c|}{\textbf{West}}                                 & \multicolumn{1}{c|}{\textbf{Northeast}}                            & \multicolumn{1}{c|}{\textbf{OOD Avg.}}                             & \textbf{ZS Avg.} \\ \hline
\multicolumn{8}{|c|}{\textbf{Multimodal MRR@10}}                                                                                                                                                                                                                                                                                                                                                                                                                                                                                                                                                           \\ \hline
\multicolumn{1}{|c|}{{\color[HTML]{340096} }}                                                                                         & \multicolumn{1}{c|}{\textbf{CE}}                               & \multicolumn{1}{c|}{0.9669 ± 0.13}                                 & \multicolumn{1}{c|}{0.9297 ± 0.20}                                 & \multicolumn{1}{c|}{0.9592 ± 0.17}                                 & \multicolumn{1}{c|}{0.9650 ± 0.14}                                 & \multicolumn{1}{c|}{0.9513 ± 0.17}                                 & -                \\ \cline{2-8} 
\multicolumn{1}{|c|}{\multirow{-2}{*}{{\color[HTML]{340096} \textbf{\begin{tabular}[c]{@{}c@{}}End2End\\ DeCLUTR-ViT\end{tabular}}}}} & \multicolumn{1}{c|}{{\color[HTML]{340096} \textbf{CE+SupCon}}} & \multicolumn{1}{c|}{{\color[HTML]{340096} \textbf{0.9859 ± 0.10}}} & \multicolumn{1}{c|}{{\color[HTML]{340096} \textbf{0.9658 ± 0.15}}} & \multicolumn{1}{c|}{{\color[HTML]{340096} \textbf{0.9834 ± 0.11}}} & \multicolumn{1}{c|}{{\color[HTML]{340096} \textbf{0.9735 ± 0.13}}} & \multicolumn{1}{c|}{{\color[HTML]{340096} \textbf{0.9742 ± 0.13}}} & -                \\ \hline
\multicolumn{1}{|c|}{}                                                                                                                & \multicolumn{1}{c|}{\textbf{ITC}}                             & \multicolumn{1}{c|}{0.6574 ± 0.35}                                 & \multicolumn{1}{c|}{0.6822 ± 0.36}                                 & \multicolumn{1}{c|}{0.7396 ± 0.36}                                 & \multicolumn{1}{c|}{0.6750 ± 0.38}                                 & \multicolumn{1}{c|}{-}                                             & 0.6886 ± 0.36    \\ \cline{2-8} 
\multicolumn{1}{|c|}{}                                                                                                                & \multicolumn{1}{c|}{\textbf{ITC+ITM}}                         & \multicolumn{1}{c|}{0.9375 ± 0.18}                                 & \multicolumn{1}{c|}{0.9389 ± 0.19}                                 & \multicolumn{1}{c|}{0.9601 ± 0.16}                                 & \multicolumn{1}{c|}{0.9715 ± 0.14}                                 & \multicolumn{1}{c|}{-}                                             & 0.9520 ± 0.17    \\ \cline{2-8} 
\multicolumn{1}{|c|}{}                                                                                                                & \multicolumn{1}{c|}{\textbf{BLIP2}}                            & \multicolumn{1}{c|}{0.6142 ± 0.36}                                 & \multicolumn{1}{c|}{0.6136 ± 0.39}                                 & \multicolumn{1}{c|}{0.6108 ± 0.41}                                 & \multicolumn{1}{c|}{0.5921 ± 0.42}                                 & \multicolumn{1}{c|}{-}                                             & 0.6077 ± 0.40    \\ \cline{2-8} 
\multicolumn{1}{|c|}{\multirow{-4}{*}{\textbf{\begin{tabular}[c]{@{}c@{}}Aligned\\ DeCLUTR-ViT\end{tabular}}}}                    & \multicolumn{1}{c|}{\textbf{BLIP2-Cond}}                       & \multicolumn{1}{c|}{0.6052 ± 0.36}                                 & \multicolumn{1}{c|}{0.6006 ± 0.39}                                 & \multicolumn{1}{c|}{0.5975 ± 0.41}                                 & \multicolumn{1}{c|}{0.5657 ± 0.42}                                 & \multicolumn{1}{c|}{-}                                             & 0.5923 ± 0.40    \\ \hline
\multicolumn{1}{|c|}{}                                                                                                                & \multicolumn{1}{c|}{\textbf{ITC}}                             & \multicolumn{1}{c|}{0.9650 ± 0.13}                                 & \multicolumn{1}{c|}{0.8331 ± 0.29}                                 & \multicolumn{1}{c|}{0.7313 ± 0.36}                                 & \multicolumn{1}{c|}{0.7641 ± 0.34}                                 & \multicolumn{1}{c|}{0.7762 ± 0.33}                                 & -                \\ \cline{2-8} 
\multicolumn{1}{|c|}{}                                                                                                                & \multicolumn{1}{c|}{\textbf{ITC+ITM}}                         & \multicolumn{1}{c|}{0.9739 ± 0.12}                                 & \multicolumn{1}{c|}{0.9285 ± 0.20}                                 & \multicolumn{1}{c|}{0.9498 ± 0.19}                                 & \multicolumn{1}{c|}{0.9655 ± 0.15}                                 & \multicolumn{1}{c|}{0.9480 ± 0.23}                                 & -                \\ \cline{2-8} 
\multicolumn{1}{|c|}{}                                                                                                                & \multicolumn{1}{c|}{\textbf{BLIP2}}                            & \multicolumn{1}{c|}{0.9774 ± 0.11}                                 & \multicolumn{1}{c|}{0.9378 ± 0.20}                                 & \multicolumn{1}{c|}{0.9559 ± 0.18}                                 & \multicolumn{1}{c|}{0.9690 ± 0.14}                                 & \multicolumn{1}{c|}{0.9542 ± 0.17}                                 & -                \\ \cline{2-8} 
\multicolumn{1}{|c|}{\multirow{-4}{*}{\textbf{\begin{tabular}[c]{@{}c@{}}Fine-tuned\\ DeCLUTR-ViT\end{tabular}}}}                     & \multicolumn{1}{c|}{\textbf{BLIP2-CE+SupCon}}                  & \multicolumn{1}{c|}{0.9814 ± 0.10}                                 & \multicolumn{1}{c|}{0.9426 ± 0.19}                                 & \multicolumn{1}{c|}{0.9648 ± 0.15}                                 & \multicolumn{1}{c|}{0.9759 ± 0.12}                                 & \multicolumn{1}{c|}{0.9602 ± 0.19}                                 &                  \\ \hline
\multicolumn{8}{|c|}{\textbf{Multimodal R-Precision@X}}                                                                                                                                                                                                                                                                                                                                                                                                                                                                                                                                                    \\ \hline
\multicolumn{1}{|c|}{{\color[HTML]{340096} }}                                                                                         & \multicolumn{1}{c|}{\textbf{CE}}                               & \multicolumn{1}{c|}{0.8040 ± 0.20}                                 & \multicolumn{1}{c|}{0.6217 ± 0.26}                                 & \multicolumn{1}{c|}{0.7429 ± 0.24}                                 & \multicolumn{1}{c|}{0.6980 ± 0.27}                                 & \multicolumn{1}{c|}{0.6875 ± 0.26}                                 & -                \\ \cline{2-8} 
\multicolumn{1}{|c|}{\multirow{-2}{*}{{\color[HTML]{340096} \textbf{\begin{tabular}[c]{@{}c@{}}End2End\\ DeCLUTR-ViT\end{tabular}}}}} & \multicolumn{1}{c|}{{\color[HTML]{340096} \textbf{CE+SupCon}}} & \multicolumn{1}{c|}{{\color[HTML]{340096} \textbf{0.9248 ± 0.14}}} & \multicolumn{1}{c|}{{\color[HTML]{340096} \textbf{0.6567 ± 0.30}}} & \multicolumn{1}{c|}{{\color[HTML]{340096} \textbf{0.7861 ± 0.25}}} & \multicolumn{1}{c|}{{\color[HTML]{340096} \textbf{0.7178 ± 0.30}}} & \multicolumn{1}{c|}{{\color[HTML]{340096} \textbf{0.7202 ± 0.28}}} & -                \\ \hline
\multicolumn{1}{|c|}{}                                                                                                                & \multicolumn{1}{c|}{\textbf{ITC}}                             & \multicolumn{1}{c|}{0.1797 ± 0.16}                                 & \multicolumn{1}{c|}{0.2373 ± 0.20}                                 & \multicolumn{1}{c|}{0.3330 ± 0.23}                                 & \multicolumn{1}{c|}{0.3076 ± 0.24}                                 & \multicolumn{1}{c|}{-}                                             & 0.2644 ± 0.21    \\ \cline{2-8} 
\multicolumn{1}{|c|}{}                                                                                                                & \multicolumn{1}{c|}{\textbf{ITC+ITM}}                         & \multicolumn{1}{c|}{0.4939 ± 0.24}                                 & \multicolumn{1}{c|}{0.5705 ± 0.26}                                 & \multicolumn{1}{c|}{0.7046 ± 0.23}                                 & \multicolumn{1}{c|}{0.6747 ± 0.26}                                 & \multicolumn{1}{c|}{-}                                             & 0.6109 ± 0.25    \\ \cline{2-8} 
\multicolumn{1}{|c|}{}                                                                                                                & \multicolumn{1}{c|}{\textbf{BLIP2}}                            & \multicolumn{1}{c|}{0.1708 ± 0.22}                                 & \multicolumn{1}{c|}{0.1847 ± 0.22}                                 & \multicolumn{1}{c|}{0.2182 ± 0.24}                                 & \multicolumn{1}{c|}{0.2841 ± 0.32}                                 & \multicolumn{1}{c|}{-}                                             & 0.2145 ± 0.25    \\ \cline{2-8} 
\multicolumn{1}{|c|}{\multirow{-4}{*}{\textbf{\begin{tabular}[c]{@{}c@{}}Aligned\\ DeCLUTR-ViT\end{tabular}}}}                    & \multicolumn{1}{c|}{\textbf{BLIP2-Cond}}                       & \multicolumn{1}{c|}{0.1455 ± 0.20}                                 & \multicolumn{1}{c|}{0.1602 ± 0.20}                                 & \multicolumn{1}{c|}{0.1830 ± 0.21}                                 & \multicolumn{1}{c|}{0.2324 ± 0.29}                                 & \multicolumn{1}{c|}{-}                                             & 0.1803 ± 0.23    \\ \hline
\multicolumn{1}{|c|}{}                                                                                                                & \multicolumn{1}{c|}{\textbf{ITC}}                             & \multicolumn{1}{c|}{0.7377 ± 0.21}                                 & \multicolumn{1}{c|}{0.3716 ± 0.22}                                 & \multicolumn{1}{c|}{0.2844 ± 0.22}                                 & \multicolumn{1}{c|}{0.3700 ± 0.26}                                 & \multicolumn{1}{c|}{0.3420 ± 0.23}                                 & -                \\ \cline{2-8} 
\multicolumn{1}{|c|}{}                                                                                                                & \multicolumn{1}{c|}{\textbf{ITC+ITM}}                         & \multicolumn{1}{c|}{0.7282 ± 0.22}                                 & \multicolumn{1}{c|}{0.4968 ± 0.23}                                 & \multicolumn{1}{c|}{0.6109 ± 0.23}                                 & \multicolumn{1}{c|}{0.6419 ± 0.27}                                 & \multicolumn{1}{c|}{0.5832 ± 0.24}                                 & -                \\ \cline{2-8} 
\multicolumn{1}{|c|}{}                                                                                                                & \multicolumn{1}{c|}{\textbf{BLIP2}}                            & \multicolumn{1}{c|}{0.7723 ± 0.2}                                  & \multicolumn{1}{c|}{0.5524 ± 0.25}                                 & \multicolumn{1}{c|}{0.6759 ± 0.23}                                 & \multicolumn{1}{c|}{0.6691 ± 0.27}                                 & \multicolumn{1}{c|}{0.6325 ± 0.25}                                 & -                \\ \cline{2-8} 
\multicolumn{1}{|c|}{\multirow{-4}{*}{\textbf{\begin{tabular}[c]{@{}c@{}}Fine-tuned\\ DeCLUTR-ViT\end{tabular}}}}                     & \multicolumn{1}{c|}{\textbf{BLIP2-CE+SupCon}}                  & \multicolumn{1}{c|}{0.7950 ± 0.19}                                 & \multicolumn{1}{c|}{0.5564 ± 0.25}                                 & \multicolumn{1}{c|}{0.6943 ± 0.23}                                 & \multicolumn{1}{c|}{0.6809 ± 0.26}                                 & \multicolumn{1}{c|}{0.6524 ± 0.25}                                 &                  \\ \hline
\multicolumn{8}{|c|}{\textbf{Multimodal Macro-F1@X}}                                                                                                                                                                                                                                                                                                                                                                                                                                                                                                                                                       \\ \hline
\multicolumn{1}{|c|}{{\color[HTML]{340096} }}                                                                                         & \multicolumn{1}{c|}{\textbf{CE}}                               & \multicolumn{1}{c|}{0.8294 ± 0.21}                                 & \multicolumn{1}{c|}{0.5618 ± 0.29}                                 & \multicolumn{1}{c|}{0.7408 ± 0.24}                                 & \multicolumn{1}{c|}{0.7053 ± 0.29}                                 & \multicolumn{1}{c|}{0.6693 ± 0.27}                                 & -                \\ \cline{2-8} 
\multicolumn{1}{|c|}{\multirow{-2}{*}{{\color[HTML]{340096} \textbf{\begin{tabular}[c]{@{}c@{}}End2End\\ DeCLUTR-ViT\end{tabular}}}}} & \multicolumn{1}{c|}{{\color[HTML]{340096} \textbf{CE+SupCon}}} & \multicolumn{1}{c|}{{\color[HTML]{340096} \textbf{0.9595 ± 0.12}}} & \multicolumn{1}{c|}{{\color[HTML]{340096} \textbf{0.5671 ± 0.33}}} & \multicolumn{1}{c|}{{\color[HTML]{340096} \textbf{0.7560 ± 0.26}}} & \multicolumn{1}{c|}{{\color[HTML]{340096} \textbf{0.7333 ± 0.30}}} & \multicolumn{1}{c|}{{\color[HTML]{340096} \textbf{0.6855 ± 0.29}}} & -                \\ \hline
\multicolumn{1}{|c|}{}                                                                                                                & \multicolumn{1}{c|}{\textbf{ITC}}                             & \multicolumn{1}{c|}{0.2519 ± 0.23}                                 & \multicolumn{1}{c|}{0.3254 ± 0.26}                                 & \multicolumn{1}{c|}{0.4687 ± 0.27}                                 & \multicolumn{1}{c|}{0.3493 ± 0.26}                                 & \multicolumn{1}{c|}{-}                                             & 0.3488 ± 0.26    \\ \cline{2-8} 
\multicolumn{1}{|c|}{}                                                                                                                & \multicolumn{1}{c|}{\textbf{ITC+ITM}}                         & \multicolumn{1}{c|}{0.4809 ± 0.27}                                 & \multicolumn{1}{c|}{0.5239 ± 0.28}                                 & \multicolumn{1}{c|}{0.7023 ± 0.23}                                 & \multicolumn{1}{c|}{0.6934 ± 0.27}                                 & \multicolumn{1}{c|}{-}                                             & 0.6001 ± 0.26    \\ \cline{2-8} 
\multicolumn{1}{|c|}{}                                                                                                                & \multicolumn{1}{c|}{\textbf{BLIP2}}                            & \multicolumn{1}{c|}{0.3263 ± 0.35}                                 & \multicolumn{1}{c|}{0.3408 ± 0.35}                                 & \multicolumn{1}{c|}{0.4612 ± 0.37}                                 & \multicolumn{1}{c|}{0.4190 ± 0.38}                                 & \multicolumn{1}{c|}{-}                                             & 0.3868 ± 0.37    \\ \cline{2-8} 
\multicolumn{1}{|c|}{\multirow{-4}{*}{\textbf{\begin{tabular}[c]{@{}c@{}}Aligned\\ DeCLUTR-ViT\end{tabular}}}}                    & \multicolumn{1}{c|}{\textbf{BLIP2-Cond}}                       & \multicolumn{1}{c|}{0.2724 ± 0.32}                                 & \multicolumn{1}{c|}{0.2850 ± 0.32}                                 & \multicolumn{1}{c|}{0.3649 ± 0.33}                                 & \multicolumn{1}{c|}{0.3353 ± 0.35}                                 & \multicolumn{1}{c|}{-}                                             & 0.3144 ± 0.33    \\ \hline
\multicolumn{1}{|c|}{}                                                                                                                & \multicolumn{1}{c|}{\textbf{ITC}}                             & \multicolumn{1}{c|}{0.7698 ± 0.23}                                 & \multicolumn{1}{c|}{0.4008 ± 0.25}                                 & \multicolumn{1}{c|}{0.4003 ± 0.27}                                 & \multicolumn{1}{c|}{0.3881 ± 0.28}                                 & \multicolumn{1}{c|}{0.3964 ± 0.27}                                 & -                \\ \cline{2-8} 
\multicolumn{1}{|c|}{}                                                                                                                & \multicolumn{1}{c|}{\textbf{ITC+ITM}}                         & \multicolumn{1}{c|}{0.7313 ± 0.25}                                 & \multicolumn{1}{c|}{0.4538 ± 0.26}                                 & \multicolumn{1}{c|}{0.6275 ± 0.24}                                 & \multicolumn{1}{c|}{0.6591 ± 0.28}                                 & \multicolumn{1}{c|}{0.5801 ± 0.27}                                 & -                \\ \cline{2-8} 
\multicolumn{1}{|c|}{}                                                                                                                & \multicolumn{1}{c|}{\textbf{BLIP2}}                            & \multicolumn{1}{c|}{0.7973 ± 0.22}                                 & \multicolumn{1}{c|}{0.5325 ± 0.28}                                 & \multicolumn{1}{c|}{0.7050 ± 0.24}                                 & \multicolumn{1}{c|}{0.6944 ± 0.29}                                 & \multicolumn{1}{c|}{0.6440 ± 0.27}                                 & -                \\ \cline{2-8} 
\multicolumn{1}{|c|}{\multirow{-4}{*}{\textbf{\begin{tabular}[c]{@{}c@{}}Fine-tuned\\ DeCLUTR-ViT\end{tabular}}}}                     & \multicolumn{1}{c|}{\textbf{BLIP2-CE+SupCon}}                  & \multicolumn{1}{c|}{0.8487 ± 0.20}                                 & \multicolumn{1}{c|}{0.5446 ± 0.29}                                 & \multicolumn{1}{c|}{0.7250 ± 0.24}                                 & \multicolumn{1}{c|}{0.7077 ± 0.29}                                 & \multicolumn{1}{c|}{0.6591 ± 0.27}                                 &                  \\ \hline
\end{tabular}
}
\caption{Comparison of multimodal retrieval performance for the DeCLUTR-ViT backbone evaluated on the multimodal (text and image) ads using MRR@10, R-Precision@X, and Macro-F1@X metrics. The End2End baselines represent the DeCLUTR-ViT backbone trained directly on the vendor identification task, while the Pre-trained baselines involve an image-text alignment task aligning text and images from the same advertisements. The Fine-tuned baselines build upon the Pre-trained models by performing vendor identification on the South region multimodal ads.}
\label{tab:multimodal_retrieval}
\end{table*}
\FloatBarrier
\clearpage

\subsection{Further Insights}
\label{app:insights}
This section evaluates the multimodal DeCLUTR-ViT backbone trained with the CE+SupCon objective, generating comprehensive insights into model learning and retrieval performance. All line plots have been smoothed for clarity and readability by setting the window size to 30.

\subsubsection{Insights from the Multimodal Classifier on the South Region Dataset}
\begin{figure*}[h]
    \centering
    \begin{minipage}[t]{\linewidth}
        \centering
        \includegraphics[width=\linewidth,keepaspectratio]{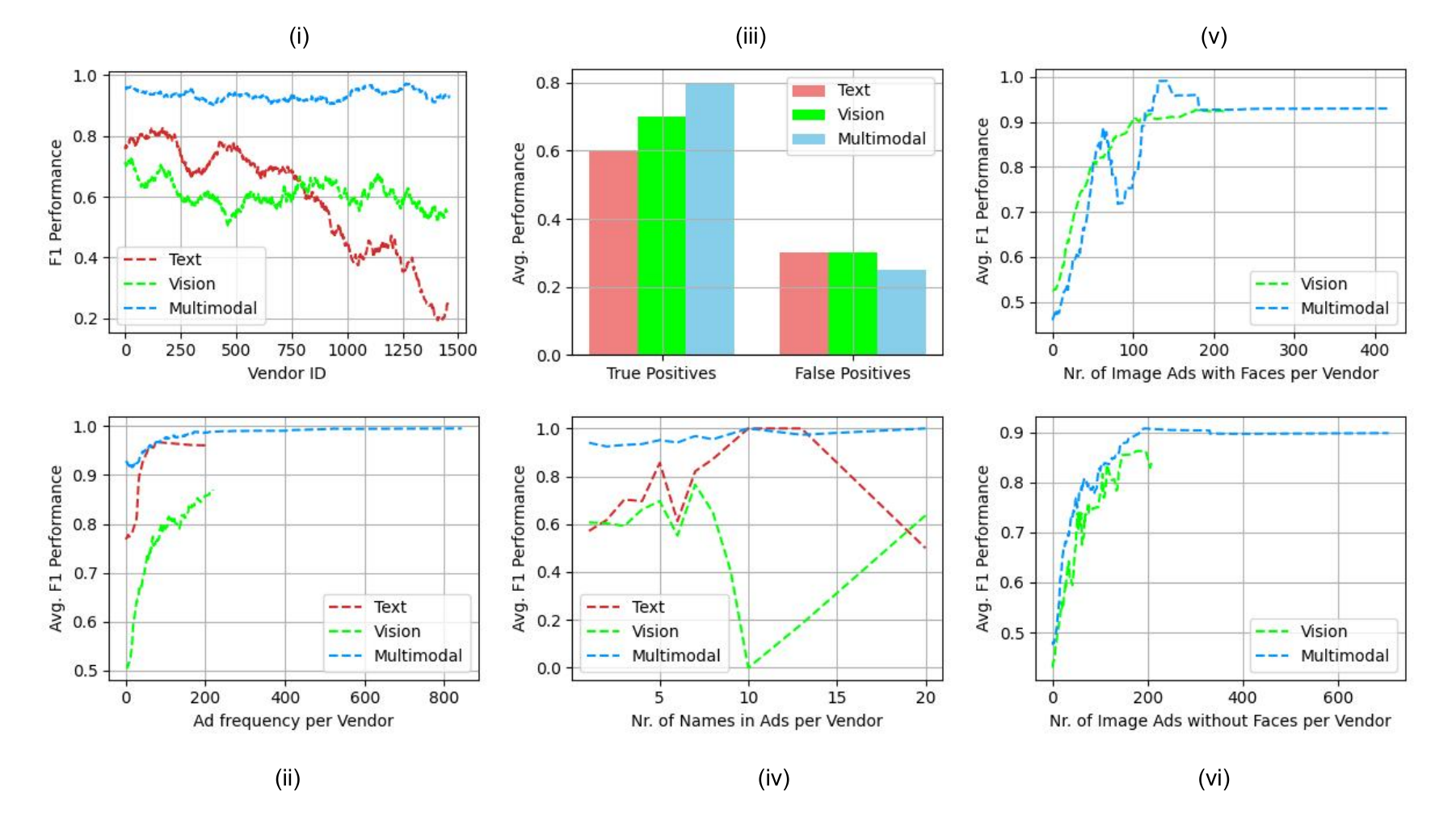}
        \caption{\label{fig:error_analysis_classification} Comparison of model performance among text-only, vision-only, and multimodal classifiers trained on the South region test dataset: (i) F1 score across different vendor IDs, (ii) Average F1 score for vendors with varying ad frequencies, (iii) Analysis of true and false positives, (iv) Average F1 score relative to the number of escort names (potentially representing different individuals) in vendor ads, and (v, vi) Average F1 score based on the number of vendor images with and without faces.}
    \end{minipage}
\end{figure*}

Figure \ref{fig:error_analysis_classification}(i) compares the average F1 performance of the DeCLUTR-small text-only, ViT-base-patch16-244 vision-only, and multimodal DeCLUTR-ViT classifiers for vendors in the South region dataset. The results show that the multimodal classifier consistently outperforms text- and vision-only baselines across all vendors. Further analysis, supported by the vendor frequency distribution in Table \ref{fig:freq_distribution} and \ref{fig:error_analysis_classification}(ii), indicates that many vendors in the text-only and vision-only datasets have very few ads, likely contributing to the lower model performance. In contrast, the multimodal classifier benefits from more training examples per vendor (at least five examples when combining text and vision data). This expanded training set allows the model to capture a broader range of stylistic and visual patterns, resulting in better performance. The findings underscore the importance of multimodal integration in enhancing model effectiveness to capture richer and more complementary stylometric cues, particularly for vendors with sparse data in individual modalities.
 
\hspace{\parindent} Figure \ref{fig:error_analysis_classification}(iii) compares the average number of true positives and false positives achieved by the text-only, vision-only, and multimodal DeCLUTR-ViT baselines across all vendors in the South region dataset. The results reveal a clear advantage for the multimodal baseline, which yields significantly more true positives while maintaining fewer false positives than the other baselines. The results emphasize the superiority of multimodal approaches in minimizing errors and improving the reliability of predictions.

\hspace{\parindent} Figure \ref{fig:error_analysis_classification}(iv) illustrates the average F1 performance of the text-only, vision-only, and multimodal baselines as a function of the number of names per vendor present in the text ads. Since multiple escort names likely represent different individuals, this analysis assesses the models' ability to link varying text descriptions and facial features to a single vendor. To extract escort names from the text ads, we utilized \citep{li-etal-2022-extracting-person-name}, though manual inspection revealed that it often failed to extract names accurately. However, the extracted entities remained consistent, allowing us to use them as unique identifiers representing escort names. The results indicate that the multimodal baseline consistently outperforms the text-only and vision-only baselines, demonstrating resilience and robust performance even as the number of escort names per vendor increases. 

\hspace{\parindent} Finally, Figure \ref{fig:error_analysis_classification}(v) and (vi) compare the average F1 performance of the vision-only and multimodal baselines as a function of the number of images with and without faces per vendor. In Figure (v), as the number of images with faces increases, the multimodal baseline performs worse than the vision-only baseline up to approximately 120 images. Beyond this threshold, the multimodal baseline either outperforms or performs on par with the vision-only baseline, indicating its ability to adapt as the data volume increases. In contrast, Figure (vi) shows that for images without faces, the multimodal baseline consistently outperforms the vision-only baseline, demonstrating its superior capacity to effectively leverage text and image features, even when facial features are absent.

\subsubsection{Insights from the Multimodal Retriever on the OOD Datasets}
\begin{figure*}[h]
    \centering
    \begin{minipage}[t]{\linewidth}
        \centering
        \includegraphics[width=\linewidth,keepaspectratio]{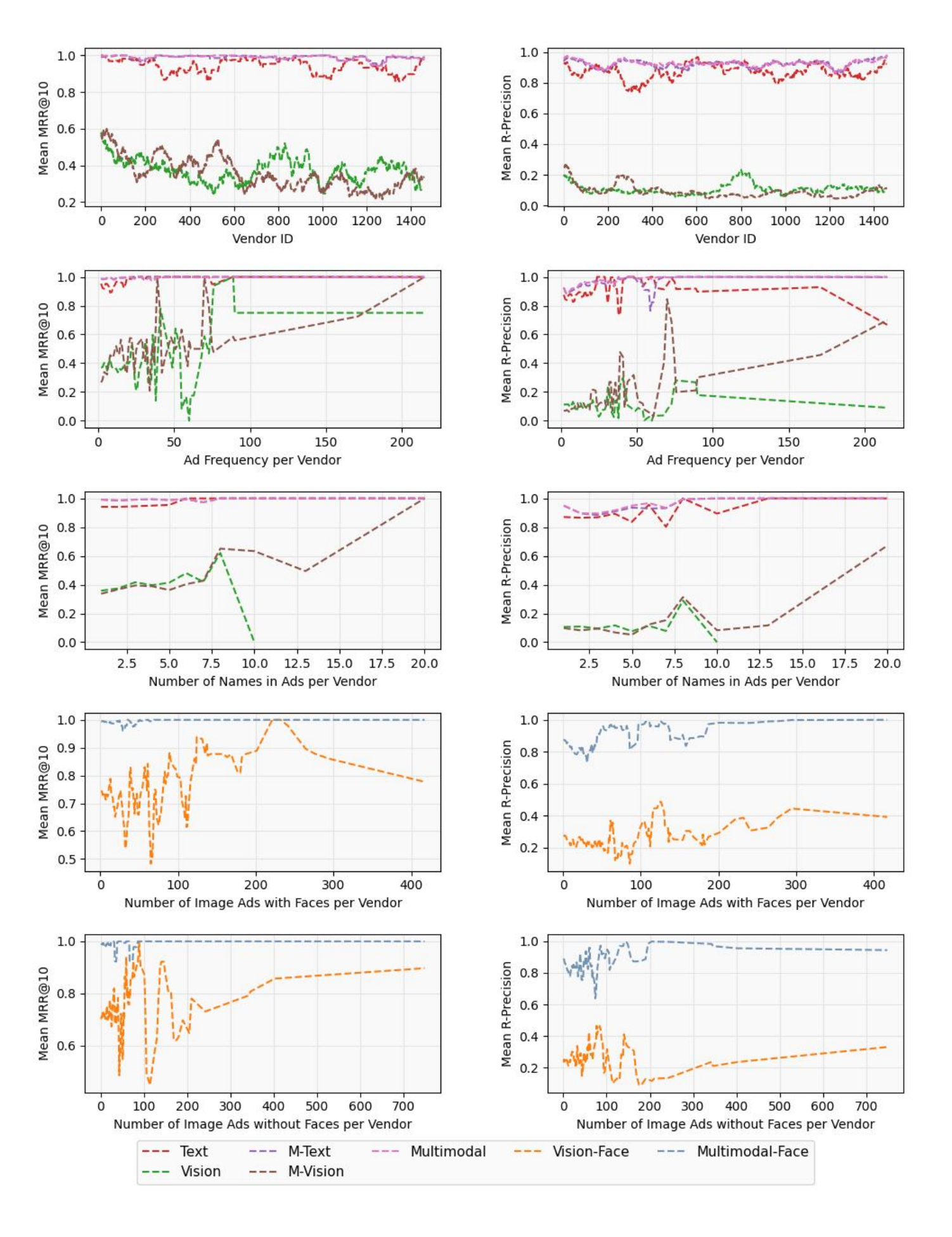}
        \caption{\label{fig:south_error_analysis_retrieval} Comparison of retrieval performance on the South region test datasets. Text, vision, and multimodal baselines (DeCLUTR-small, ViT-base-patch16-224, and DeCLUTR-ViT, respectively) are trained end-to-end for vendor identification using the joint CE+SupCon objective on the South region dataset. M-Text and M-Vision represent text-only and image-only embeddings from the multimodal system. Vision-Face and Multimodal-Face denote evaluations of escort images with and without faces.}
    \end{minipage}
\end{figure*}

\begin{figure*}[h]
    \centering
    \begin{minipage}[t]{\linewidth}
        \centering
        \includegraphics[width=\linewidth,keepaspectratio]{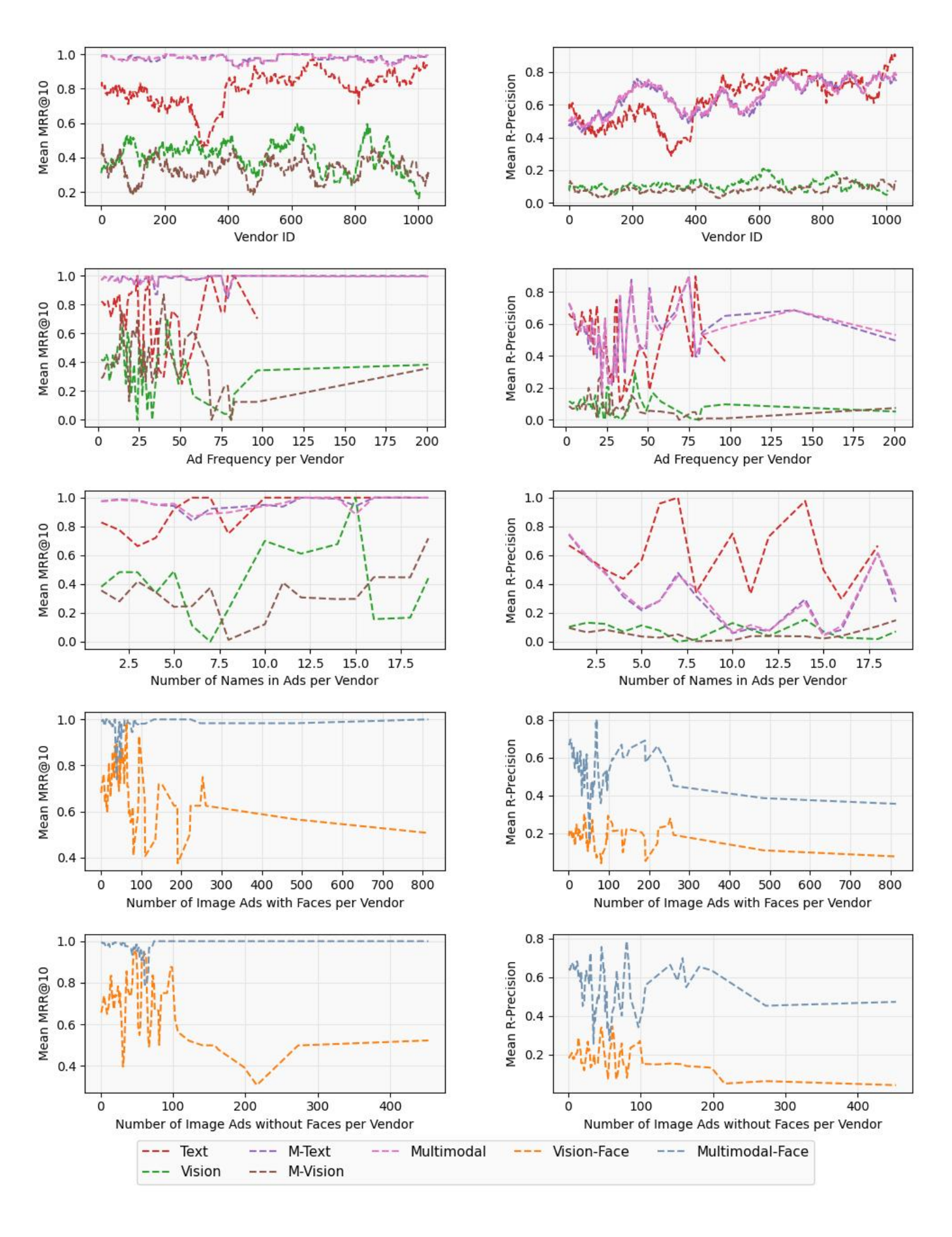}
        \caption{\label{fig:midwest_error_analysis_retrieval} Comparison of retrieval performance on the Midwest region test datasets. Text, vision, and multimodal baselines (DeCLUTR-small, ViT-base-patch16-224, and DeCLUTR-ViT, respectively) are trained end-to-end for vendor identification using the joint CE+SupCon objective on the South region dataset. M-Text and M-Vision represent text-only and image-only embeddings from the multimodal system. Vision-Face and Multimodal-Face denote evaluations of escort images with and without faces. }
    \end{minipage}
\end{figure*}

\begin{figure*}[h]
    \centering
    \begin{minipage}[t]{\linewidth}
        \centering
        \includegraphics[width=\linewidth,keepaspectratio]{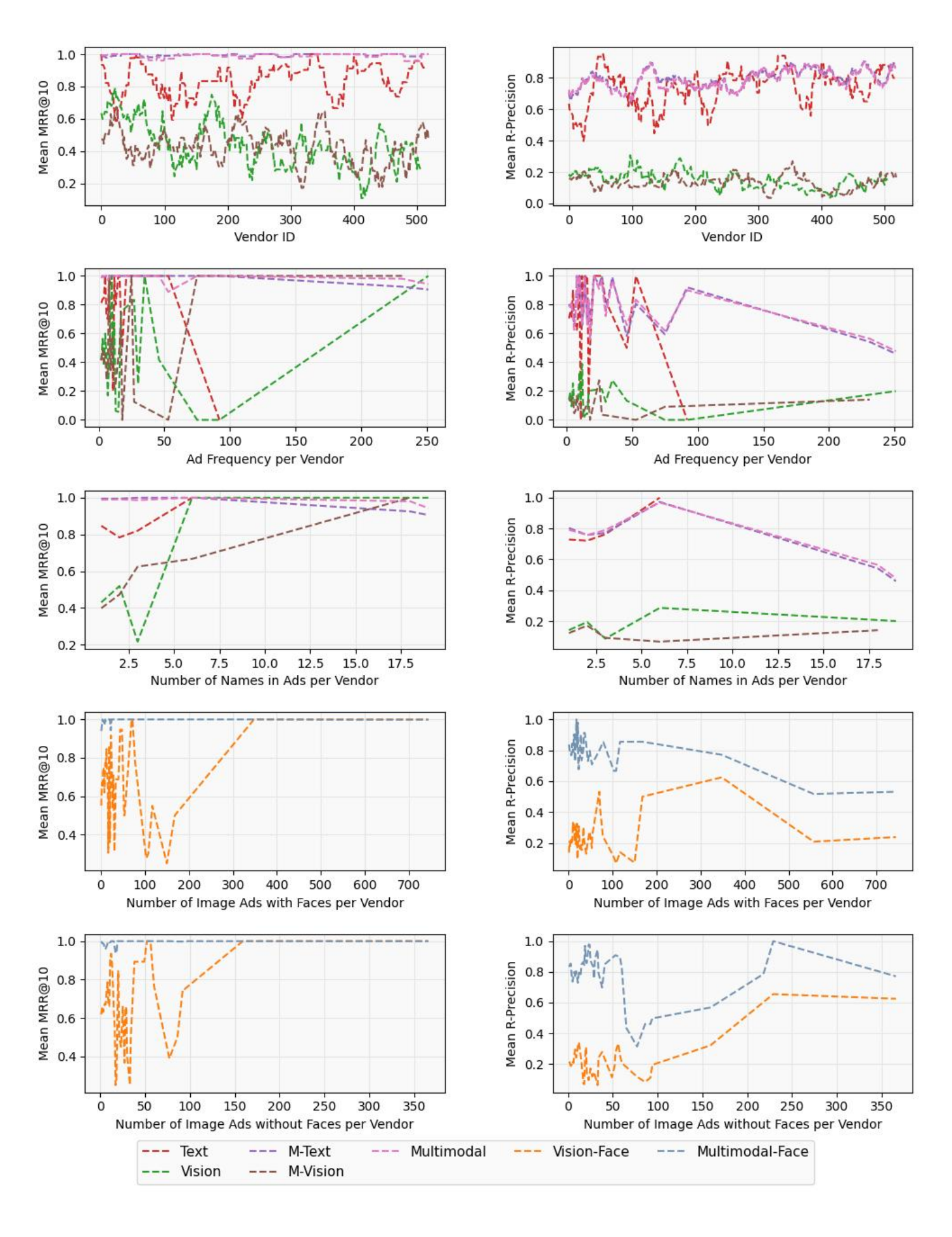}
        \caption{\label{fig:west_error_analysis_retrieval} Comparison of retrieval performance on the West region test datasets. Text, vision, and multimodal baselines (DeCLUTR-small, ViT-base-patch16-224, and DeCLUTR-ViT, respectively) are trained end-to-end for vendor identification using the joint CE+SupCon objective on the South region dataset. M-Text and M-Vision represent text-only and image-only embeddings from the multimodal system. Vision-Face and Multimodal-Face denote evaluations of escort images with and without faces. }
    \end{minipage}
\end{figure*}

\begin{figure*}[h]
    \centering
    \begin{minipage}[t]{\linewidth}
        \centering
        \includegraphics[width=\linewidth,keepaspectratio]{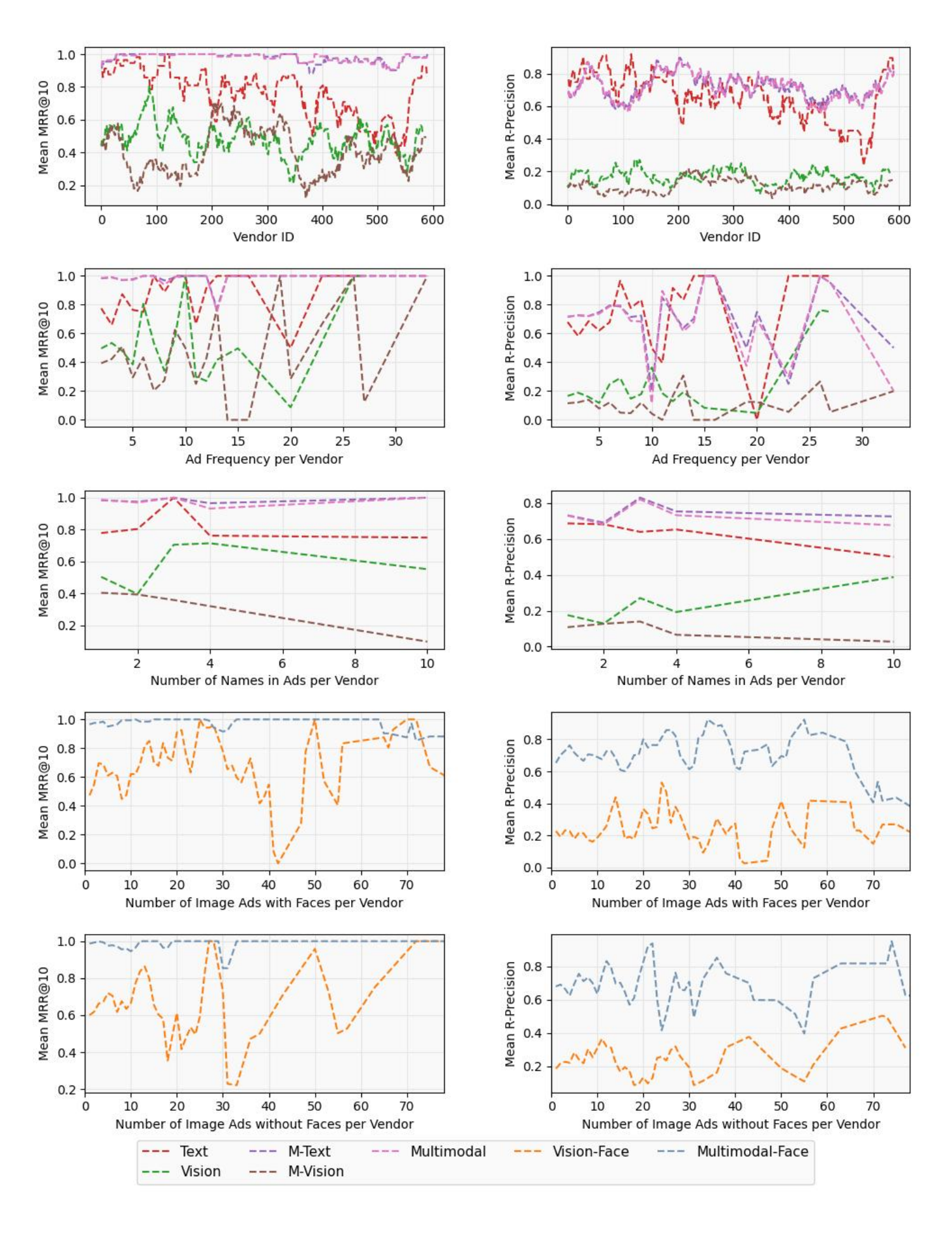}
        \caption{\label{fig:northeast_error_analysis_retrieval} Comparison of retrieval performance on the Northeast region test datasets. Text, vision, and multimodal baselines (DeCLUTR-small, ViT-base-patch16-224, and DeCLUTR-ViT, respectively) are trained end-to-end for vendor identification using the joint CE+SupCon objective on the South region dataset. M-Text and M-Vision represent text-only and image-only embeddings from the multimodal system. Vision-Face and Multimodal-Face denote evaluations of escort images with and without faces. }
    \end{minipage}
\end{figure*}

In this section, we analyze retrieval performance by comparing our multimodal DeCLUTR-ViT baseline against the text-only (DeCLUTR-small) and vision-only (ViT-base-patch16-244) baselines, all trained with the CE+SupCon objective on the South (Figure \ref{fig:south_error_analysis_retrieval}), Midwest (Figure \ref{fig:midwest_error_analysis_retrieval}), West (Figure \ref{fig:west_error_analysis_retrieval}), and Northeast (Figure \ref{fig:northeast_error_analysis_retrieval}) region datasets. To further contextualize our findings, we also evaluate the text-only (M-Text) and vision-only (M-Vision) representations extracted from the multimodal baseline, comparing their performance against the standalone text-only and vision-only baselines. Additionally, we assess the Vision-Face and Multimodal-Face baselines, which analyze the performance of vision-only and multimodal models, specifically on images with and without faces. Below, we present the consolidated insights across all regions, structured according to the key factors influencing performance: vendors, ad frequency, number of names, and the presence or absence of faces in images.

\begin{table*}
\centering
\resizebox{0.95\linewidth}{!}{%
\begin{tabular}{|l |l |c|c|c|}
\hline
\textbf{Retrieval}                                                & \textbf{Metric}      & \textbf{Midwest}                                               & \textbf{West}                                                  & \textbf{Northeast}                                             \\ \hline
                                          & \textbf{MRR@10}         & \begin{tabular}[c]{@{}c@{}}Shared: 0.7164 ± 0.41 \\ Unique: 0.7910 ± 0.37\end{tabular} & \begin{tabular}[c]{@{}c@{}}Shared: 0.8581 ± 0.33 \\ Unique: 0.8498 ± 0.33\end{tabular} & \begin{tabular}[c]{@{}c@{}}Shared: 0.7859 ± 0.38 \\ Unique: 0.7013 ± 0.42\end{tabular} \\ \cline{2-5} 
\multirow{-2}{*}{\textbf{Text-to-Text}}   & \textbf{R-Precision@X} & \begin{tabular}[c]{@{}c@{}}Shared: 0.5027 ± 0.38 \\ Unique: 0.6251 ± 0.37\end{tabular} & \begin{tabular}[c]{@{}c@{}}Shared: 0.7128 ± 0.36 \\ Unique: 0.7234 ± 0.36\end{tabular} & \begin{tabular}[c]{@{}c@{}}Shared: 0.6553 ± 0.40 \\ Unique: 0.5817 ± 0.44\end{tabular} \\ \hline
                                          & \textbf{MRR@10}         & \begin{tabular}[c]{@{}c@{}}Shared: 0.3462 ± 0.36 \\ Unique: 0.3583 ± 0.38\end{tabular} & \begin{tabular}[c]{@{}c@{}}Shared: 0.3506 ± 0.37 \\ Unique: 0.3728 ± 0.37\end{tabular} & \begin{tabular}[c]{@{}c@{}}Shared: 0.3031 ± 0.38 \\ Unique: 0.2432 ± 0.32\end{tabular} \\ \cline{2-5} 
\multirow{-2}{*}{\textbf{Image-to-Image}} & \textbf{R-Precision@X} & \begin{tabular}[c]{@{}c@{}}Shared: 0.0673 ± 0.09 \\ Unique: 0.0914 ± 0.14\end{tabular} & \begin{tabular}[c]{@{}c@{}}Shared: 0.0896 ± 0.12 \\ Unique: 0.1168 ± 0.16\end{tabular} & \begin{tabular}[c]{@{}c@{}}Shared: 0.0816 ± 0.13 \\ Unique: 0.0807 ± 0.14\end{tabular} \\ \hline
                                          & \textbf{MRR@10}         & \begin{tabular}[c]{@{}c@{}}Shared: 0.7862 ± 0.36 \\ Unique: 0.8355 ± 0.31\end{tabular} & \begin{tabular}[c]{@{}c@{}}Shared: 0.8909 ± 0.28 \\ Unique: 0.8693 ± 0.29\end{tabular} & \begin{tabular}[c]{@{}c@{}}Shared: 0.8138 ± 0.35 \\ Unique: 0.7920 ± 0.29\end{tabular} \\ \cline{2-5} 
\multirow{-2}{*}{\textbf{Multimodal}} & \textbf{R-Precision@X} & \begin{tabular}[c]{@{}c@{}}Shared: 0.5026 ± 0.35 \\ Unique: 0.6196 ± 0.34\end{tabular} & \begin{tabular}[c]{@{}c@{}}Shared: 0.7103 ± 0.33 \\ Unique: 0.7266 ± 0.33\end{tabular} & \begin{tabular}[c]{@{}c@{}}Shared: 0.6436 ± 0.37 \\ Unique: 0.5550 ± 0.41\end{tabular} \\ \hline
\end{tabular}
}
\caption{Text-to-Text, Image-to-Image, and multimodal retrieval performance for shared and unique vendors between South and  Midwest, West, and Northeast region dataset. All the representations are extracted from the multimodal DeCLUTR-ViT backbone trained with CE+SupCon objective on the South region dataset.}
\label{tab:sanity_check_unique_and_common_performance}
\end{table*}

\paragraph{Performance per Vendor:} Across all regions, the multimodal baseline consistently outperforms text-only and vision-only baselines for both MRR@10 and R-Precision@X. This performance advantage underscores the power of integrating textual and visual cues, which capture complementary information. The M-Text and M-Vision representations, extracted from the multimodal model, also outperform their respective standalone baselines. Notably, the text-only baseline performs better than the vision-only baseline, emphasizing the dominant role of text in vendor identification and retrieval tasks. However, the multimodal baseline demonstrates lower performance variability than unimodal approaches, indicating its robustness across diverse vendors. This consistency is critical for addressing real-world applications where vendor behaviors vary significantly.

\paragraph{Performance by Ad Frequency:} The relationship between retrieval performance and the frequency of ads per vendor remains consistent across regions. The multimodal baseline achieves high performance across all ad frequencies, particularly excelling for vendors with lower ad frequencies. This suggests that multimodal integration effectively compensates for data sparsity by leveraging both textual and visual features. The M-Text representation follows closely, showing a significant improvement over the standalone text-only baseline, particularly as ad frequency increases. While the vision-only baseline struggles with sparse data, the M-Vision representation extracted from the multimodal model provides a noticeable improvement, albeit still trailing behind M-Text. These results reinforce the strength of multimodal baselines in handling scenarios with limited vendor representation.

\paragraph{Performance by Number of Names:} As mentioned earlier, analyzing retrieval performance by the number of names associated with each vendor reveals the robustness of the multimodal baseline in linking ads with varied linguistic and visual patterns. Across all regions, the multimodal baseline maintains superior performance as the number of names increases, outperforming text-only and vision-only baselines. The M-Text representation consistently surpasses the standalone text-only baseline, demonstrating that multimodal training enhances the textual representation's robustness. While the vision-only baseline experiences noticeable drops in performance with increasing names, the M-Vision representation extracted from the multimodal model maintains steadier performance. These findings highlight the ability of multimodal baselines to capture stylistic and semantic variations better than unimodal baselines, which is crucial for identifying vendors with diverse aliases.

\paragraph{Performance by Images with and without Faces:} The analysis of retrieval performance based on the presence or absence of faces in images provides critical insights into the multimodal baseline's ability to leverage facial features. Across all regions, the Multimodal-Face baseline consistently outperforms the Vision-Face baseline for both MRR@10 and R-Precision@X, demonstrating its effectiveness in combining facial and textual cues. For images with faces, the multimodal baseline initially struggles as faces increase but eventually outperforms the vision-only baseline when more visual data becomes available. This trend reflects the model's ability to adapt and utilize visual information effectively when sufficient samples are present. For images without faces, the Multimodal-Face baseline consistently surpasses the Vision-Face baseline, leveraging non-facial visual patterns and textual information to improve retrieval performance. 

\subsubsection{Multimodal Retrieval Performance on Shared and Unseen Vendors in OOD Datasets}
Here, the evaluation focuses on a retrieval task, distinguishing between shared vendors—those present in the South and OOD datasets—and unknown vendors exclusive to the OOD datasets. While Figure \ref{fig:dataSimilarity} highlights an overlap of vendors between the South and OOD datasets, it is important to note that the OOD datasets were never exposed to the model during training.
Table \ref{tab:sanity_check_unique_and_common_performance} presents a detailed analysis of the model’s MRR@10 and R-Precision@X performance across text-to-text, image-to-image, and multimodal retrieval tasks for both shared and unseen vendors. Representations for these evaluations are derived from the multimodal DeCLUTR-ViT classifier trained on the South dataset. The results confirm the model’s robust performance on shared and unseen vendors, showcasing its ability to generalize effectively to unseen scenarios. This demonstrates the model’s capability to link ads to vendors, further underscoring its practical utility in real-world HT applications regardless of prior exposure to vendors and ads.

\FloatBarrier

\begin{figure*}[h]
    \centering
    
    \begin{subfigure}[b]{\textwidth}  
    \centering
    \includegraphics[width=\textwidth,keepaspectratio]{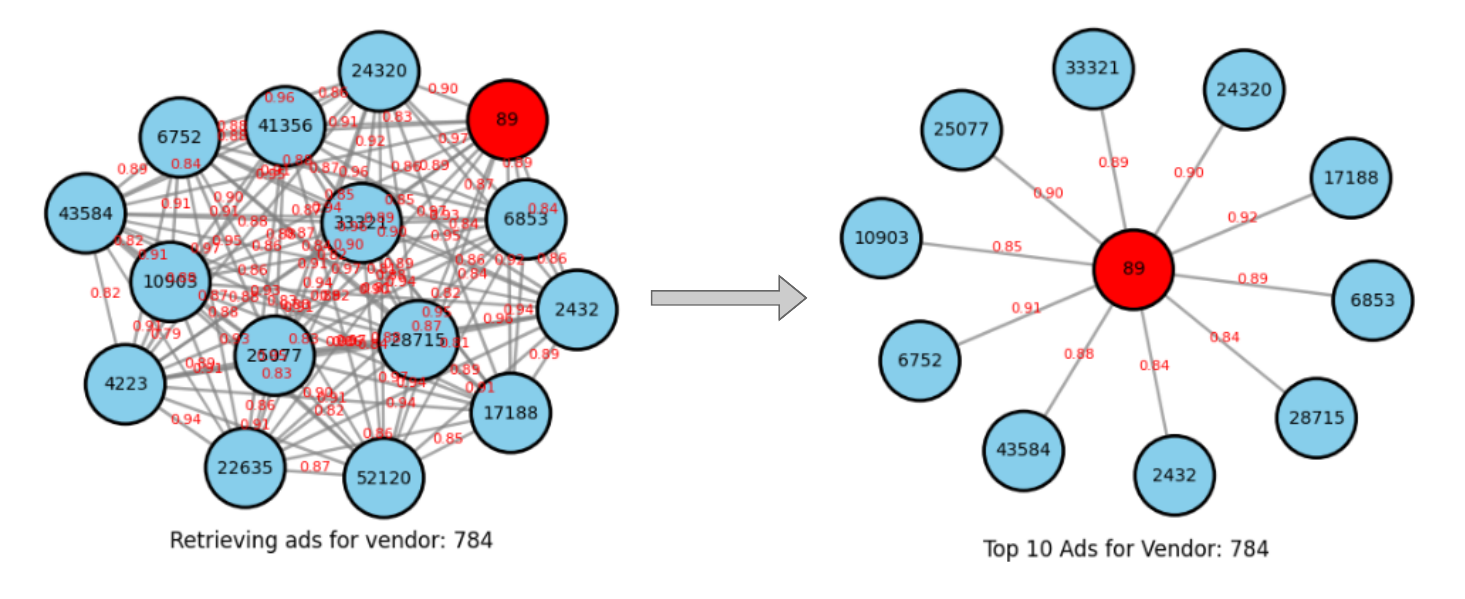}
    \caption{\label{fig:knowledgegraphvendor784} Vendor 784}
    \end{subfigure}
    
    \hspace{\textwidth} 
    
    \begin{subfigure}[b]{\textwidth}  
        \centering
        \includegraphics[width=\textwidth,keepaspectratio]{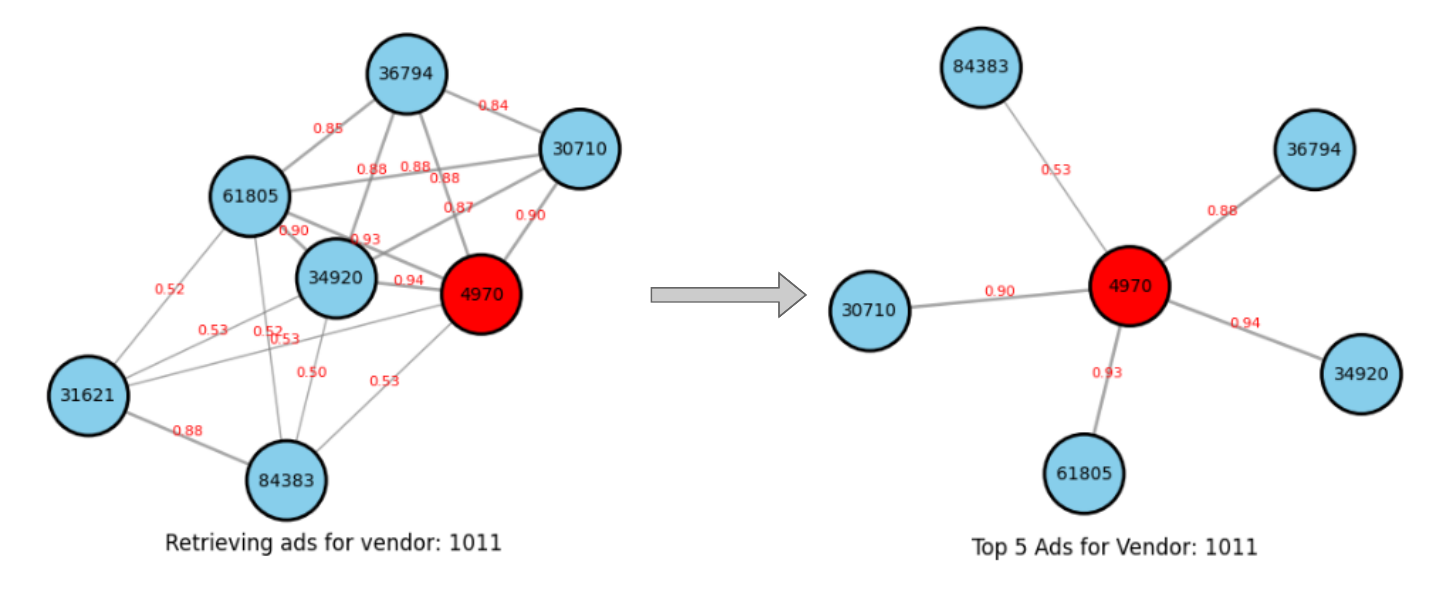}
        \caption{\label{fig:knowledgegraphvendor1101} Vendor 1101}
    \end{subfigure}
    
    \caption{Knowledge graph representation generated using AA retrieval for Vendor labels 784 and 1101 from the South region dataset. The left graph utilizes R-Precision metrics to link all relevant ads for a query ad (highlighted in red), while the right graph applies (a) MRR@10 and (b) MRR@5 to identify the top-10 most likely relevant ads. Nodes represent advertisement IDs, and edges denote the similarity between ads, both in relation to each other and the query ad, showcasing the effectiveness of AA retrieval in constructing relational insights.}
    \label{fig:knowledge_graphs}
\end{figure*}

\subsection{Practical Utility}
\label{app:utility}

To demonstrate the practical utility of our research, we employ the multimodal DeCLUTR-ViT model, trained with the CE+SupCon objective on the South region dataset, to create knowledge graphs using retrieval-based methods. The choice of representations for constructing these graphs is informed by the retrieval performance of text, vision, and multimodal embeddings on R-Precision and MRR@10 metrics. Since text-only representations from the multimodal baseline exhibit superior retrieval performance across both metrics for our dataset, we utilize them to perform our retrieval analysis.

\hspace{\parindent} Figures \ref{fig:knowledgegraphvendor784} and \ref{fig:knowledgegraphvendor1101} illustrate knowledge graphs generated for vendor labels 784 and 1101 from the South region datasets, respectively. To construct these graphs, we begin with a query advertisement (highlighted in red) and retrieve all relevant ads from the training dataset based on R-Precision performance. Each advertisement is represented as a node in the graph and labeled with its unique ID. Notably, these IDs serve as anonymous identifiers, as all personally identifiable information in the dataset has been removed using comprehensive masking techniques. Edges in the graph encode the similarity scores between connected nodes and the query advertisement, providing a quantifiable measure of relatedness. The graphs on the left of the figures depict all retrieved ads for a given query, visualizing the comprehensive network of connected advertisements for a specific vendor. To provide flexibility for researchers, investigators, and law enforcement agencies (LEAs), we propose an alternative approach using MRR@K. This allows stakeholders to retrieve the top-K most relevant ads based on similarity, enabling focused analysis depending on investigative confidence or manual verification thresholds. The resulting knowledge graphs, visualized on the right side of the figures, present a filtered view, facilitating efficient examination of high-confidence matches.

\hspace{\parindent} By leveraging these knowledge graphs, stakeholders can visualize vendor activity across advertisements, identify patterns, and establish connections, using it to initiate investigations into identifying HT identifiers.

\end{document}